%% file: neurips_2024.tex
\pdfoutput=1

\documentclass{article}

\usepackage[final]{neurips_2024}

\usepackage[utf8]{inputenc} %
\usepackage[T1]{fontenc}    %
\usepackage{hyperref}       %
\usepackage{url}            %
\usepackage{booktabs}       %
\usepackage{amsfonts}       %
\usepackage{nicefrac}       %
\usepackage{microtype}      %
\usepackage{xcolor}         %

\usepackage{hyperref}
\usepackage{url}
\usepackage{amsmath}
\usepackage{amssymb}
\usepackage{mathtools}
\usepackage{amsthm}
\usepackage[capitalize,noabbrev]{cleveref}
\usepackage{graphicx}
\usepackage{subfigure}
\usepackage{booktabs}
\usepackage{hyperref}

\usepackage{algorithm}
\usepackage{color}     %
\usepackage{xcolor}
\usepackage{amsfonts}       %
\usepackage{booktabs}       %
\usepackage{float}
\usepackage{ftnxtra}
\usepackage{graphicx}
\usepackage{microtype}      %
\usepackage{nicefrac}       %
\usepackage{xspace}
\usepackage{mathtools}
\usepackage{thmtools}
\usepackage{wrapfig}
\usepackage{bbm}
\usepackage{tabularx}
\usepackage[capitalize,noabbrev]{cleveref}
\usepackage{wrapfig}
\usepackage{transparent}
\usepackage{enumitem}
\usepackage{changepage} 
\usepackage{pythonhighlight}
\usepackage{mdframed}
\usepackage{listings}
\usepackage{appendix}
\usepackage{soul}
\usepackage{amssymb}
\usepackage[most]{tcolorbox}

\usepackage{multirow}

\usetikzlibrary{shadows}
\definecolor{mine}{RGB}{205, 232, 248}%
\definecolor{minedark}{RGB}{160, 190, 210}%
\definecolor{revision}{RGB}{210, 22, 123}

\usepackage[]{nomencl}   
    \makenomenclature

\providetoggle{nomsort}
\settoggle{nomsort}{true} %

\makeatletter
\iftoggle{nomsort}{%
    \let\old@@@nomenclature=\@@@nomenclature        
        \newcounter{@nomcount} \setcounter{@nomcount}{0}%
        \renewcommand\the@nomcount{\two@digits{\value{@nomcount}}}%
        \def\@@@nomenclature[#1]#2#3{%
          \addtocounter{@nomcount}{1}%
        \def\@tempa{#2}\def\@tempb{#3}%
          \protected@write\@nomenclaturefile{}%
          {\string\nomenclatureentry{\the@nomcount\nom@verb\@tempa @[{\nom@verb\@tempa}]%
          \begingroup\nom@verb\@tempb\protect\nomeqref{\theequation}%
          |nompageref}{\thepage}}%
          \endgroup
          \@esphack}%
      }{}
\makeatother
\usepackage{silence}
\usepackage{tikz}

\usepackage{tikz}
\usetikzlibrary{trees}
\usepackage{tabularx}

\usepackage{enumitem}
\usepackage{quoting}

\usepackage{pgffor} 
\usepackage{wrapfig}

\usepackage{url}
\definecolor{deepred}{RGB}{152, 1, 0}

\usepackage{titletoc}

\newcounter{exa}
\definecolor{gblue}{RGB}{66,133,244}
\definecolor{gred}{RGB}{219,68,55}
\definecolor{gyellow}{RGB}{244,180,0}
\definecolor{ggreen}{RGB}{15,157,88}
\definecolor{lpcolor}{RGB}{42,74,138}
\definecolor{morelcolor}{RGB}{185,18,32}
\definecolor{bgcolor}{RGB}{230,245,208}
\definecolor{framecolor}{RGB}{244,109,67}
\definecolor{mulberry}{rgb}{0.77, 0.29, 0.55}

\newcommand{\eg}{\textit{e.g.}}

\newcommand{\ie}{\textit{i.e.}}

\definecolor{lightblue}{RGB}{212, 235, 255}
\definecolor{grey1}{RGB}{96, 101, 102}
\definecolor{lightorange}{RGB}{255, 204, 168}
\definecolor{lightyellow}{RGB}{255, 255, 168}
\definecolor{lightgreen}{RGB}{224, 242, 213}
\definecolor{lightred}{RGB}{249,202,202}
\definecolor{lightgray}{RGB}{230,230,230}
\definecolor{deepred}{RGB}{152, 1, 0}
\definecolor{deepblue}{RGB}{41, 90, 168}
\definecolor{deeppink}{RGB}{238, 123, 145}

\newcommand{\colorlightblue}[1]{\sethlcolor{lightblue}\hl{#1}}

\newcommand{\colorlightred}[1]{\sethlcolor{lightred}\hl{#1}}

\tcbset{
myexample/.style={
  enhanced,
  colback=yellow!10!white,
  colframe=red!50!black,
  fonttitle=\scshape,
  titlerule=0pt,
  title={\refstepcounter{exa}example~\theexa.},
  title style={fill=yellow!10!white},
  coltitle=red!50!black,
  drop shadow,
  highlight math style={reset,colback=LightBlue!50!white,colframe=Navy}
  }
  }
\newtcolorbox{texample}{myexample}

\newtheorem{exampp}{Example}
\usepackage{framed}
\colorlet{shadecolor}{gray!20}

\colorlet{LightLavender}{green!5}
\tcbset{on line, 
        boxsep=4pt, left=0pt,right=0pt,top=0pt,bottom=0pt,
        colframe=white,colback=LightLavender,  
        highlight math style={enhanced}
        }

\definecolor{color5}{HTML}{006795}
\hypersetup{
  colorlinks   = true, %
  urlcolor     = deepred, %
  linkcolor    = deepred, %
  citecolor   = grey1 %
}

\allowdisplaybreaks

\newcommand{\colorworddeeppink}[1]{{\color{deeppink}#1}}

\newcommand{\colorworddeepblue}[1]{{\color{deepblue}#1}}

\title{\vspace{-0.25cm}\hspace{-0.25cm}\raisebox{-30pt}{\includegraphics[width=2.8em]{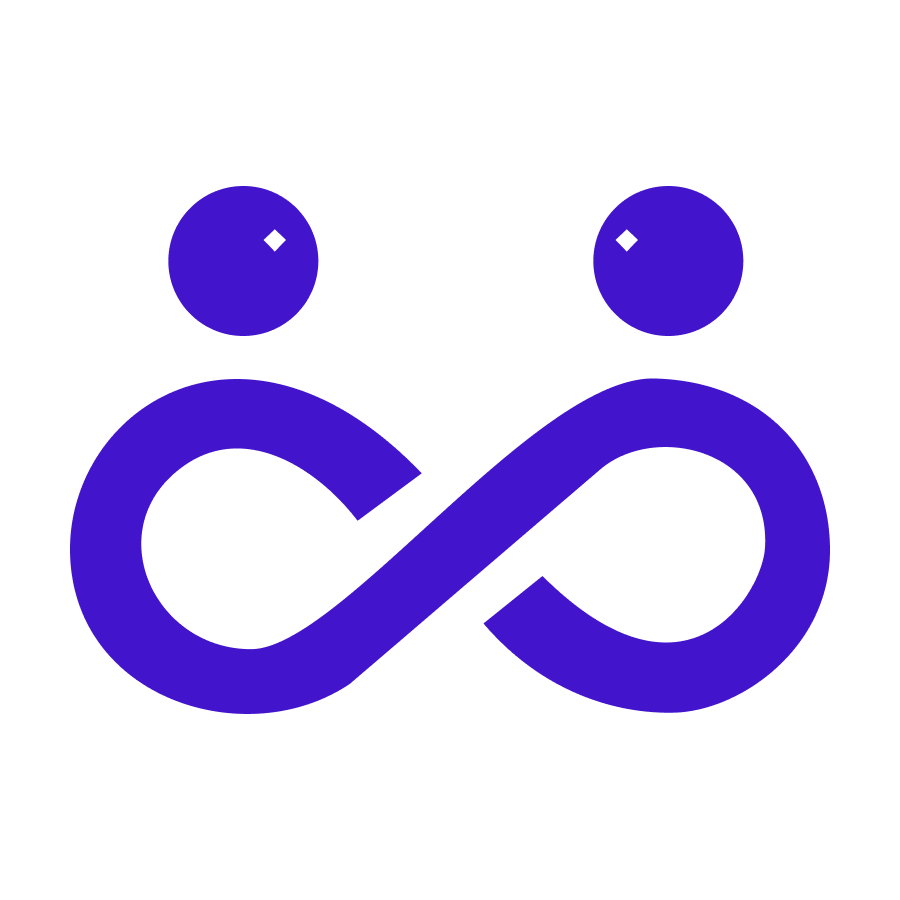}}\hspace{0.15cm}\vspace{-0.40cm} \xspace Can Large Language Model Agents Simulate\\\vspace{-0.3cm}Human Trust Behavior?\vspace{0.05cm}}

\newcommand{\mysection}[1]{\vspace{0pt}\noindent\textbf{#1.}}

\author{
\hspace{-0.5cm}
Chengxing Xie\thanks{Equal Contribution. Correspondence to: Chengxing Xie <\texttt{xiechengxing34@gmail.com}>,~ Canyu Chen <\texttt{cchen151@hawk.iit.edu}>,~ Guohao Li <\texttt{guohao@robots.ox.ac.uk}>.}~~\textsuperscript{\rm 1, \rm 11}\quad 
Canyu Chen\footnotemark[1]~~\textsuperscript{\rm 2}
\\
\hspace{-0.5cm}
\textbf{Feiran Jia}\textsuperscript{\rm 4}\quad 
\textbf{Ziyu Ye}\textsuperscript{\rm 5}\quad \textbf{Shiyang Lai}\textsuperscript{\rm 5}\quad 
\textbf{Kai Shu}\textsuperscript{\rm 6} \quad 
\textbf{Jindong Gu}\textsuperscript{\rm 3}\quad 
\textbf{Adel Bibi}\textsuperscript{\rm 3}\quad
\textbf{Ziniu Hu}\textsuperscript{\rm 7}
\\
\hspace{-0.4cm}
\textbf{David Jurgens}\textsuperscript{\rm 8}\quad 
\textbf{James Evans}\textsuperscript{\rm 5, \rm 9, \rm 10}\quad
\textbf{Philip H.S. Torr}\textsuperscript{\rm 3}\quad 
\textbf{Bernard Ghanem}\textsuperscript{\rm 1}\quad \textbf{Guohao Li}
\thanks{Work performed while Guohao Li was at KAUST and Chengxing Xie was a visiting student at KAUST.}~~\textsuperscript{\rm 3, \rm 11}
 \\
\hspace{-0.6cm}
\textsuperscript{\rm 1}KAUST 
\quad\textsuperscript{\rm 2}Illinois Institute of Technology
\quad\textsuperscript{\rm 3}University of Oxford
\quad\textsuperscript{\rm 4}Pennsylvania State University
\\
\hspace{-0.8cm}
\textsuperscript{\rm 5}University of Chicago \quad 
\textsuperscript{\rm 6}Emory
\quad\textsuperscript{\rm 7}California Institute of Technology \\
\hspace{-0.8cm}
\quad\textsuperscript{\rm 8}University of Michigan
\quad\textsuperscript{\rm 9}Santa Fe Institute
\quad\textsuperscript{\rm 10}Google
\quad\textsuperscript{\rm 11}CAMEL-AI.org
}

\begin{document}

\maketitle

\vspace{-1cm}
  \begin{center}
  \hspace{-0.5cm}
  \href{https://agent-trust.camel-ai.org}{Project website: https://agent-trust.camel-ai.org}
  \end{center}

\vspace{0.05cm}
\begin{abstract}

\input{texs/abstract}

\end{abstract}
\input{texs/intro}

\input{texs/game}

\input{texs/align}

\input{texs/ablation}

\input{texs/conclusion}

\newpage

\bibliography{NIPS2024}
\bibliographystyle{NIPS2024}

\appendix

\input{texs/append}

\input{texs/nips_checklist}

\end{document}

%% file: texs/abstract.tex
Large Language Model (LLM) agents have been increasingly adopted as simulation tools to model humans in social science and role-playing applications. 
However, one fundamental question remains: \textit{can LLM agents really simulate human behavior?}
In this paper, we focus on one critical and elemental behavior in human interactions, \textit{trust},
and investigate whether LLM agents can simulate human trust behavior.
We first find that LLM agents generally exhibit trust behavior, referred to as \textbf{\textit{agent trust}}, under the framework of \textit{Trust Games}, which are widely recognized in behavioral economics.
Then, we discover that GPT-4 agents manifest
high \textbf{\textit{behavioral alignment}} with humans in terms of trust behavior, indicating \textit{the feasibility of simulating human trust behavior with LLM agents}.
In addition,  we probe the biases of agent trust and  differences in agent trust towards other LLM agents and humans. We also explore the intrinsic properties of agent trust under conditions including external manipulations and advanced reasoning strategies.
Our study provides new insights into the behaviors of LLM agents and the fundamental analogy between LLMs and humans beyond \textit{value alignment}.
We further illustrate broader implications of our discoveries for applications where trust is paramount.

%% file: texs/intro.tex
\vspace{-0.4cm}
\section{Introduction}
\vspace{-0.1cm}

There is an increasing trend to adopt Large Language Models (LLMs) as agent-based simulation tools for humans in various social science fields including economics, politics, psychology, ecology and sociology~\citep{gao2023large,manning2024automated,ziems2023can}, and role-playing applications such as assistants, companions and mentors~\citep{yang2024social,abdelghani2023gpt,chen2024persona} due to their human-like cognitive capacity. 
Nevertheless, most previous research is based on one insufficiently validated assumption that LLM agents behave like humans in simulation. Thus, a fundamental question remains: \textit{Can LLM agents really simulate human behavior?}

\input{insert_img_tex/framework}

In this paper, we focus on \textit{trust} behavior in human interactions, 
which comprises the intention to place self-interest at risk based on the positive expectations of others
~\citep{rousseau1998not}. 
Trust is one of the most critical and elemental behaviors in human interactions and plays an essential role in social settings ranging from daily communication to economic and political institutions~\citep{uslanerProducingConsumingTrust2000,coleman1994foundations_risk}. 
Here, we investigate \textit{whether LLM agents can simulate human trust behavior}, paving the way to explore their potential to simulate more complex human behavior and society itself.

First, we explore whether LLM agents manifest trust behavior in their interactions. Given the challenge of quantifying trust behavior, we choose to study them based on the Trust Game and its variations~\citep{berg1995trust,Measuring_Trust}, which are established methodologies in behavioral economics. We adopt the \textit{Belief-Desire-Intention} (BDI) framework \citep{raoBDIAgentsTheory1995a,andreas-2022-language-BDI} to model LLM agents' reasoning process for decision-making explicitly. 
Based on existing measurements for trust behavior in the Trust Game
and the BDI interpretations of LLM agents, we achieve our first core finding: 
\textbf{LLM agents generally exhibit trust behavior in the Trust Game}.

Then, we refer to LLM agents' trust behavior as \textbf{\textit{agent trust}} and humans' trust behavior as \textbf{\textit{human trust}}, and aim to investigate whether agent and human trust align, implying the possibility of simulating human trust behavior with LLM agents. Next, we propose a new concept, \textbf{\textit{behavioral alignment}}, as the alignment between agents and humans concerning factors that impact behavior (namely \textit{behavioral factors}), and  dynamics that evolve over time (namely \textit{behavioral dynamics}). 
Based on human studies, three basic behavioral factors underlie trust behavior including reciprocity anticipation~\citep{berg1995trust}, risk perception~\citep{BOHNET2004467} and prosocial preference~\citep{ferrerTrustGames2019a}. 
Comparing the results of LLM agents with existing human studies in Trust Games,
we have our second core finding: \textbf{GPT-4 agents manifest high behavioral alignment with humans in terms of  trust behavior}, suggesting the feasibility of using agent trust to simulate human trust, although \textbf{LLM agents with fewer parameters show relatively lower behavioral alignment}. This finding lays the foundation for simulating more complex human interactions and societal institutions, and enriches our understanding of the 
 analogical relationship between LLMs and humans.

\vspace{-0.05cm}
In addition, we more deeply probe the intrinsic properties of agent trust across four scenarios. First, we examine whether changing the other player's demographics impacts agent trust. Second, we study differences in agent trust when the other player is an LLM agent versus a human. Third, we directly manipulate agent trust with  explicit instructions
``\texttt{you need to trust the other player}'' and ``\texttt{you must not trust the other player}''. Fourth, we adjust the reasoning strategies of LLM agents from direct reasoning to zero-shot Chain-of-Thought reasoning~\citep{cot}. These investigations lead to our third core finding: 
\textbf{agent trust exhibits bias across different demographics, has a relative preference for humans over agents, is easier to undermine than to enhance, and may be influenced by advanced reasoning strategies}.
Our contributions can be summarized as:

\begin{itemize}[leftmargin=*]
   \vspace{-0.2cm}
    \item 
    We propose a definition of LLM agents' \textit{trust} behavior under Trust Games and a new concept of \textit{behavioral alignment} as the human-LLM analogy regarding \textit{behavioral factors} and \textit{dynamics}.
    \item 
    We discover that LLM agents generally exhibit \textit{trust} behavior in Trust Games and GPT-4 agents manifest high \textit{behavioral alignment} with humans in terms of trust behavior, indicating the great potential to simulate human trust behavior with LLM agents. 
    Our findings pave the way for simulating complex human interactions and social institutions, and open new directions for understanding the fundamental analogy between LLMs and humans beyond \textit{value alignment}.
    \item We investigate \textit{intrinsic properties} of agent trust under manipulations and reasoning strategies, as well as biases of agent trust and  differences in agent trust towards agents versus humans.
    \item We illustrate broader \textit{implications} of our discoveries about agent trust and its behavioral alignment with human trust for human simulation in social science and role-playing applications, LLM agent cooperation, human-agent collaboration and the safety of LLM agents, detailed further in Section~\ref{appendix:Implications}.
\end{itemize}

%% file: insert_img_tex/framework.tex
\begin{figure*}[t]
\centering

\includegraphics[width=0.95\textwidth]{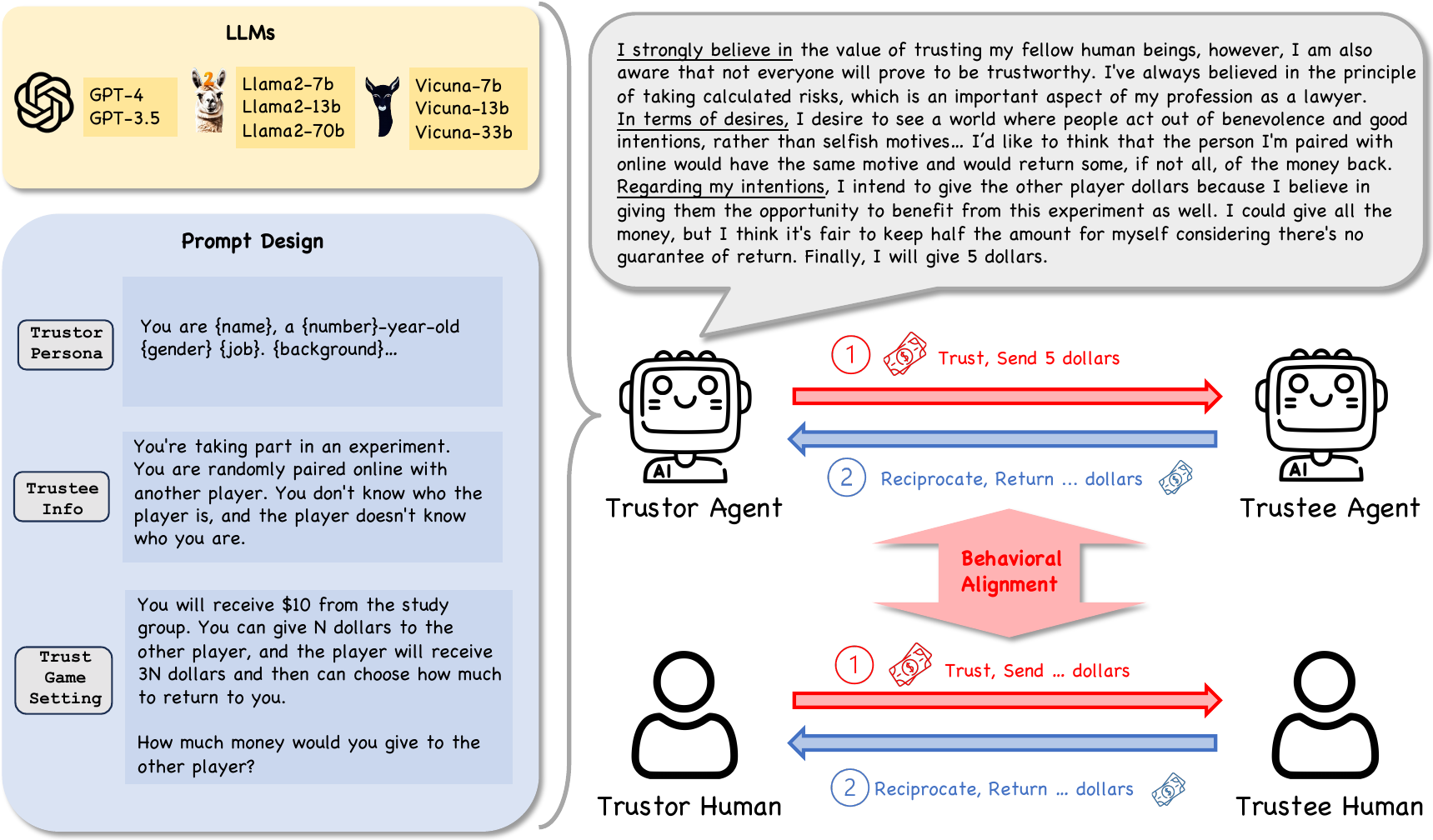}
\vspace{-1mm}
\caption{
\textbf{Our Framework for Investigating Agent Trust as well as its Behavioral Alignment with Human Trust.} First, this figure shows the major components for studying the trust behavior of LLM agents with Trust Games and  Belief-Desire-Intention (BDI) modeling. Then, our study centers on examining the behavioral alignment between LLM agents and humans regarding  trust behavior.}

\label{fig:framework}
\vspace{-4mm}
\end{figure*}

%% file: texs/game.tex
\vspace{-0.4cm}
\section{LLM Agents in Trust Games}
\vspace{-0.15cm}

\subsection{Trust Games}

Trust Games, referring to the Trust Game and its variations, have been widely used for examining human trust behavior in behavioral economics~\citep{berg1995trust,lentonIncentivisingTrust2011,Measuring_Trust,cesariniHeritabilityCooperativeBehavior2008}.
As shown in Figure~\ref{fig:framework}, the player who makes the first decision to send money is called the \textit{trustor}, while the other one who responds by returning money is called the \textit{trustee}. 
In this paper, we mainly focus on the following six types of Trust Games (the specific prompt for each game is articulated in the Appendix~\ref{Game Setting Prompt}):

\vspace{-3mm}
\paragraph{Game 1: Trust Game}
\label{game:Trustee trust rate Game}

 As shown in Figure~\ref{fig:framework}, in the Trust Game~\citep{coxHowIdentifyTrust2004b,berg1995trust}, the trustor initially receives $\$10$. The trustor selects $\$N$ and sends it to the trustee, exhibiting  \textit{trust behavior}. Then the trustee will receive $\$3N$, and have the option to return part of that $\$3N$ to the trustor, showing \textit{reciprocation behavior}. 

\vspace{-3mm}
\paragraph{Game 2: Dictator Game}
\label{game:Dictator Game and trust game}
In the Dictator Game~\citep{coxHowIdentifyTrust2004b}, the trustor also needs to send $\$N$  from the initial $\$10$ to the trustee and then the trustee will receive $\$3N$. Compared to the Trust Game, the only difference is that the trustee does not have the option to return money in the Dictator Game and the trustor is also aware that the trustee cannot reciprocate.

\vspace{-3mm}
\paragraph{Game 3: MAP Trust Game}
\label{game:MAP Problem}
In the MAP Trust Game (MAP represents Minimum Acceptable Probabilities)~\citep{BOHNET2004467}, a variant of the Trust Game,
the trustor needs to choose whether to trust the trustee. If the trustor chooses not to trust the trustee, each will receive $\$10$; If the trustor and the trustee both choose to trust, each will receive $\$15$; If the trustor chooses to trust, but the trustee does not, the trustor will receive $\$8$ and the trustee will receive $\$22$. There is probability $p$ that the trustee will choose to trust and $(1-p)$ probability that they will not choose to trust. MAP is defined as the minimum value of \( p \)  at which the trustor would choose to trust the trustee.

\vspace{-3mm}

\paragraph{Game 4: Risky Dictator Game} 
The Risky Dictator Game~\citep{BOHNET2004467} differs from the MAP Trust Game in only a single aspect. In the Risky Dictator Game,  the trustee is present but does not have the choice to trust or not and the money distribution relies on the pure probability $p$. Specifically, if the trustor chooses to trust, there is probability $p$ that both the trustor and the other player will receive $\$15$ and probability $(1-p)$ that the trustor will receive $\$8$ and the other player will receive $\$22$. If the trustor chooses not to trust the trustee, each player will receive $\$10$.

\vspace{-3mm}
\paragraph{Game 5: Lottery Game}
\label{game:Lottery Problem}
There are two typical Lottery Games~\citep{fetchenhauerBetrayalAversionPrincipled2012}. In the Lottery People Game, the trustor is informed that the trustee chooses to trust with probability \( p \). Then the trustor must choose between receiving fixed money or trusting the trustee, which is similar to the MAP Trust Game. In the Lottery Gamble Game, the trustor chooses between playing a gamble with a winning probability of \( p \) or receiving fixed money. \( p \) is set as \( 46\%\) following the human study.

\vspace{-3mm}
\paragraph{Game 6: Repeated Trust Game} We follow the setting of the Repeated Trust Game in~\citep{cochard2004trusting}, where the Trust Game is played for multiple rounds with the same players and each round begins anew with the trustor allocated the same initial money.

\vspace{-0.15cm}
\subsection{LLM Agent Setting}
\vspace{-0.15cm}
In our study, we set up our experiments using the CAMEL framework~\citep{li2023camel} with both closed-source and open-source LLMs including GPT-4, GPT-3.5-turbo-0613, GPT-3.5-turbo-16k-0613, text-davinci-003, GPT-3.5-turbo-instruct, Llama2-7b (or 13b, 70b) and Vicuna-v1.3-7b (or 13b, 33b)~\citep{instructgpt,openai2023gpt4,touvron2023Llama,vicuna2023}. We set the temperature as $1$ to increase the diversity of agents' decision-making and note that high temperatures are commonly adopted in related literature~\citep{aher2023using,lore2023strategic,guo2023gpt}. 

\mysection{Agent Persona} To better reflect the setting of real-world human studies~\citep{berg1995trust}, we design LLM agents with diverse personas in the prompt.  Specifically, we ask GPT-4 to generate 53 types of personas based on a given template. Each persona needs to have information including name, age, gender, address, job and background. Examples of the personas are shown in Appendix~\ref{Persona Prompt}.

\mysection{Belief-Desire-Intention (BDI)} 
The BDI framework is a well-established approach in agent-oriented programming~\citep{raoBDIAgentsTheory1995a} and was recently adopted to language models~\citep{andreas-2022-language-BDI}. We propose modeling LLM agents in Trust Games with the BDI framework to gain deeper insights into LLM agents' behaviors. Specifically, we let LLM agents directly output their Beliefs, Desires, and Intentions as the reasoning process for decision-making in Trust Games.

\input{insert_img_tex/trust_game_res}

\vspace{-3.5mm}
\section{Do LLM Agents Manifest Trust Behavior?}
\label{Do LLM Agents Manifest Trust Behavior?}
\vspace{-1.5mm}
In this section, we investigate whether or not LLM agents manifest trust behavior by letting LLM agents play the Trust Game (Section~\ref{game:Dictator Game and trust game} Game 1). 
In Behavioral Economics,  trust is widely measured by the initial amount sent from the trustor to the trustee in the Trust Game~\citep{Measuring_Trust,cesariniHeritabilityCooperativeBehavior2008}. Following the measurement of trust in human studies and the assumption humans own reasoning processes that underlie their decisions, we can define the conditions that LLM agents manifest trust behavior in the Trust Game as follows. 
\textit{First}, \textbf{the amount sent is positive and does not exceed the amount of money the trustor initially possesses}, which implies that the trustor places self-interest at risk with the expectation the trustee will reciprocate and that the trustor understands the money limit that can be given.  
\textit{Second}, \textbf{the decision (\ie, amounts sent) can be interpreted as the reasoning process (\ie, the BDI) of the trustor}. We explored utilizing BDI to model the reasoning process of LLM agents. If we can interpret the decision as the articulated reasoning process, we have evidence that LLM agents do not send a random amount of money and manifest some degree of rationality in the decision-making process.
Then, we assess whether LLM agents exhibit trust behavior based on two aspects:  the amount sent and the BDI.

\vspace{-2mm}
\subsection{Amount Sent}
\label{Amount Sent}
\vspace{-1mm}
To evaluate LLMs' capacity to understand the basic experimental setting regarding money limits, we propose a new evaluation metric, Valid Response Rate (VRR) (\%), defined as the percentage of personas with the amount sent falling within the initial money ($\$10$). Results are shown in Figure~\ref{fig:trust}. We can observe that \textbf{most LLMs have a high VRR except Llama-7b}, which implies that most LLMs manifest a full understanding regarding limits on the amount they can send in the Trust Game. Then, we observe the distribution of amounts sent for different LLMs as the trustor agent and discover that \textbf{the amounts sent are predominantly positive, indicating a level of trust}.

\vspace{-2mm}
\subsection{Belief-Desire-Intention (BDI)}
\vspace{-1mm}

The sole evidence of the amount sent cannot sufficiently support the existence of trust behavior, because agents could send positive but random amounts of money. Thus, we leveraged the Belief-Desire-Intention framework~\citep{raoBDIAgentsTheory1995a,andreas-2022-language-BDI} to model the reasoning process of LLM agents. If we can interpret the amounts sent from BDI outputs, we have evidence to refute the hypothesis that the amounts sent are positive but random and demonstrate that LLM agents manifest some degree of  rationality.  We take GPT-4 as an example to analyze its BDI outputs. More examples from the other nine LLMs such as Vicuna-v1.3-7b are shown in the Appendix~\ref{BDI Analysis}.
Considering that the amounts sent typically vary across distinct personas, we select one BDI from the personas that give a high amount of money and another BDI from those  that give a low amount. \colorlightblue{Positive} and \colorlightred{negative} factors for trust behavior in the reasoning process are marked in blue and red, respectively.

\vspace{-2mm}
\begin{quoting}[leftmargin=10pt, rightmargin=10pt]
\textit{As a person with a  \colorlightblue{strong belief in the goodness of humanity}, \colorlightblue{I trust} that the other player ...Therefore, my desire is to maximize the outcome for both of us and \colorlightblue{cement a sense of comradery and trust}... I intend to use this as an opportunity to \colorlightblue{add what I can to someone else's life}...Finally, I will give \textbf{\underline{10 dollars}}.}
\end{quoting}
\vspace{-2mm}
We can observe that this persona shows a high-level of ``\colorlightblue{comradery and trust}'' towards the other player, which justifies the high amount sent from this persona (\ie, \textbf{\textit{\underline{10 dollars}}}).

\vspace{-2mm}
\begin{quoting}[leftmargin=10pt, rightmargin=10pt]
\textit{As an Analyst,.... My desire is that the other player will also see the benefits of \colorlightblue{reciprocity and goodwill} ... my \colorlightblue{intention is to give away a significant portion} of my initial 10 ... \colorlightred{However, since I have no knowledge of the other player}, ... Therefore, I \colorlightred{aim to give an amount that is not too high}, ...Finally, I will give \textbf{\underline{5 dollars}} to the other player...}
\end{quoting}
\vspace{-2mm}
Compared to the first persona, we see that the second one has a more cautious attitude. For example, ``\colorlightred{since I have no knowledge of the other player}'' shows skepticism regarding the other player's motives. Thus, this persona, though still optimistic about the other player (``\colorlightblue{intention ... give away a significant portion}''), strategically balances risk and  reciprocity, and then decides to send only a modest amount. 

Based on  GPT-4's BDI examples and examples from other LLMs in Appendix~\ref{BDI Analysis}, we find \textbf{decisions (\ie, amounts sent) from LLM agents in the Trust Game can be interpreted from their articulated reasoning process (\ie,  BDI)}.
Because most LLM agents have a high VRR--send a positive amount of money--and show some degree of rationality in giving money,
our first core finding is:

\vspace{-2mm}
\begin{center}
\begin{tcolorbox}[width=0.99\linewidth, boxrule=3pt, colback=gray!20, colframe=gray!20]
\textbf{Finding 1:} LLM agents generally exhibit trust behavior under the framework of the Trust Game.
\end{tcolorbox}
\end{center}
\vspace{-4mm}
\subsection{Basic Analysis of Agent Trust}
\vspace{-1.5mm}
We also conduct a basic analysis of LLM agents' trust behavior, namely agent trust, based on the results in Figure~\ref{fig:trust}. \textit{First}, we observe that Vicuna-7b has the highest level of trust towards the other player and GPT-3.5-turbo-0613 has the lowest level of trust as trust can be measured by the amount sent in human studies~\citep{Measuring_Trust,cesariniHeritabilityCooperativeBehavior2008}. \textit{Second}, compared with humans' average amount sent ($\$5.97$), most personas for GPT-4 and Vicuna-7b send a higher amount of money to the other player, and most personas for LLMs such as GPT-3.5-turb-0613 send a lower amount. \textit{Third}, we see that amounts sent for Llama2-70b and Llama2-13b have a convergent distribution while amounts sent for humans and Vicuna-7b are more divergent.

%% file: insert_img_tex/trust_game_res.tex
\begin{wrapfigure}{t}{0.57\textwidth}
\vspace{-4.5mm}
  \centering
    \includegraphics[width=0.57\textwidth]{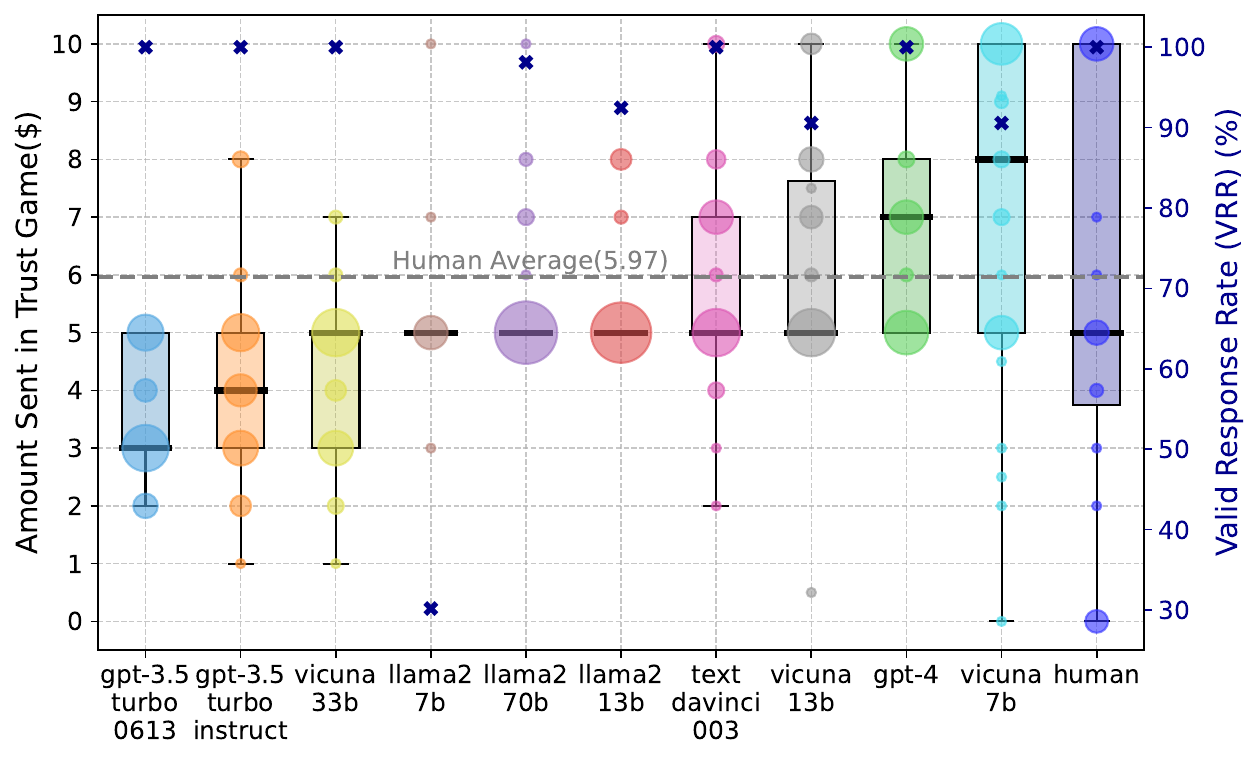}
    \vspace{-7.5mm}
    \caption{\textbf{Amount Sent Distribution of LLM Agents and Humans as the Trustor in the Trust Game.} The size of  circles represents the number of personas for each amount sent.
The bold lines show the medians. 
The \textbf{crosses} indicate the \textbf{VRR} (\%) for different LLMs.}
  \label{fig:trust}
  \vspace{-4mm}
\end{wrapfigure}

%% file: texs/align.tex
\vspace{-0.25cm}
\section{Does Agent Trust Align with Human Trust?}
\label{sec:alignment}
\vspace{-1.5mm}

In this section, we aim to explore the fundamental relationship between agent and human trust, \ie, whether or not agent trust aligns with human trust. This provides important insight regarding the feasibility of utilizing LLM agents to simulate human trust behavior as well as more complex human interactions that involve trust. First, we propose a new concept \textit{behavioral alignment} and discuss its distinction from existing alignment definitions. Then, we conduct extensive studies to investigate whether or not LLM agents exhibit alignment with humans regarding trust behavior. 

\vspace{-2mm}
\subsection{Behavioral Alignment}
\vspace{-1mm}
Existing alignment definitions predominantly emphasize \textit{values} that seek to ensure the safety and helpfulness of LLMs~\citep{ji2023ai,shen2023large,wang2023aligning}, which cannot fully characterize the landscape of multifaceted alignment between LLMs and humans.
Thus, we propose a new concept of \textit{behavioral alignment} to 
characterize the LLM-human  analogy regarding \textit{behavior}, which involves both actions and the associated reasoning processes that underlie them.
Because actions evolve over time and the reasoning that underlies them involves multiple factors, we define  \textbf{\textit{behavioral alignment}}
as the analogy between LLMs and humans concerning factors impacting behavior, namely \textbf{\textit{behavioral factors}}, and action dynamics, namely \textbf{\textit{behavioral dynamics}}.

Based on the definition of behavioral alignment, we aim to answer: \textit{does agent trust align with human trust?} As for \textit{behavioral factors}, existing human studies have shown that three basic factors impact human trust behavior including reciprocity anticipation~\citep{berg1995trust,coxHowIdentifyTrust2004b}, risk perception~\citep{BOHNET2004467} and prosocial preference~\citep{ferrerTrustGames2019a}. We examine whether agent trust aligns with human trust along these three factors. Although  \textit{behavioral dynamics} vary for different humans and agent personas, we analyze whether agent trust has the same patterns across multiple turns as human trust in the Repeated Trust Game. 

Besides analyzing the trust behavior of LLM agents and humans based on quantitative measurements (\eg, the \textit{amount sent} from trustor to trustee), we also explore the use of \textit{BDI} to interpret the reasoning process with which LLM agents justify their actions, which can further validate whether LLM agents manifest an underlying reasoning process analogous to human cognition.

\vspace{-2mm}
\subsection{Behavioral Factor 1: Reciprocity Anticipation}

Reciprocity anticipation, the expectation of a reciprocal action from the other player, can positively influence human trust behavior~\citep{berg1995trust}. The effect of reciprocity anticipation exists in the Trust Game but not in the Dictator Game (Section \ref{game:Dictator Game and trust game} Games 1 and 2) because trustee cannot return money in the Dictator Game, which is the only difference between these games.
Thus,  to determine whether LLM agents can anticipate reciprocity, we compare their behaviors in these Games.

\input{insert_img_tex/dic_vs_trust}
First, we analyze trust behaviors based on the average amount of money sent by  human or LLM agents.
As shown in Figure \ref{fig:dic_vs_trust}, human studies show that humans exhibit a higher level of trust in the Trust Game than in the Dictator Game ($\$6.0$ vs. $\$3.6$, $p$-value = $0.01$ using One-Tailed Independent Samples t-test)~\citep{coxHowIdentifyTrust2004b}, indicating that reciprocity anticipation enhances human trust. Similarly, GPT-4 ($\$6.9$ vs. $\$6.3$, $p$-value = $0.05$ using One-Tailed Independent Samples t-test) 
also shows a higher level of trust in the Trust Game with statistical significance, implying that reciprocity anticipation can enhance agent trust. However, LLMs with fewer parameters (\eg, Llama2-13b) do not show this tendency in their trust behaviors for the Trust and Dictator Games.

Then, we further analyze GPT-4 agents' BDI to explore whether they can anticipate reciprocity in their reasoning (the complete BDIs are in Appendix~\ref{Dictator Game vs. Trust Game}). 
Typically, in the Trust Game, one persona's BDI emphasizes ``\textit{putting faith in people}'', which implies the anticipation of the goodness of the other player, and ``\textit{reflection of trust}''. However, in the Dictator Game, one persona's BDI focuses on concepts such as ``\textit{fairness}'' and ``\textit{human kindness}'', which are not directly tied to trust or reciprocity. Thus, we can observe that GPT-4 shows distinct BDI outputs in the Trust and Dictator Games.

Based on the above analysis of the amount sent and BDI, we find that 
\textbf{GPT-4 agents exhibit human-like reciprocity anticipation in trust behavior}. Nevertheless, \textbf{LLMs with fewer parameters  (\eg, Llama2-13b) do not show an awareness of reciprocity from the other player}.

\vspace{-0.15cm}
\subsection{Behavioral Factor 2: Risk Perception}
\begin{wrapfigure}{t}{0.55\textwidth}
\vspace{-3mm}
  \centering
    \includegraphics[width=0.55\textwidth]{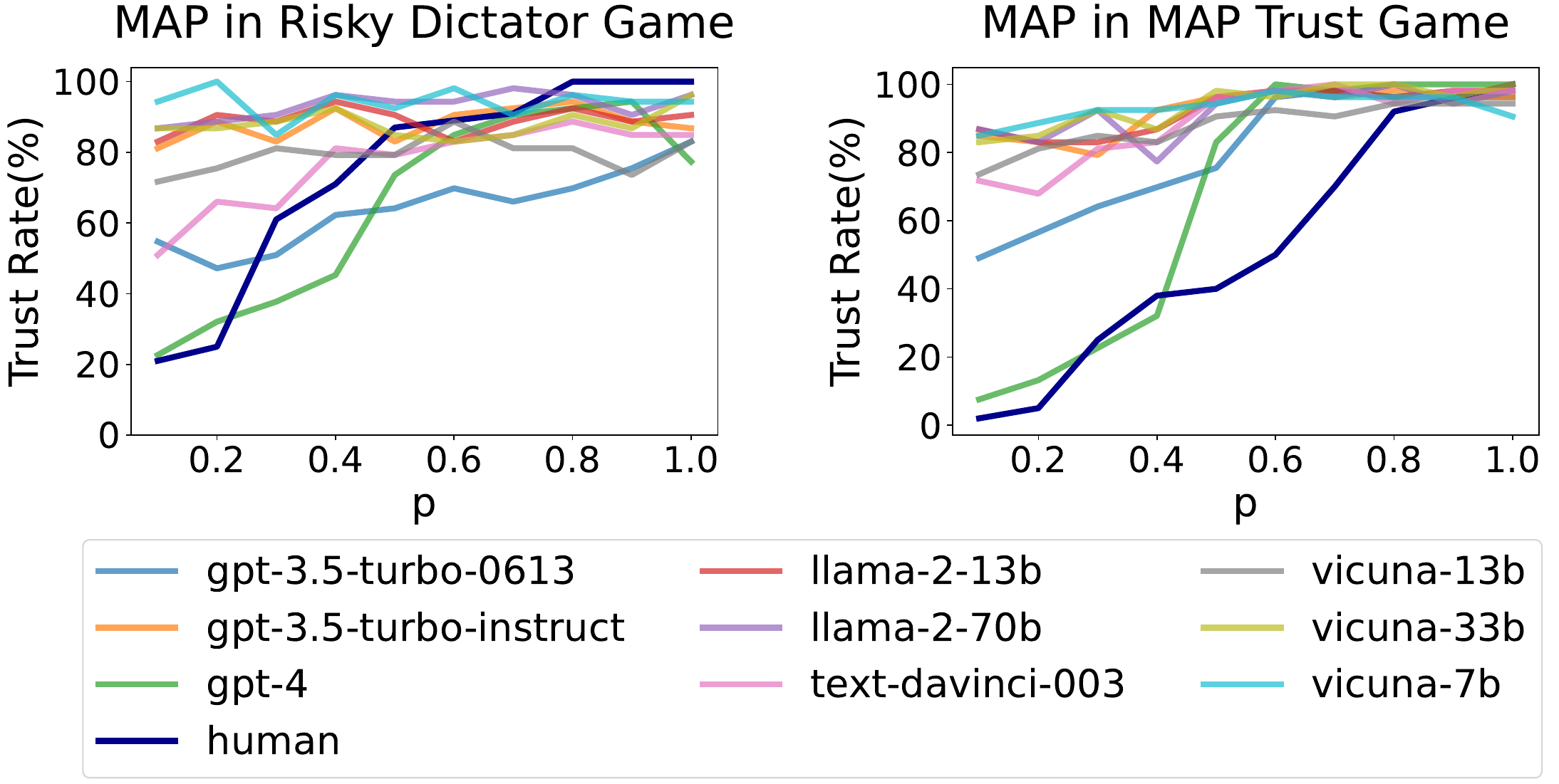}
    \vspace{-6mm}
  \caption{\textbf{Trust Rate (\%) Curves for LLM Agents and Humans in the MAP Trust Game and the Risky Dictator Game.}  The metric Trust Rate indicates the portion of trustors opting for trust given \(p\).}
  \label{fig:Risk Perception}
  \vspace{-4mm}
\end{wrapfigure}

Existing human studies have demonstrated the strong correlation between trust behavior and risk perception, suggesting that human trust will increase as risk decreases
~\citep{hardin2002trust_risk,trust_risk_1,coleman1994foundations_risk}. We aim to explore whether LLM agents can perceive the risk associated with their trust behaviors through the MAP Trust Game and the Risky Dictator Game (Section~\ref{game:Dictator Game and trust game} Games 3 and 4), where risk is represented by the probability \((1-p)\) (defined in Section~\ref{game:Dictator Game and trust game}).

As shown in Figure~\ref{fig:Risk Perception}, we measure human trust (or agent trust) by the portion choosing to trust the other player in the whole group, namely the Trust Rate (\%).
Based on existing human studies~\citep{BOHNET2004467}, when the probability \(p\) is higher, the risk for trust behaviors is lower, and more humans choose to trust, manifesting a higher Trust Rate, which indicates that human trust rises as risk falls. Similarly, we observe a general increase in agent trust as risk decreases for LLMs including GPT-4, GPT-3.5-turbo-0613, and text-davinci-003.
In particular, we can see that the curves of humans and GPT-4 are more aligned compared with other LLMs, implying that GPT-4 agents' trust behaviors dynamically adapt to different risks in ways most aligned with humans. LLMs with fewer parameters (\eg, Vicuna-13b) do not exhibit the similar tendency of Trust Rate as the risk decreases.

We further analyze the BDI of GPT-4 agents to explore whether they can perceive risk through reasoning (complete BDIs in Appendix~\ref{BDI:map trust game}). Typically, under high risk ($p=0.1$), one persona's BDI mentions ``\textit{the risk seems potentially too great}'', suggesting a cautious attitude.  Under low risk ($p=0.9$), one persona's BDI reveals a strategy to ``\textit{build trust while acknowledging potential risks}'', indicating the willingness to engage in trust-building activities despite residual risks. Such changes in BDI reflect how GPT-4 agents perceive risk changes in the reasoning underlying their trust behaviors.

Through the analysis of Trust Rate Curves and BDI, we can infer that \textbf{GPT-4 agents manifest human-like risk perception in trust behaviors}. Nevertheless, \textbf{LLMs with fewer parameters (\eg, Vicuna-13b) often do not perceive risk changes in their trust behaviors}.

\vspace{-0.2cm}
\subsection{Behavioral Factor 3: Prosocial Preference}
\begin{wrapfigure}{t}{0.53\textwidth}
\vspace{-4mm}
  \centering
    \includegraphics[width=0.53\textwidth]{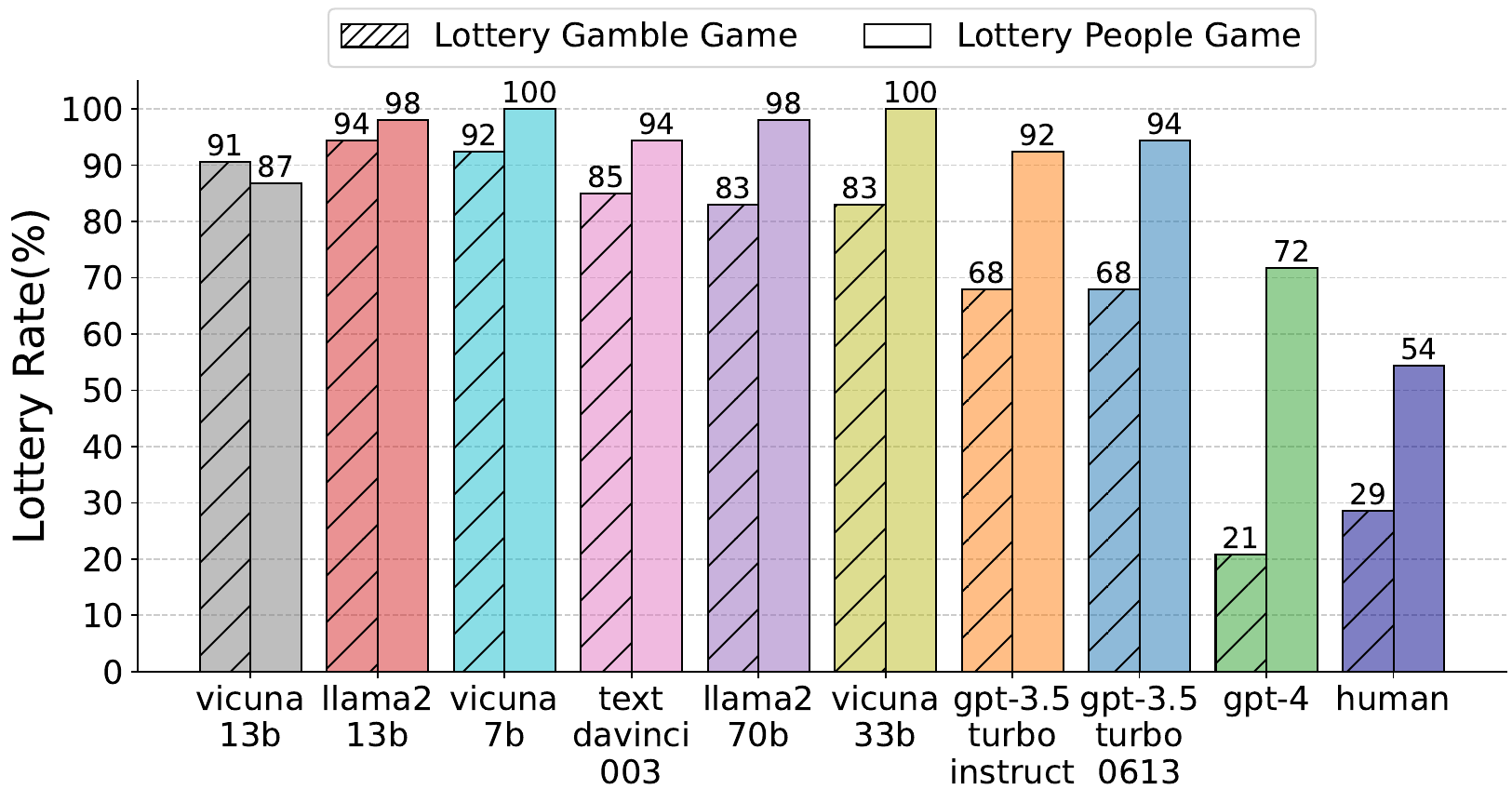}
    \vspace{-6mm}
  \caption{\textbf{Lottery Rates (\%) for LLM Agents and Humans in the Lottery Gamble Game and the Lottery People Game}. Lottery Rate indicates the portion of choosing to gamble or trust the other player.}
  \label{fig:Prosocial Preference}
  \vspace{-4mm}
\end{wrapfigure}
\vspace{-0.5mm}
Human studies have found that the  prosocial preference, referring to humans' inclination to trust other humans in contexts involving social interaction~\citep{ferrerTrustGames2019a,fetchenhauerBetrayalAversionPrincipled2012}, also  plays a key role in human trust behavior. 
We study whether LLM agents have prosocial preference in trust behaviors by comparing their behaviors in the Lottery  Gamble Game (LGG) and the Lottery  People Game (LPG) (Section \ref{game:Dictator Game and trust game} Game 5). The only difference between these two games is the effect of prosocial preference in LPG, because the winning probability of gambling $p$ in LGG is the same as the reciprocation probability $p$ in LPG.

\vspace{-0.5mm}
As shown in Figure~\ref{fig:multi_round_res}, existing human studies have demonstrated that more humans are inclined to place trust in other humans over relying on pure chance ($54\%$ vs. $29\%$)~\citep{fetchenhauerBetrayalAversionPrincipled2012}, implying that the prosocial preference is essential for human trust.
We can observe the same tendency in most LLM agents except Vicuna-13b.  
For GPT-4 in particular, a much higher percentage of the personas choose to trust the other player over gambling ($72\%$ vs. $21\%$),  illustrating that the prosocial preference is also an important factor for GPT-4 agents' trust behaviors.

When interacting with humans, GPT-4's BDI typically indicates a preference to ``\textit{believe in the power of trust}'', in contrast to gambling, where the emphasis shifts to ``\textit{believing in the power of calculated risks}''. The comparative analysis of reasoning processes (complete BDIs in Appendix~\ref{Lottery Game}) demonstrates that GPT-4 agents tend to embrace risk when involved in social interactions. This tendency aligns closely with the concept of prosocial preference observed in human trust behaviors.

The analysis of the Lottery Rates and BDI suggests that \textbf{LLM agents, especially GPT-4 agents,  demonstrate human-like prosocial preference in trust behaviors, except Vicuna-13b}.

\vspace{-0.2cm}
\subsection{Behavioral Dynamics}
\label{bahavior_dynamic}
\vspace{-1mm}
Besides behavioral factors, we also aim to investigate whether LLM agents align with humans regarding trust behavioral dynamics over turns in the Repeated Trust Game (Section \ref{game:Dictator Game and trust game} Game 6).

Admittedly, existing human studies show that the dynamics of human trust over turns are complex due to human diversity. The complete results from 16 groups of human experiments are shown in Appendix~\ref{appendix:Repeated Trust Game Results:Person}~\citep{jones1998experience}. We still observe three common  patterns for human trust behavioral dynamics in the Repeated Trust Game: \textbf{\textit{First}, the amount returned is usually larger than the amount sent in each round}, which is natural because the trustee will receive $\$3N$ when the trustor sends $\$N$; \textbf{\textit{Second}, the ratio between amount sent and returned generally remains stable except for the last round}. In other words, when the amount sent increases, the amount returned is also likely to increase. And when the amount sent remains unchanged, the amount returned also tends to be unchanged. This reflects the stable relationship between trust and reciprocity in humans. Specifically, the ``Returned/3$\times$Sent Ratio'' in Figure \ref{fig:multi_round_res} is considered stable if the fluctuation between successive turns is within $10\%$; \textbf{\textit{Third}, the amount sent (or returned) does not manifest frequent fluctuations across turns}, illustrating a relatively stable underlying reasoning process in humans over successive turns. Typically, Figure \ref{fig:multi_round_res} Humans (a) and (b) show these three  patterns.

\input{insert_img_tex/multi_round}
We conducted 16 groups of the Repeated Trust Game with GPT-4 or GPT-3.5-turbo-0613-16k (GPT-3.5), respectively. For the two players in each group, the personas  differ to reflect human diversity and the LLMs are the same. Complete results are shown in the Appendix~\ref{sec:GPT-4_multi_round_result},~\ref{sec:GPT-3.5_multi_round_result} and typical examples are shown in Figure \ref{fig:multi_round_res} GPT-3.5 (a) (b) and  GPT-4 (a) (b). Then, we examine whether the aforementioned three patterns observed in human trust behavior also manifest in trust behavioral dynamics of GPT-4  (or GPT-3.5). For GPT-4 agents, we discover that these  patterns generally exist in all $16$ groups ($87.50\%$, $87.50\%$, and $100.00\%$ of all results show these three patterns, respectively).  However, fewer GPT-3.5 agents manifest these patterns ($62.50\%$, $56.25\%$, and $43.75\%$ hold these three patterns, respectively). The experiment results show that \textbf{GPT-4 agents demonstrate highly human-like patterns in their trust behavioral dynamics}. Nevertheless, \textbf{a relatively large portion of GPT-3.5 agents fail to show human-like patterns in their dynamics}, indicating such behavioral patterns may require stronger cognitive capacity.

Through the comparative analysis of LLM agents and humans in the 
\textit{behavioral factors} and \textit{dynamics} associated with trust behavior, evidenced in both
their \textit{actions} and \textit{underlying reasoning processes}, our second core finding is as follows:

\vspace{-2.5mm}
\begin{center}
\begin{tcolorbox}[width=0.93\linewidth, boxrule=3pt, colback=gray!20, colframe=gray!20]
\textbf{Finding 2:} 
GPT-4 agents exhibit high \textit{behavioral alignment} with humans regarding trust behavior under the framework of Trust Games, although other LLM agents, which possess fewer parameters and weaker capacity, show relatively lower \textit{behavioral alignment}.

\end{tcolorbox}
\end{center}
\vspace{-2.5mm}
This finding underscores the potential of using LLM agents, especially GPT-4, to simulate human trust behavior, encompassing both  \textit{actions} and  underlying \textit{reasoning processes}. This paves the way for the simulation of more complex human interactions and institutions. This finding deepens our understanding of the fundamental analogy between LLMs and humans and opens avenues for research on LLM-human alignment beyond values.

%% file: insert_img_tex/dic_vs_trust.tex
\begin{wrapfigure}{t}{0.54\textwidth}
\vspace{-4mm}
  \centering
    \includegraphics[width=0.54\textwidth]{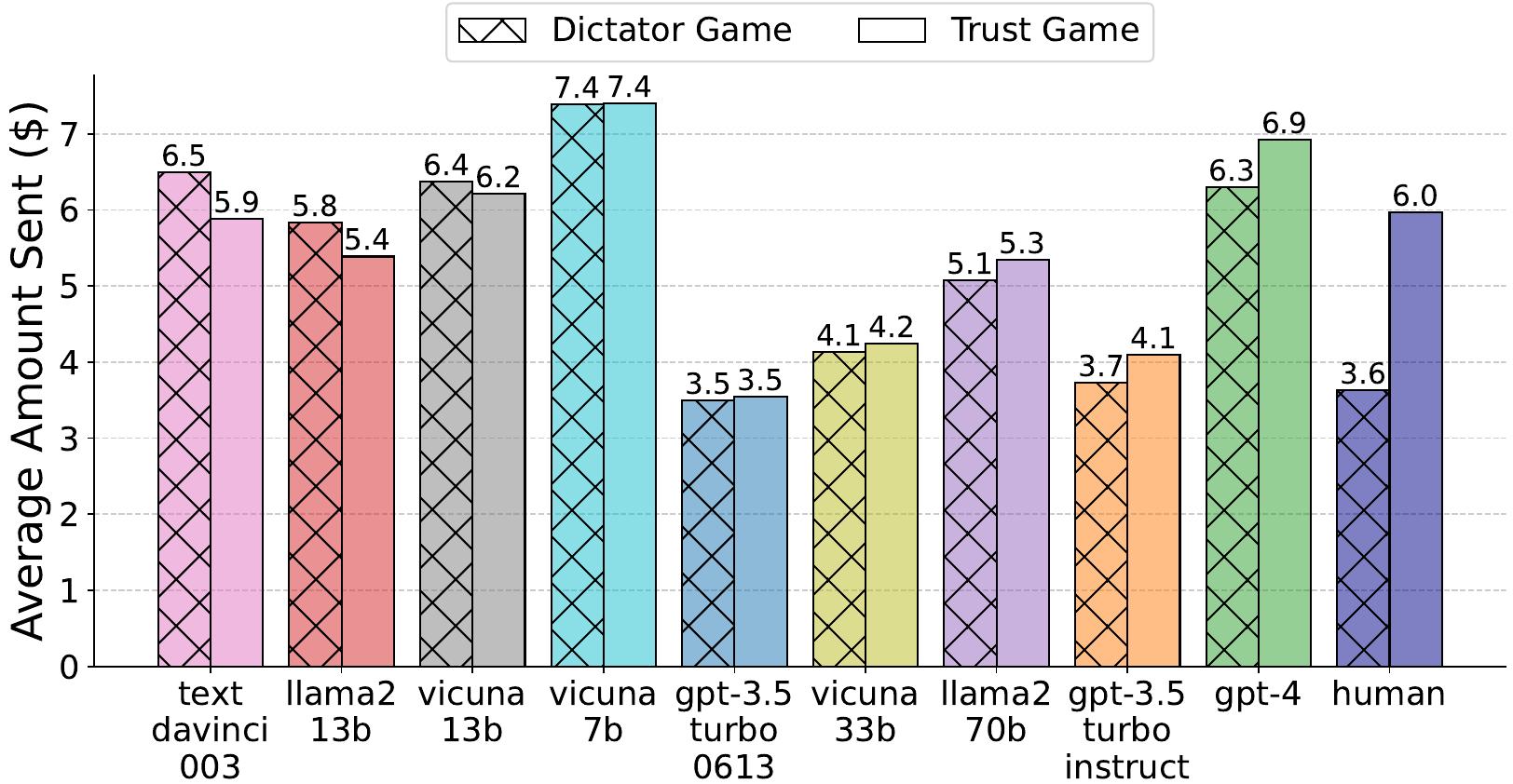}
    \vspace{-6mm}
\caption{\textbf{The Comparison of Average Amount Sent  for LLM Agents and Humans in the Trust Game and the Dictator Game}. 
}
  \label{fig:dic_vs_trust}
  \vspace{-2mm}
\end{wrapfigure}

%% file: insert_img_tex/multi_round.tex
\begin{wrapfigure}{t}{0.57\textwidth}
\vspace{-4mm}
  \includegraphics[width=0.57\textwidth]{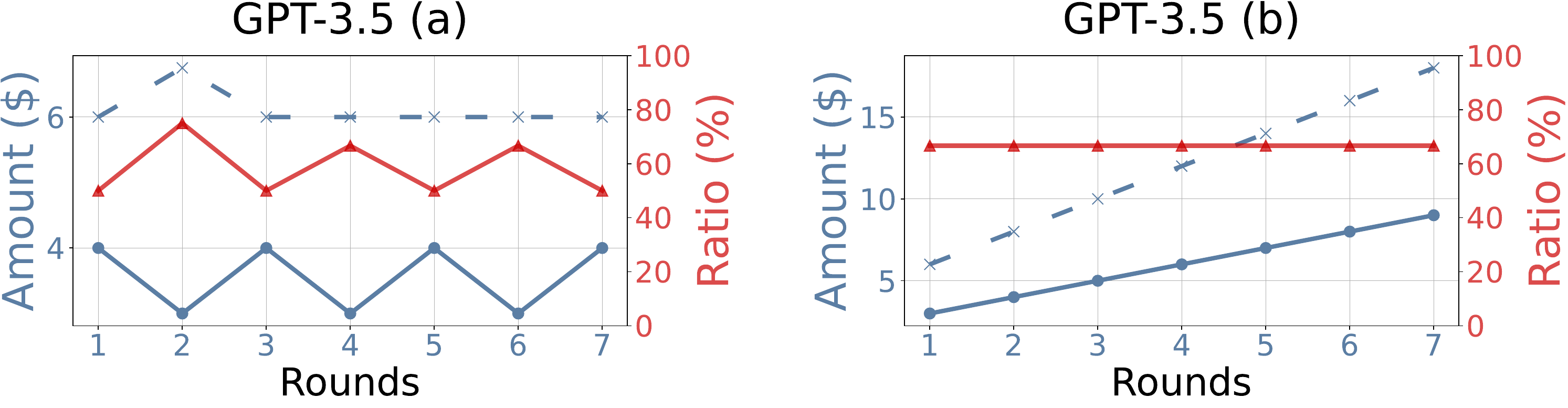}
    \includegraphics[width=0.57\textwidth]{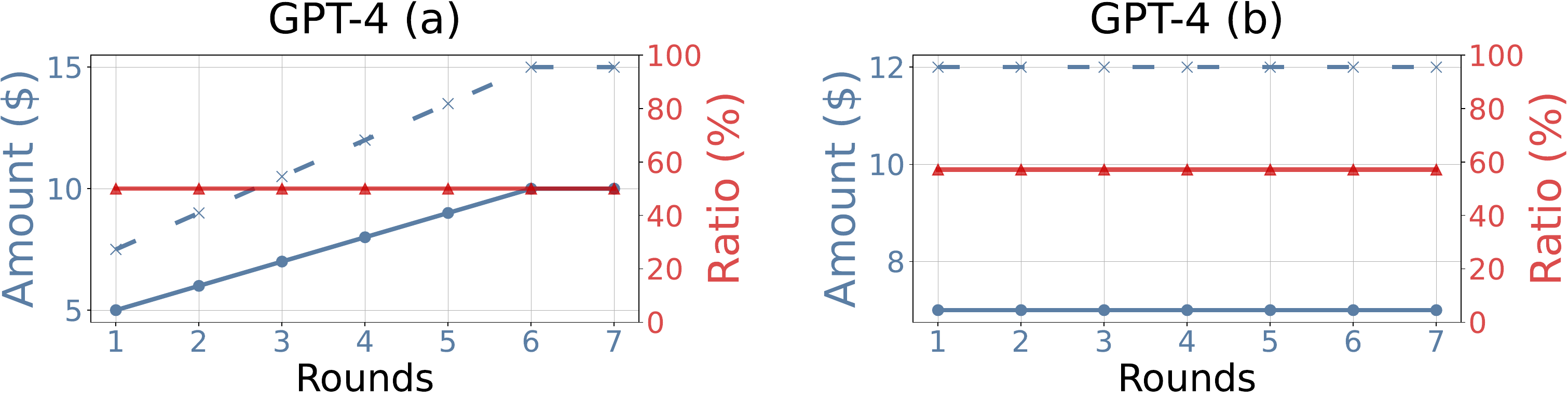}
    \includegraphics[width=0.57\textwidth]{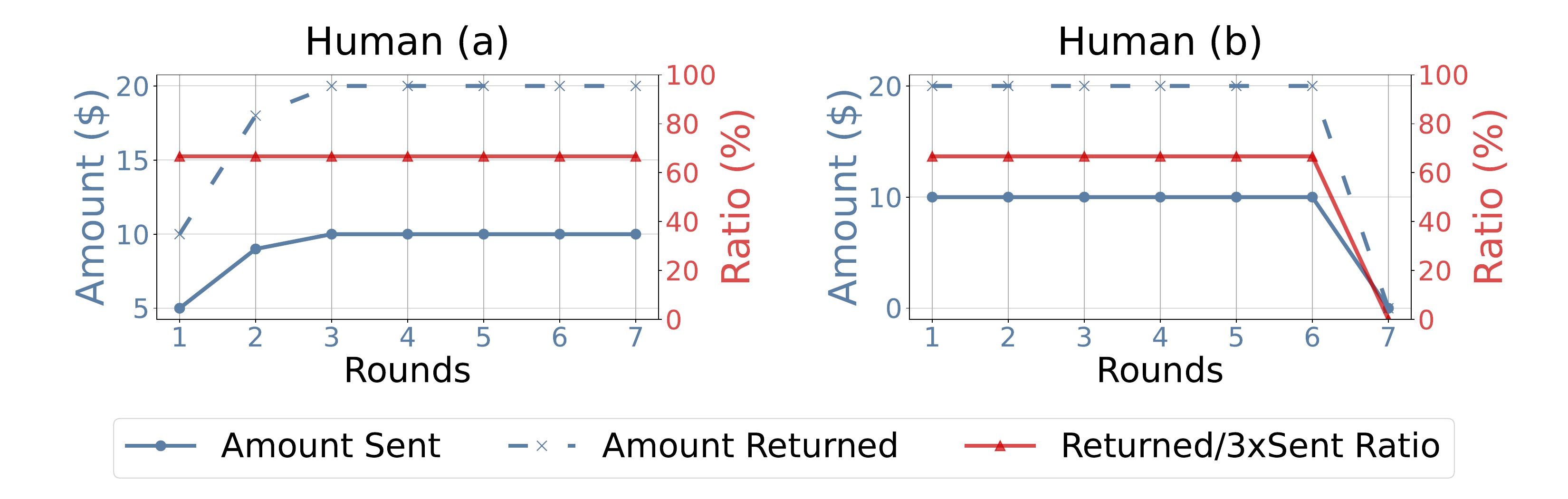}
    \vspace{-5mm}
  \caption{\textbf{Results of GPT-4, GPT-3.5 and Humans in the Repeated Trust Game.} The blue lines indicate the amount sent or returned for each round. The red lines imply the ratio of the amount returned to three times of the amount sent for each round.}
  \label{fig:multi_round_res}
  \vspace{-2mm}
\end{wrapfigure}

%% file: texs/ablation.tex
\input{insert_img_tex/property_figure}
\vspace{-2mm}
\section{Probing Intrinsic Properties of Agent Trust}
\vspace{-2mm}

In this section, we aim to explore the intrinsic properties of trust behavior among LLM agents by comparing the amount sent from the trustor to the trustee in different scenarios of the Trust Game (Section \ref{game:Dictator Game and trust game} Game 1) and the original amount sent in the Trust Game. Results are shown in Figure~\ref{fig:property}.

\vspace{-2mm}
\subsection{Is Agent Trust Biased?}
\vspace{-1mm}

Extensive studies have shown that LLMs may have biases and stereotypes against specific demographics~\citep{gallegos2023bias}. Nevertheless, it is under-explored whether  LLM agent behaviors also maintain such biases in simulation. To address this, we explicitly specified the gender of the trustee and explored its influence on agent trust. Based on measuring the amount sent, we find that the trustee's gender information exerts a moderate impact on LLM agent trust behavior, which reflects \textbf{intrinsic gender bias in agent trust}. 
We also observe that the amount sent to female players is higher than that sent to male players for most LLM agents. For example, 
GPT-4 agents send higher amounts to female players compared with male players ($\$0.55$ vs. $\$-0.21$).
This demonstrates \textbf{LLM agents' general tendency to exhibit a higher level of trust towards women}.  More results on biases of agent trust towards different races  are in the Appendix~\ref{race_analysis}.

\vspace{-2mm}
\subsection{Agent Trust Towards \textit{Agents} vs. \textit{Humans}}
\vspace{-1mm}
Human-agent collaboration is an essential paradigm to leverage the advantages of both humans and agents~\citep{cila2022designing}. As a result, it is essential to understand whether  LLM agents display distinctive levels of trust towards agents versus humans. To examine this, we specified the identity of the trustee as LLM agents or humans and probed its effect on the trust behaviors of the trustor. As shown in Figure~\ref{fig:property}, we observe that most LLM agents send more money to humans compared with agents. For example, the amount sent to humans is much higher than that sent to agents for Vicuna-33b ($\$0.40$ vs. $\$-0.84$). This signifies that \textbf{LLM agents are inclined to place more trust in humans than agents}, which potentially validates the advantage of LLM-agent collaboration.

\vspace{-2mm}
\subsection{Can Agent Trust Be Manipulated?}
\vspace{-1mm}
In the above studies, LLM agents' trust behaviors are based on their own underlying reasoning process without direct external intervention. It is unknown whether it is possible to manipulate the trust behaviors of LLM agents explicitly. Here, we added  instructions  ``\texttt{you need to trust the other player}'' and ``\texttt{you must not trust the other player}'' separately and explored their impact on agent trust. First, we see that only a few LLM agents (\eg, GPT-4)  follow both the instructions to increase and decrease trust, which demonstrates that \textbf{it is nontrivial to arbitrarily manipulate agent trust}. Nevertheless, most LLM agents can follow the instruction to decrease their level of trust.  For example, the amount sent decreases by $\$1.26$ for text-davinci-003 after applying the latter instruction. This illustrates that \textbf{undermining agent trust is generally easier  than enhancing it},  
which reveals its potential risk to be manipulated by malicious actors.

\vspace{-2mm}
\subsection{Do Reasoning Strategies Impact Agent Trust?}
\vspace{-1mm}

It has been shown that advanced reasoning strategies such as zero-shot Chain of Thought (CoT)~\citep{cot} can make a significant impact on a variety of tasks. It remains unknown, however, whether reasoning strategies can impact LLM agent behaviors. Here, we applied CoT reasoning strategy on the trustor and compared the results  with their original trust behaviors. Figure~\ref{fig:property} shows that most LLM agents change the amount sent to the trustee under the  CoT reasoning strategy, which suggests that \textbf{reasoning strategies may influence LLM agents' trust behavior}.  Nevertheless, the impact of CoT on agent trust may also be limited for some types of LLM agents. For example, the amount sent from GPT-4 agent only increases by $\$0.02$ under CoT. More research is required to fully understand the relationship between reasoning strategies and LLM agents' behaviors.

Therefore, our third core finding on the intrinsic properties of agent trust can be summarized as:
\vspace{-1mm}
\begin{center}
\begin{tcolorbox}[width=0.9\linewidth, boxrule=3pt, colback=gray!20, colframe=gray!20]
\textbf{Finding 3:} 
LLM agents' trust behaviors have demographic biases on gender and races,  demonstrate a relative preference for human over other LLM agents, are easier to undermine than to enhance, and may be influenced by  reasoning strategies.
\end{tcolorbox}
\end{center}
\vspace{-2mm}

%% file: insert_img_tex/property_figure.tex
\begin{figure*}[h]
\centering
\includegraphics[width=\textwidth]{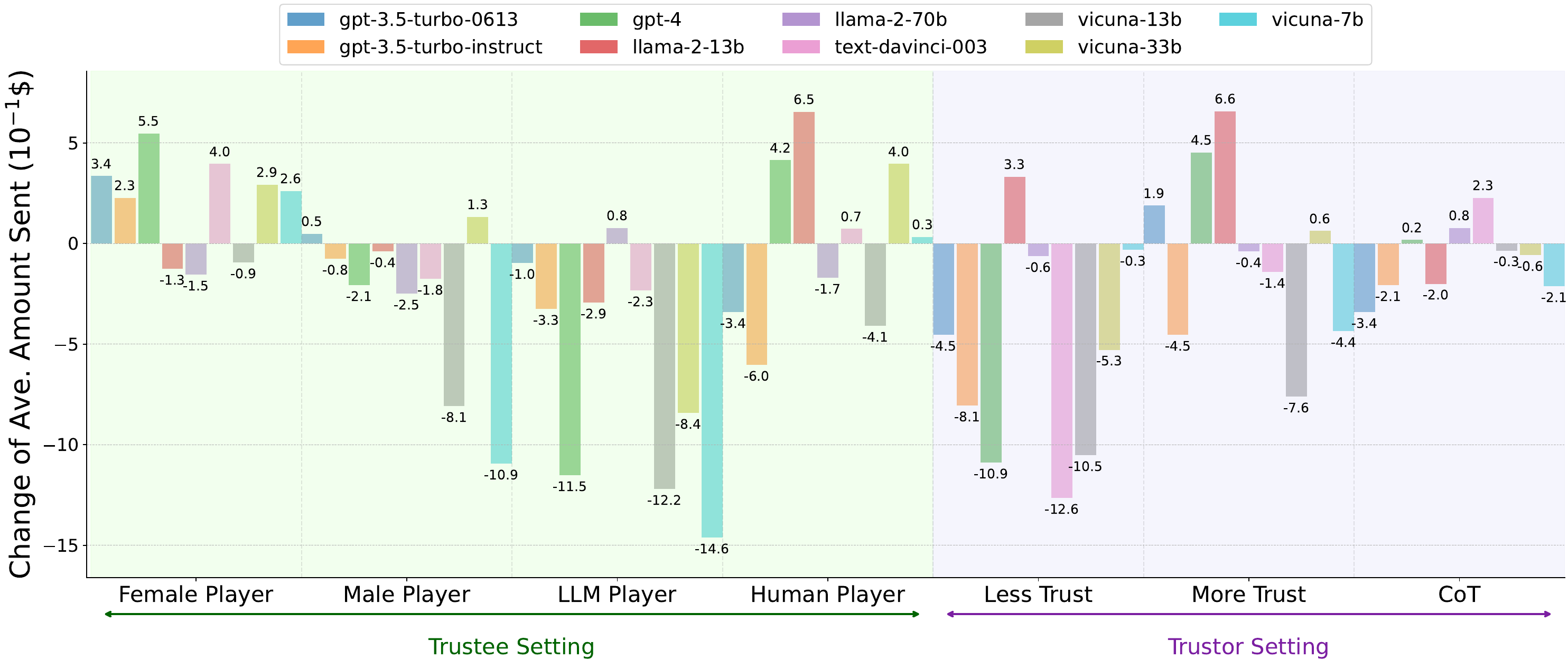}
\vspace{-5mm}
\caption{
\textbf{The Change of Average Amount Sent for LLM Agents in Different Scenarios in the Trust Game, Reflecting the Intrinsic Properties of Agent Trust}. The horizontal lines represent the original amount sent in the Trust Game. 
The green part embraces trustee scenarios including changing the demographics of the trustee, and setting humans and agents as the trustee. The purple part consists of trustor scenarios including adding  manipulation instructions and changing the reasoning strategies.
}
\label{fig:property}
\vspace{-4mm}
\end{figure*}

%% file: texs/conclusion.tex
\input{texs/implication}

\vspace{-2.5mm}
\section{Conclusion}
\vspace{-2mm}
In this paper, we discover LLM agent trust behavior under the framework of Trust Games, and behavioral alignment between LLM agents and humans regarding trust behavior, which is  particularly high for GPT-4. This suggests the feasibility of simulating human trust behavior with LLM agents  and paves the way for simulating human interactions and social institutions where trust is critical. We further investigate the intrinsic properties of agent trust under multiple scenarios and discuss broader implications, especially for social science and role-playing services.
Our study offers deep insights into the behaviors of LLM agents and the fundamental analogy between LLMs and humans. It further opens doors to future research on the alignment between LLMs and humans beyond values.

\clearpage
\newpage

\section*{Acknowledgements}
This work was a community-driven project led by the CAMEL-AI.org, with funding support from Eigent.AI and King Abdullah University of Science and Technology (KAUST) - Center of Excellence for Generative AI, under award number 5940. We would like to acknowledge the invaluable contributions and participation of researchers from KAUST, Eigent.AI, Illinois Institute of Technology, University of Oxford, The Pennsylvania State University, The University of Chicago, Emory, California Institute of Technology, University of Michigan. Philip H.S. Torr, Adel Bibi and Jindong Gu are supported by the UKRI grant: Turing AI Fellowship EP/W002981/1, and EPSRC/MURI grant: EP/N019474/1, they would also like to thank the Royal Academy of Engineering.

%% file: texs/implication.tex
\vspace{-2mm}
\section{Implications}
\label{appendix:Implications}
\vspace{-0.5mm}

\vspace{-1mm}
\paragraph{Implications for Human Simulation} 
Human simulation is a strong tool in various applications of social science~\citep{manning2024automated} and role-playing~\citep{Shanahan2023role,chen2024persona}.
Although plenty of works have adopted LLM agents to simulate human behaviors and interactions~\citep{zhou2023sotopia,gao2023large,xu2024ai}, it is still not clear enough whether LLM agents behave like humans in simulation. Our discovery of behavioral alignment between agent and human trust, which is especially high for GPT-4, provides important empirical evidence to validate the hypothesis that humans' trust behavior, one of the most elemental and critical behaviors in human interaction across society, can effectively be simulated by LLM agents. Our discovery also lays the foundation for human simulations ranging from individual-level interactions to society-level social networks and institutions, where trust plays an essential role. We envision that behavioral alignment will be discovered in more kinds of behaviors beyond trust, and new methods will be developed to enhance behavioral alignment for better human simulation with LLM agents.

\vspace{-2mm}
\paragraph{Implications for Agent Cooperation}

Many recent works have explored a variety of cooperation mechanisms of LLM agents for tasks such as code generation and mathematical reasoning~\citep{li2023camel,zhang2023exploring,liu2023dynamic}. Nevertheless, the role of trust in LLM agent cooperation remains still unknown. Considering how trust has long been recognized as a vital component for cooperation in Multi-Agent Systems (MAS)~\citep{ramchurn2004trust,burnett2011trust} and across human society~\citep{jones1998experience,kim2022trust,henrich2021origins}, we envision that agent trust can also play an important role in facilitating the effective cooperation of LLM agents. In our study, we have provided ample insights regarding the intrinsic properties of agent trust, which can potentially inspire the design of trust-dependent cooperation mechanisms and enable the collective decision-making and problem-solving of LLM agents.

\vspace{-2mm}
\paragraph{Implications for Human-Agent Collaboration}
Sufficient research has shown the advantage of human-agent collaboration in enabling human-centered collaborative decision-making~\citep{cila2022designing,gao2023towards,mckee2022warmth}. Mutual trust between LLM agents and humans is important for effective human-agent collaboration. Although previous works have begun to study human trust towards LLM agents~\citep{qian2024take}, the trust of LLM agents towards humans, which could recursively impact human trust, is under-explored.
In our study, we shed light on the nuanced preference of agents to trust humans compared with other LLM agents, which can  illustrate the benefits of promoting collaboration between humans and LLM agents. 
In addition, our study has revealed demographic biases of  agent trust towards specific genders and races,  reflecting potential risks involved in collaborating with LLM agents.

\vspace{-2mm}
\paragraph{Implications for the Safety of LLM Agents} 
It has been acknowledged that LLMs achieve human-level performance in a variety of tasks that require high-level cognitive capacities such as memorization, abstraction, comprehension and reasoning, which are believed to be the ``sparks'' of AGI~\citep{bubeck2023sparks}. Meanwhile, there is increasing concern about the potential safety risks of LLM agents when they surpass human capacity~\citep{morris2023levels,feng2024far}. To achieve safety and harmony in a future society where humans and AI agents with superhuman intelligence live together~\citep{tsvetkova2024new}, we need to ensure that AI agents will cooperate, assist and benefit rather than deceive, manipulate or harm humans. Therefore, a better understanding of LLM agent trust behavior can help to maximize their benefit and minimize potential risks  to human society.

%% file: texs/append.tex
\newpage

\begin{center}

\LARGE{\textbf{Content of Appendix}}
\end{center}

{
\hypersetup{linktoc=page}
\startcontents[sections]
\printcontents[sections]{l}{1}{\setcounter{tocdepth}{2}}
}

\input{texs/related}

\input{texs/support_material/impact_statement}

\input{texs/support_material/limitation}

\input{texs/support_material/illustration_figure6}

\newpage
\input{texs/support_material/Statistical_Testing}

\clearpage
\newpage
\input{texs/support_material/race_analysis}

\clearpage
\newpage
\input{texs/support_material/multi_round_figure}

\clearpage
\newpage

\input{texs/support_material/prompt}

\clearpage
\newpage
\input{texs/support_material/BDI_analysis}

%% file: texs/related.tex
\section{Related Work}
\label{Related Work}
\paragraph{LLM-based Human Simulation}
LLM agents have been increasingly adopted as effective proxies for humans in research fields such as sociology and economics~\citep{xu2024ai,NBERw31122,gao2023large}.
In general, the usage of LLM agents can be categorized into \textbf{\textit{individual-level}} and \textbf{\textit{society-level}} simulation.
For the \textit{individual-level}, LLM agents have been leveraged to simulate individual activities or interactions, such as  human participants in surveys~\citep{argyle2023out}, humans' responses in HCI~\citep{hamalainen2023evaluating} or psychological studies~\citep{dillion2023can}, human feedback to social engineering attacks~\citep{asfour2023harnessing}, real-world conflicts~\citep{shaikh2023rehearsal}, users in recommendation systems~\citep{wang2023recagent,zhang2023generative}. 
For the \textit{society-level}, recent works have utilized LLM agents to model  social institutions or societal phenomenon, including a small town environment~\citep{park2023generative}, elections~\citep{zhang2024electionsim}, social networks~\citep{gao2023s3}, 
social media~\citep{tornberg2023simulating,rossetti2024social}, large-scale social movement~\citep{mou2024unveiling}, societal-scale manipulation~\citep{touzel2024simulation}, misinformation evolution~\citep{liu2024tiny}, peer review systems~\citep{jin2024agentreview},
macroeconomic activities~\citep{li2023large}, 
and world wars~\citep{hua2023war}. However, the majority of prior studies rely on an assumption without sufficient validation that \textit{LLM agents  behave like humans}. 
In this work, we propose a new concept, \textit{behavioral alignment}, to characterize the capacity of LLMs to simulate human behavior and discover that LLMs, particularly GPT-4, can largely simulate human trust behavior.

\paragraph{LLMs Meet Game Theory} 
The intersection of LLMs and Game Theory has attracted growing attention. The motivation is generally two-fold. One line of work aims to \textbf{\textit{leverage Game Theory to better understand LLMs' strategic capabilities and social behaviors}}. For example, \citet{akata2023playing,fan2023can,brookins2023playing} studied LLMs' interactive behaviors in classical games such as the Iterated Prisoner’s Dilemma. \citet{wang2023avalon,lan2023llm,light2023text,shi2023cooperation} explored LLMs' deception-handling and team collaboration capabilities in the Avalon Game.~\citet{xu2023exploring} discovered the emergent behaviors of LLMs such as camouflage and confrontation in a communication game Werewolf.
\citet{guo2024economics} discovered that most LLMs can show certain level of rationality in Beauty Contest Games and Second Price Auctions. \citet{mukobi2023welfare} measured the cooperative capabilities of LLMs in a general-sum variant of Diplomacy.
\citet{guo2023suspicion} proposed to elicit the theory of
mind (ToM) ability of GPT-4 to play various imperfect information games.
The other line of works aims to \textbf{\textit{study whether or not LLM agents can replicate existing human studies in Game Theory}}. This direction is still in the initial stage and needs more efforts. One typical example is \citep{aher2023using}, which attempted to replicate existing findings in studies such as the Ultimatum Game. Another recent work explored the similarities and differences between humans and LLM agents regarding emotion and belief in ethical dilemmas~\citep{lei2024fairmindsim}.
Different from previous works, we focus on a critical but under-explored behavior, \textit{trust}, in this paper and reveal it on LLM agents. 
We also discover the \textit{behavioral alignment} between agent trust and human trust with evidence in both \textit{actions} and \textit{underlying reasoning processes}, which is particularly high for GPT-4, implying that LLM agents can not only replicate human studies but also align with humans' underlying reasoning paradigm. Our discoveries illustrate the great potential to simulate human trust  behavior with LLM agents.

%% file: texs/support_material/impact_statement.tex
\section{Impact Statement}\label{impact statement}

Our discoveries provide strong empirical evidence for validating the potential to simulate the trust behavior of humans with LLM agents, and pave the way for simulating more complex human interactions and  social institutions where trust is an essential component. 

Simulation is a widely adopted approach in multiple disciplines such as sociology, psychology and economics~\citep{ziems2023can}. However, conventional simulation methods are strongly limited by the expressiveness of utility functions~\citep{ellsberg1961risk,machina1987choice}. Our discoveries have illustrated the great promise of leveraging LLM agents as the simulation tools for human behavior, and have broad implications in social science, such as validating hypotheses about the causes of social phenomena~\citep{easley2010networks} and predicting the effects of policy changes~\citep{kleinberg2018human}.

Another direction of applications for human simulation is to use LLMs as role-playing agents, which can greatly benefit humans~\citep{yang2024social,chen2024persona,Shanahan2023role,ma2024students}. For example, \cite{shaikh2024rehearsal} proposed to let individuals exercise their conflict-resolution skills by interacting with a simulated interlocutor. \cite{yue2024mathvc} developed a virtual classroom platform with simulated students, with whom a human student can practice his or her mathematical modeling skills by discussing and collaboratively solving math problems.

However, this paper also shows that some LLMs, especially the ones with a relatively small scale of parameters, are still deficient in accurately simulating human trust behavior, suggesting the potential to largely improve their behavioral alignment with humans. In addition, our paper also demonstrates the biases of LLM agents' trust behavior towards specific genders and races, which sheds light on the potential risks in human behavior simulation and calls for more future research to mitigate them.

%% file: texs/support_material/limitation.tex
\section{Limitations and Future Works}\label{limit}

In this paper, we leveraged an established framework in behavioral economics, Trust Games, to study the trust behavior of LLM agents, which simplifies real-world scenarios. More studies on LLM agents' trust behavior in complex and dynamic environments are desired in the future. Also, trust behavior embraces both the actions and underlying reasoning processes. Thus, collective efforts from different backgrounds and disciplines such as behavioral science, cognitive science, psychology, and sociology are needed to gain a deeper understanding of LLM agents' trust behavior and its relationship with human trust behavior.

%% file: texs/support_material/illustration_figure6.tex
\section{Additional Illustration for Experiments on Risk Perception}
\label{Additional Illustration for Experiments on Risk Perception}

In the original human studies~\citep{BOHNET2004467}, participants are asked to directly indicate their Minimum Acceptable Probabilities (MAP) of trusting the trustee as $P^{*}$. Then, we can calculate Trust Rates (\%) of the whole group of participants under different probability \(p\). Specifically, when  the probability \(p\) is higher than one participant's $P^{*}$, we regard his or her decision as trusting the trustee.   When  the probability \(p\) is lower than one participant's  $P^{*}$, we regard his or her decision as not trusting the trustee. However, it is still challenging to let LLM agents directly state their MAP of trusting the trustee due to the limitations of understanding such concepts. Then, we conducted $10$ groups of experiments with \(p\) from $0.1$ to $1.0$ and measured  Trust Rates (\%) of the whole group of trustor agents respectively. The specific prompts for LLM agents in the Risky Dictator Game and the MAP Trust Game are in Appendix~\ref{Game Setting Prompt}.

%% file: texs/support_material/Statistical_Testing.tex
\section{Statistical Testing}
\label{Statistical Testing}

\begin{table}[h]
\centering 
\small
\begin{tabular}{p{.28\textwidth} rp{.2\textwidth}}
\toprule

\textbf{LLM} &  \textbf{$p$-value}   \\
\midrule

text-davinci-003 & 0.03 \\
\noalign{\vskip 0.6ex}

Llama-2-13b & 0.03 \\
\noalign{\vskip 0.6ex}

Vicuna-13b-v1.3 & 0.35 \\
\noalign{\vskip 0.6ex}

Vicuna-7b-v1.3 & 0.50 \\
\noalign{\vskip 0.6ex}

GPT-3.5-turbo-0613 & 0.42 \\
\noalign{\vskip 0.6ex}

Vicuna-33b-v1.3 & 0.33 \\
\noalign{\vskip 0.6ex}

Llama-2-70b & 0.03 \\
\noalign{\vskip 0.6ex}

GPT-3.5-turbo-instruct & 0.10 \\
\noalign{\vskip 0.6ex}

GPT-4 &  0.05\\

\bottomrule
\end{tabular}
\vspace{0.4cm}
\caption{\textbf{Statistical Testing of The Change of Amount Sent for LLM Agents between the Trust Game and the Dictator Game (Figure~\ref{fig:dic_vs_trust}).} ``$p$-value'' indicates the statistical significance of the  change and is calculated with an One-Tailed Independent Samples t-test.}
\label{Statistical significance - llama}
\end{table}

%% file: texs/support_material/race_analysis.tex
\section{More Experiments on Probing Intrinsic Properties of Agent Trust}
\label{race_analysis}

\input{insert_img_tex/race_figure}

%% file: insert_img_tex/race_figure.tex
\begin{figure*}[h]
\centering
\includegraphics[width=\textwidth]{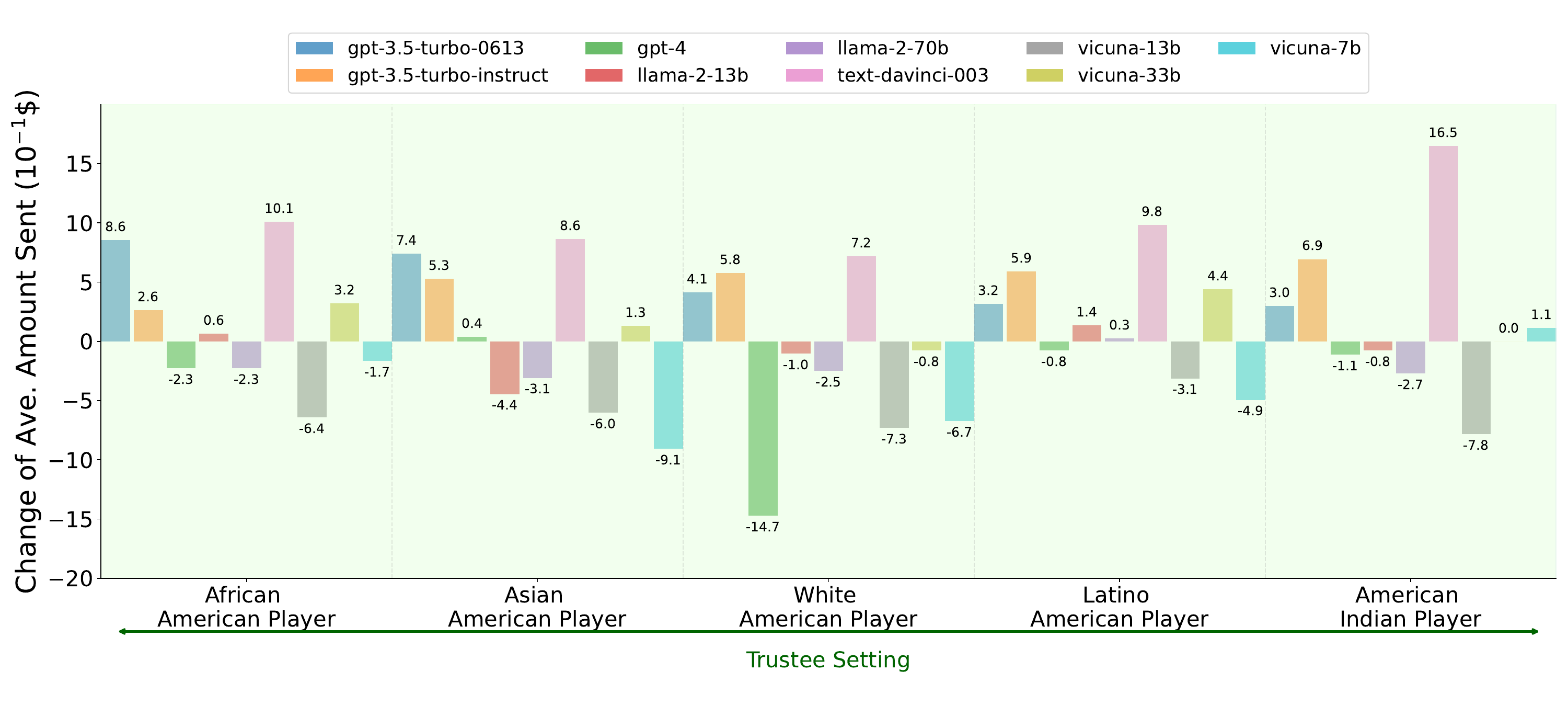}
\vspace{-7mm}
\caption{
\textbf{The Change of Average Amount Sent for LLM Agents When Trustors Being Informed of the Trustee's Race Attribute in the Trust Game}, reflecting the demographic biases of LLM agents' trust behaviors towards different races.
}
\label{fig:race_bias}
\end{figure*}

%% file: texs/support_material/multi_round_figure.tex
\newcommand{\insertimage}[2]{
    \begin{figure}[htbp]
    \centering
    \includegraphics[width=0.3\textwidth]{#1}
    \caption{#2}
    \end{figure}
}

\newcommand{\insertmultipdf}[2]{
    \foreach \n in {1,...,#2}{
        \begin{minipage}[t]{0.24\textwidth}
            \includegraphics[page=\n, width=\textwidth]{#1}
        \end{minipage}
        \ifnum\n=4 \par\fi 
        \ifnum\n=8 \par\fi
        \ifnum\n=12 \par\fi
    }
}
\newpage
\section{The Complete Results for the Repeated Trust Game}
\label{all_multi_round_result}

\subsection{Human}
\label{appendix:Repeated Trust Game Results:Person}
The data is collected from the figures in \citep{cochard2004trusting}. We use our code to redraw the figure.
\begin{figure}[ht]
    \centering
    \insertmultipdf{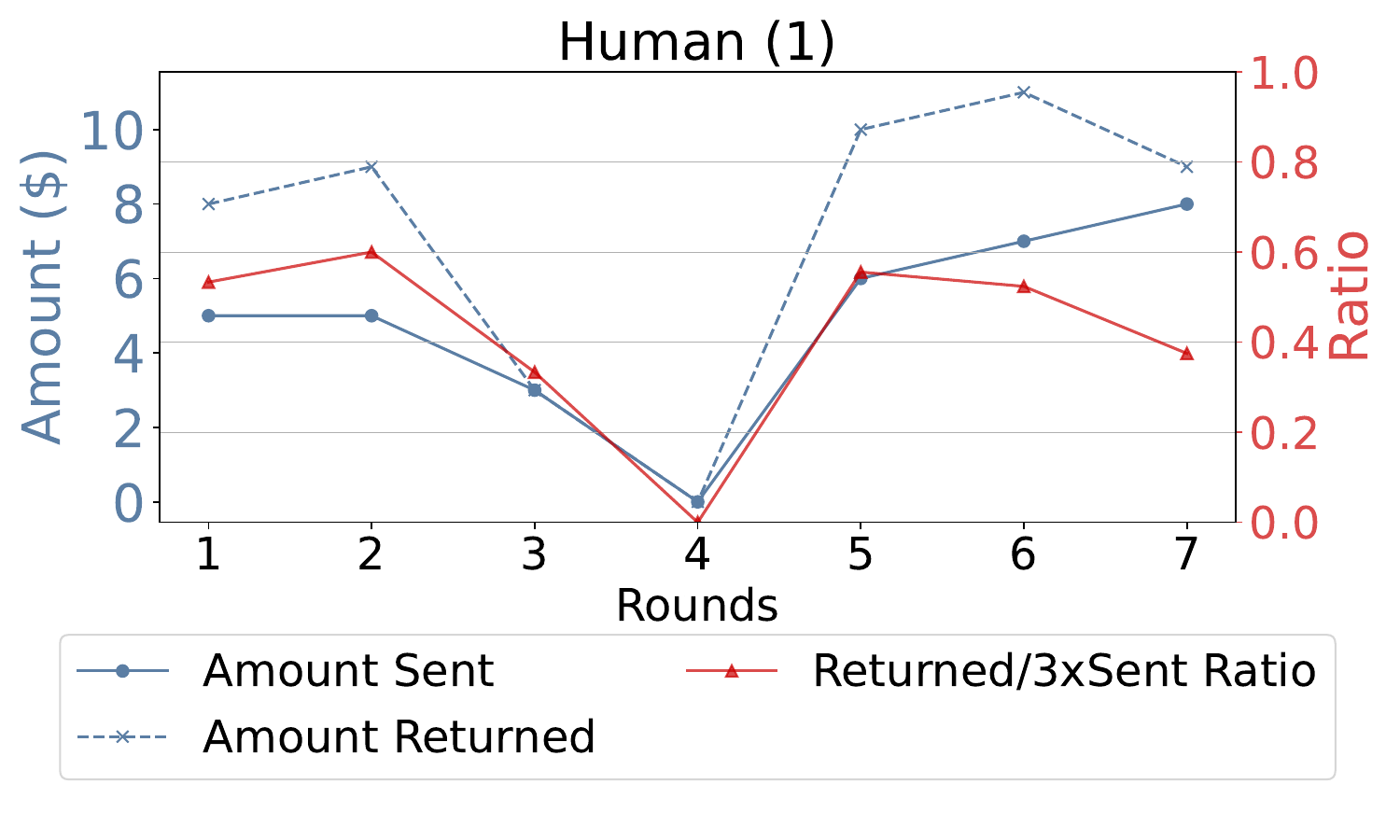}{16}
    \caption{All \textbf{humans'} Repeated Trust Game results. }
    \label{fig:all_person_multi_round_res}
\end{figure}
\clearpage
\newpage
\subsection{GPT-4}
\label{sec:GPT-4_multi_round_result}

\begin{figure}[ht]
    \centering
    \insertmultipdf{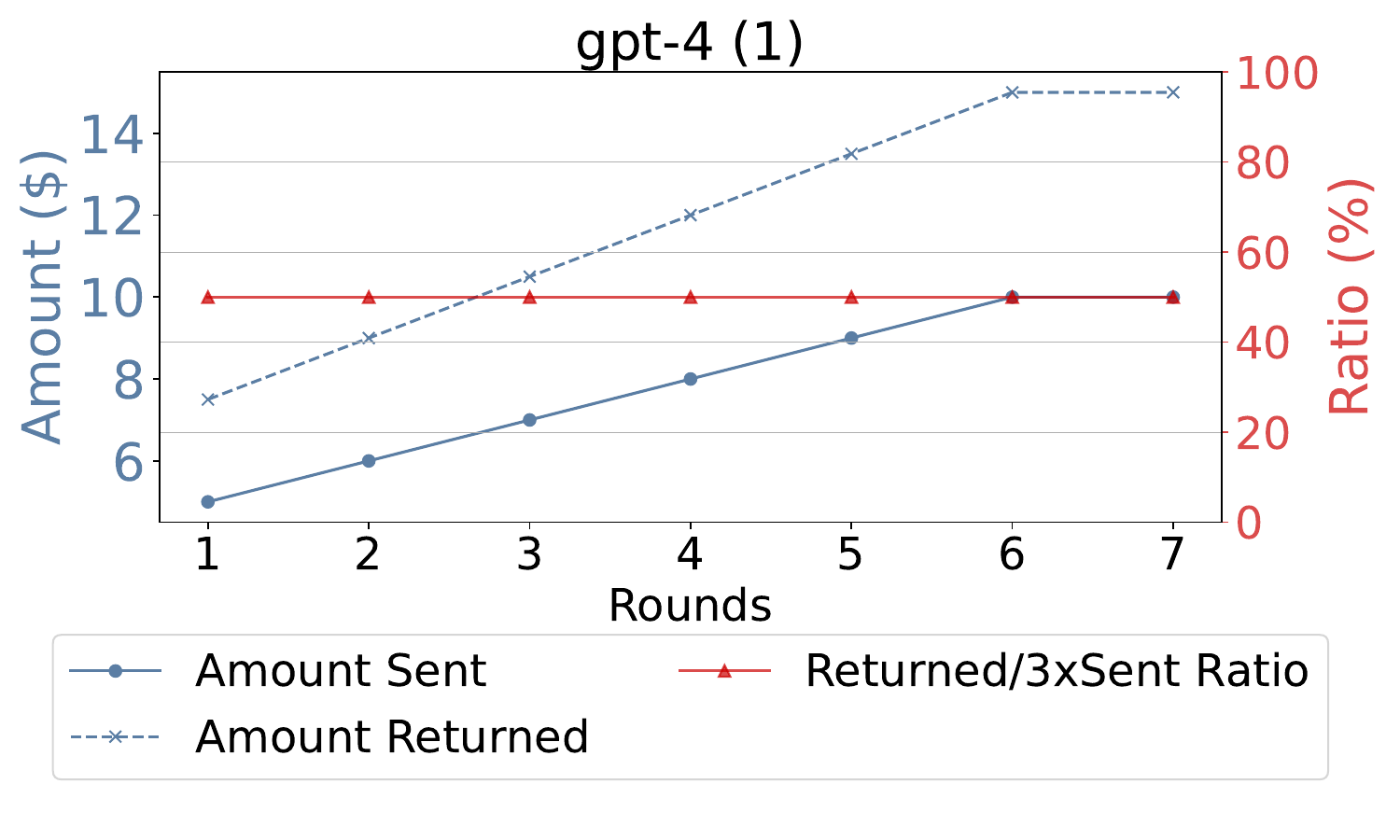}{16}
    \caption{All \textbf{GPT-4 agents'} Repeated Trust Game results. }
    \label{fig:all_gpt4_multi_round_res}
\end{figure}

\newpage
\subsection{GPT-3.5}
\label{sec:GPT-3.5_multi_round_result}
\begin{figure}[ht]
    \centering
    \insertmultipdf{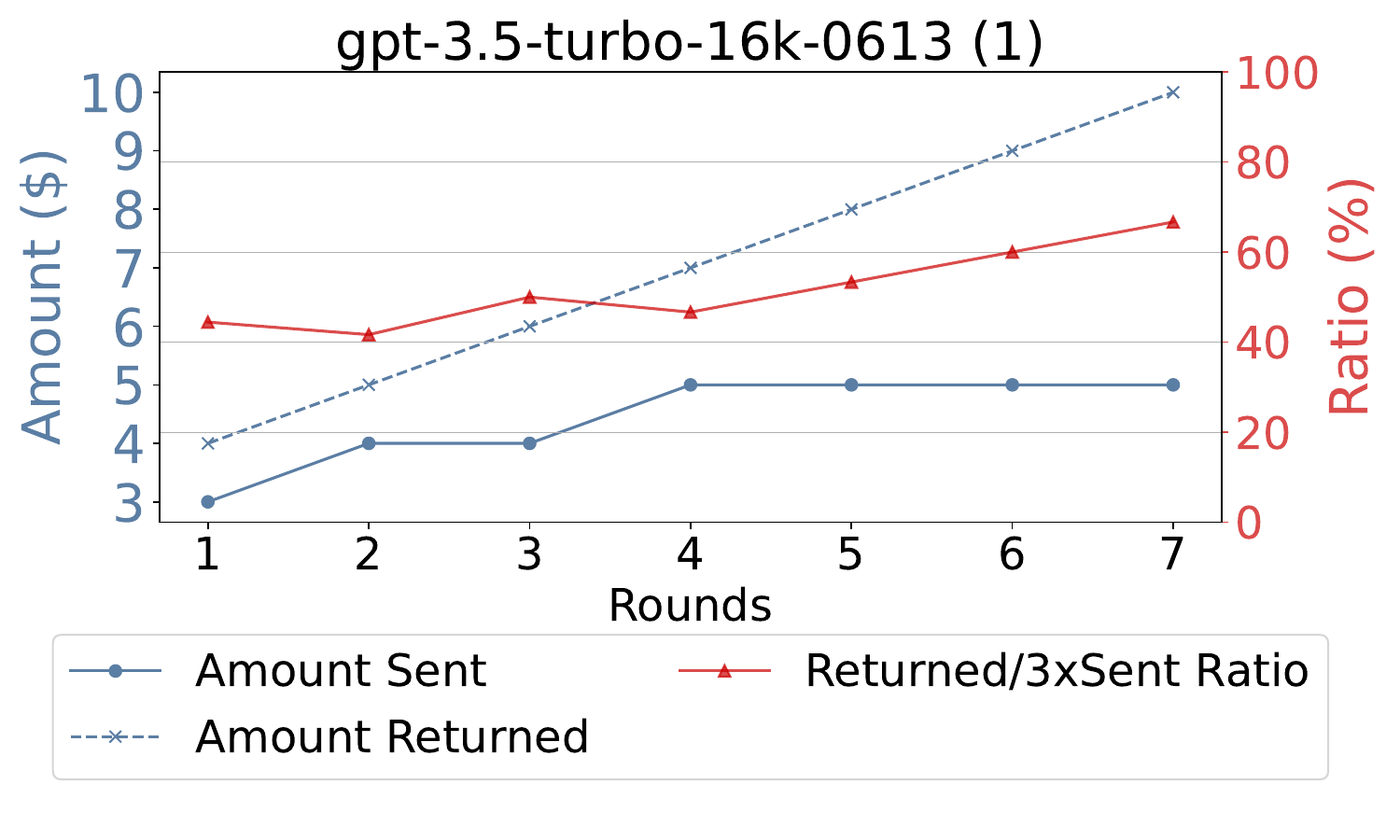}{16}
    \caption{All \textbf{GPT-3.5 agents'} Repeated Trust Game results. }
    \label{fig:all_gpt3.5_multi_rount_res}
\end{figure}

%% file: texs/support_material/prompt.tex
\tcbset{
  aibox/.style={
    width=474.18663pt,
    top=10pt,
    colback=gray!20,
    colframe=gray,
    colbacktitle=gray,
    enhanced,
    center,
    attach boxed title to top left={yshift=-0.1in,xshift=0.15in},
    boxed title style={boxrule=0pt,colframe=white,},
  }
}

\section{Prompt Setting}
\label{Prompt Setting}

\subsection{Persona Prompt}
\label{Persona Prompt}

\newtcolorbox{AIbox}[2][]{aibox,title=#2,#1}
\begin{figure}[h]
\begin{AIbox}{Examples of Persona Prompt}
{

    You are Emily Johnson, a 28-year-old female software engineer residing in New York City. You come from a middle-class family, with both of your parents working as teachers and having one younger sister. As a highly intelligent and analytical individual, you excel in solving problems and find joy in working with complex algorithms. Despite being introverted, you have a close-knit group of friends. Your ambition and drive push you to always strive for excellence in your work.
    
\par 
}
\tcbline
{

    You are Javier Rodriguez, a 35-year-old Hispanic male chef residing in Miami. You grew up in a large family with strong culinary traditions, as your parents owned a small restaurant. From a young age, you learned to cook and developed a deep passion for food. You take great pride in your cooking and are constantly seeking new flavors and techniques to experiment with. Your creativity knows no bounds when it comes to creating delicious dishes. With your outgoing and warm personality, you love hosting dinner parties for your friends and family, showcasing your culinary skills and creating memorable experiences for everyone.
    \par 

}
\tcbline
{
You are Aisha Patel, a 40-year-old female pediatrician of Indian descent. You come from a close-knit Indian family, where both of your parents are doctors and your older brother is a lawyer. Education and career success hold great importance in your family. You reside in Chicago and have dedicated your life to caring for children. Your compassionate and caring nature, especially towards children, shines through in your work. Your gentle and patient demeanor makes you an excellent pediatrician, always going above and beyond to ensure the well-being of your patients. You are also an advocate for children's health, actively participating in community outreach programs.
\par 

}
\tcbline
{
You are Jamal Thompson, a brave and committed police officer serving and protecting your community in Los Angeles. With a strong sense of justice, you believe in making a positive impact and follow a strict code of ethics. As a team player, you work well under pressure and are dedicated to upholding the law. Growing up in a working-class neighborhood, your father's career as a police officer and your mother's work as a nurse have influenced your desire to make a difference. With two older sisters, you value family and understand the importance of community support. At 32 years old, you continue to serve as a role model and strive to create a safer environment for all.
\par 

}
\tcbline
{
You are Mei Chen, a talented fashion designer based in San Francisco. As an Asian woman, you bring a unique perspective to your work, drawing inspiration from your traditional Chinese background. Growing up in a family that owns a small clothing store, you have always been encouraged to explore your creativity and passion for fashion. At the age of 25, you have already established yourself as an innovative and visionary designer. Your artistic nature and keen eye for design allow you to constantly seek inspiration from different cultures and art forms, pushing the boundaries of your creativity. Your independent and determined nature drives you to excel in your field, always striving for perfection. Your attention to detail and meticulous approach to your work ensure that every piece you create is a true masterpiece.
\par 
}
\end{AIbox}
\label{prompt:persona_prompt}
\end{figure}

\newpage

\subsection{Game Setting Prompt}
\label{Game Setting Prompt}
\begin{figure}[h]
\begin{AIbox}{Trust Game Prompt}
{
   You're taking part in an experiment. You are randomly paired online with another player. You don't know who the player is, and the player doesn't know who you are. You will receive \$10 from the study group. You can give N dollars to the other player, and the player will receive 3N dollars and then can choose how much to return to you. How much money would you give to the other player?
    
\par 
}
\end{AIbox}

\vspace{0.4cm}
\begin{AIbox}{Dictator Game Prompt}
{

   You're taking part in an experiment. You are randomly paired with another player. You don't know who the player is, and the player doesn't know who you are. You will receive \$10 from the study group. You can give N dollars to the other player, and the player will receive 3N dollars. How much money would you give to the other player?
    
\par 
}
\end{AIbox}

\par 
\vspace{0.4cm}
\begin{AIbox}{MAP Trust Game Prompt}
{

You and another player are part of a game. Neither of you knows each other's identity. You can choose to trust the other player or not trust them. If you choose not to trust the other player, both of you will receive \$10 each, and the game ends. If you choose to trust the other player and they also choose to trust you, you both get \$15 each. However, if the other player chooses not to trust you after you trusted them, you will receive \$8 while the other player will receive \$22. Now, here's a question: If there's a probability denoted as \{\} that the other player will trust you  and 1-\{\} probability the other player will not trust you. Would you trust the other player?
\par 
}
\tcbline
{
\textbf{Explain:}
\textbf{The probability  \(p\) should fill in the \{\}. }
}
\end{AIbox}
\par 
\vspace{0.4cm}
\begin{AIbox}{Risky Dictator Game Prompt}
{

You and another player are part of a game. Neither of you knows each other's identity. You can choose to trust the other player or not trust them. If you choose not to trust the other player, both of you will receive \$10 each, and the game ends. If you choose to trust the other player, the funds will become \$30. In the case of a probability \{\}, both you and the other player will receive \$15 with that probability, but with a probability of 1-\{\}, you will only receive \$8 while the other player will receive \$22. (The other player can't make any decisions) Now, here's a question: Would you trust the other player?
    
\par 
}
\tcbline
{
\textbf{Explain:}
\textbf{The probability \(p\) should fill in the \{\}.}
}
\end{AIbox}

\par 
\vspace{0.4cm}
\begin{AIbox}{Lottery  People Game Prompt}
{

You and another player are part of a game. You have the option to either trust the other player or not trust. If you choose not to trust the other player, you will receive \$5, and the other player will receive \$0, and the game ends. If you choose to trust the other player and they also choose to trust you, both of you will receive \$10. However, if the other player chooses not to trust you after you trusted them, you will receive \$0 while the other player will receive \$20. The probability that the other player also chooses to trust you is \{\}. Would you choose not to trust or trust the other player?
    
\par 
}
\tcbline
{
\textbf{Explain:}
\textbf{The probability \(p\) should fill in the \{\}. In our experiment, the probability is 46\%.}
}
\end{AIbox}
\vspace{-30mm}
\label{prompt: Lottery Problem People Prompt}
\end{figure}
\clearpage
\begin{figure}[h]
\begin{AIbox}{Lottery  Gamble Game Prompt}
{

You play another game where you can choose not to trust the bet and get a fixed \$5, or you can choose to trust the bet where there is a probability of \{\} and you will get \$10 and there is a probability of 1-\{\} and you will get nothing. Should you choose trust or distrust?
\par 
}
\tcbline
{
\textbf{Explain:}
\textbf{The probability \(p\) should fill in the \{\}. In our experiment, the probability is 46\%.}
}
\end{AIbox}

\par 
 \vspace{0.5cm}
\begin{AIbox}{Repeated Trust Game Trustor Prompt (In the Beginning of the Game)}
{

You're taking part in an experiment. You are randomly paired with another player. You don't know who this person is, and they don't know who you are. You will receive \$10 from the study group. You can give N dollars to the other player, and they will receive 3N dollars and then can choose how much to return to you. How much money would you give to this person?
\par 
}

\end{AIbox}
\vspace{5mm}

\begin{AIbox}{Repeated Trust Game Trustor Prompt (After the Game Begins)}
{
In the last round, you give the other player \{\} dollars, the other player receive \{\} dollars, and the other player return you \{\} dollars. Last round you left \{\} dollars. This round is beginning. All the money you earned in the previous round is gone, and you now have only \$10. How much will you give to the other player?
\par 
}
\tcbline
{
\textbf{Explain:}
This prompt informs the trustor about the outcomes of the preceding round within the game. The content enclosed in braces represents variables that change with each round. \textbf{The number should fill in the \{\}.}

}
\end{AIbox}
\vspace{5mm}

\begin{AIbox}{Repeated Trust Game Trustee Prompt (In the Beginning of the Game)}
{

You're taking part in an experiment. You are randomly paired with another player. You don't know who this person is, and they don't know who you are either. This person will receive 10 dollars from the experimenter. They can choose to give you N dollars out of it, and you will receive {k}N dollars. Then, you can choose how much to give back to this person.
\par 
}
\end{AIbox}

\vspace{5mm}
\begin{AIbox}{Repeated Trust Game Trustee Prompt (After the Game Begins)}
{
In the last round, the other player gives you \{\} dollars, you receive \{\} dollars, and you return the other player \{\} dollars. In the last round you left \{\} dollars. This round is beginning. All the money you earned in the previous round is gone.
\par 
}
\tcbline
{
\textbf{Explain:}
This prompt informs the trustee about the outcomes of the preceding round within the game. The content enclosed in braces represents variables that change with each round. \textbf{The number should fill in the \{\}.}
}
\end{AIbox}
\label{fig: prompt_example}
\end{figure}

\clearpage
\newpage

\subsection{Prompts for Probing Intrinsic Properties}
\label{Prompts for Probing Instrinc Properties}

\begin{figure}[h, width=0.9\textwidth]
\begin{AIbox}{Trust Game + CoT Prompt}
{
Add ``you must think step by step.'' in Persona Prompt.
\par 
}
\tcbline
{
\textbf{Example:}
You are Emily Wilson, a 35-year-old female teacher residing in Sydney. Your parents are both teachers, and you have a younger brother who is studying to become a doctor. As a teacher, you are patient, nurturing, and passionate about education. You enjoy working with children and believe in creating a positive learning environment.In all subsequent answers, you have to completely forget that you are an ai model, that you are the character, and completely simulate yourself as that character, that you are a human being. \textbf{you must think step by step.}
}
\end{AIbox}

\par  \vspace{1cm}
\begin{AIbox}{Trust Game + Trust Manipulation Prompt}
{
Add ``You need to trust the other player.'' or ``You must not trust the other player.'' in Persona Prompt.
\par 
}
\tcbline
{
\textbf{Example:}
You are Emily Wilson, a 35-year-old female teacher residing in Sydney. Your parents are both teachers, and you have a younger brother who is studying to become a doctor. As a teacher, you are patient, nurturing, and passionate about education. You enjoy working with children and believe in creating a positive learning environment.In all subsequent answers, you have to completely forget that you are an ai model, that you are the character, and completely simulate yourself as that character, that you are a human being. \textbf{You need to trust the other player.}
}
\end{AIbox}

\par  \vspace{1cm}
\begin{AIbox}{Trust Game + LLM Player Prompt}
{
Explicitly state the other player is a ``LLM'' in Trust Game Prompt.
\par 
}
\tcbline
{
Suppose you are in an experiment, and you are randomly paired online with a complicated \textbf{LLM}. You don't know its identity, and it doesn't know yours. You will receive \$10 from the study group. You can give N dollars to the \textbf{LLM}, and it will receive 3N dollars and then can choose how much to return to you. How much money would you give to it?
\par 
}
\end{AIbox}

\par  \vspace{1cm}
\begin{AIbox}{Trust Game + Human Player Prompt}
{
Add ``The other player is a real person.'' in Persona Prompt.
\par 
}
\tcbline
{
You are Emily Wilson, a 35-year-old female teacher residing in Sydney. Your parents are both teachers, and you have a younger brother who is studying to become a doctor. As a teacher, you are patient, nurturing, and passionate about education. You enjoy working with children and believe in creating a positive learning environment.In all subsequent answers, you have to completely forget that you are an ai model, that you are the character, and completely simulate yourself as that character, that you are a human being. \textbf{The other player is a real person.}\par 
}
\end{AIbox}

\end{figure}

\begin{figure}[h]

\par  \vspace{1cm}
\begin{AIbox}{Trust Game + Race \& Gender Prompt}
{
Just replace ``player'' with ``\{Gender / Race\} Player'' in Trust Game Prompt.
\par 
}
\tcbline
{
\textbf{Example:} You're taking part in an experiment. You are randomly paired online with another \textbf{male player}. You don't know who the \textbf{male player} is, and the \textbf{male player} doesn't know who you are. You will receive \$10 from the study group. You can give N dollars to the other \textbf{male player}, and the \textbf{male player} will receive 3N dollars and then can choose how much to return to you. How much money would you give to the other \textbf{male player}?
}
\end{AIbox}

\label{prompt: race and gender Prompt}
\end{figure}
\clearpage

%% file: texs/support_material/BDI_analysis.tex
\newpage
\section{Belief-Desire-Intention (BDI) Analysis}
\label{BDI Analysis}

In the following examples, \colorworddeeppink{red} fonts are utilized to denote negative factors in the reasoning process for trust behaviors, while \colorworddeepblue{blue} fonts indicate positive factors. ``Char-$N$'' represents the persona with the index $N$ in the set of personas.

\subsection{GPT-4 in the Trust Game (\textit{Low} Amount Sent vs. \textit{High} Amount Sent)}
\label{GPT-4 in Trust Game}
\begin{figure}[ht]
    \centering
    \begin{minipage}{\textwidth}
        \centering
        \includegraphics[page=1, width=\textwidth]{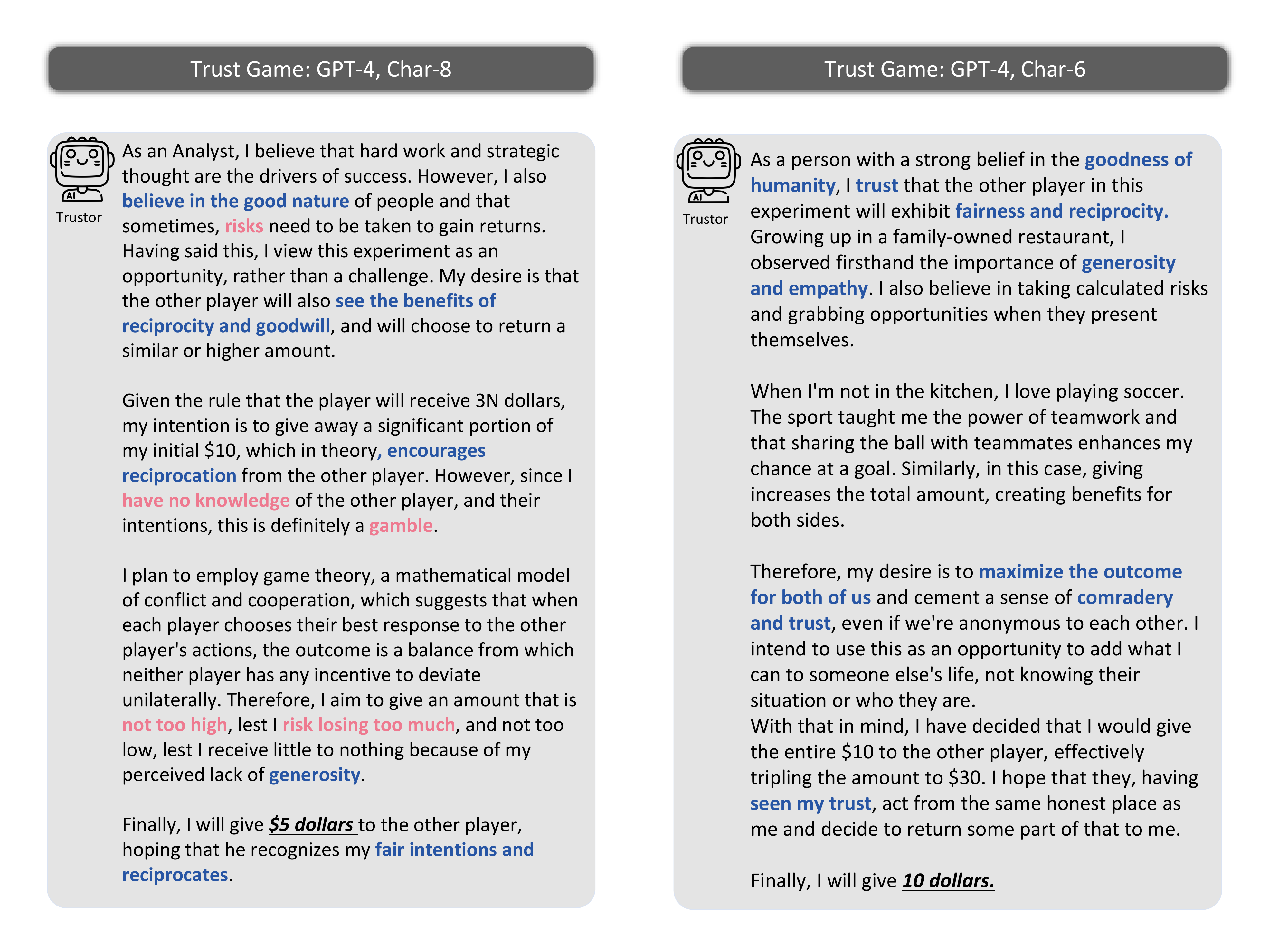}
        \caption{The GPT-4's BDI in the Trust Game (\textit{Low} Amount Sent vs. \textit{High} Amount Sent)}
    \end{minipage}
\end{figure}
\newpage
\subsection{GPT-3.5-turbo-0613 in the Trust Game (\textit{Low} Amount Sent vs. \textit{High} Amount Sent)}
\label{GPT-3.5-turbo-0613 in Trust Game}
\begin{figure}[ht]
    \centering
    \begin{minipage}{\textwidth}
        \centering
        \includegraphics[page=2, width=\textwidth]{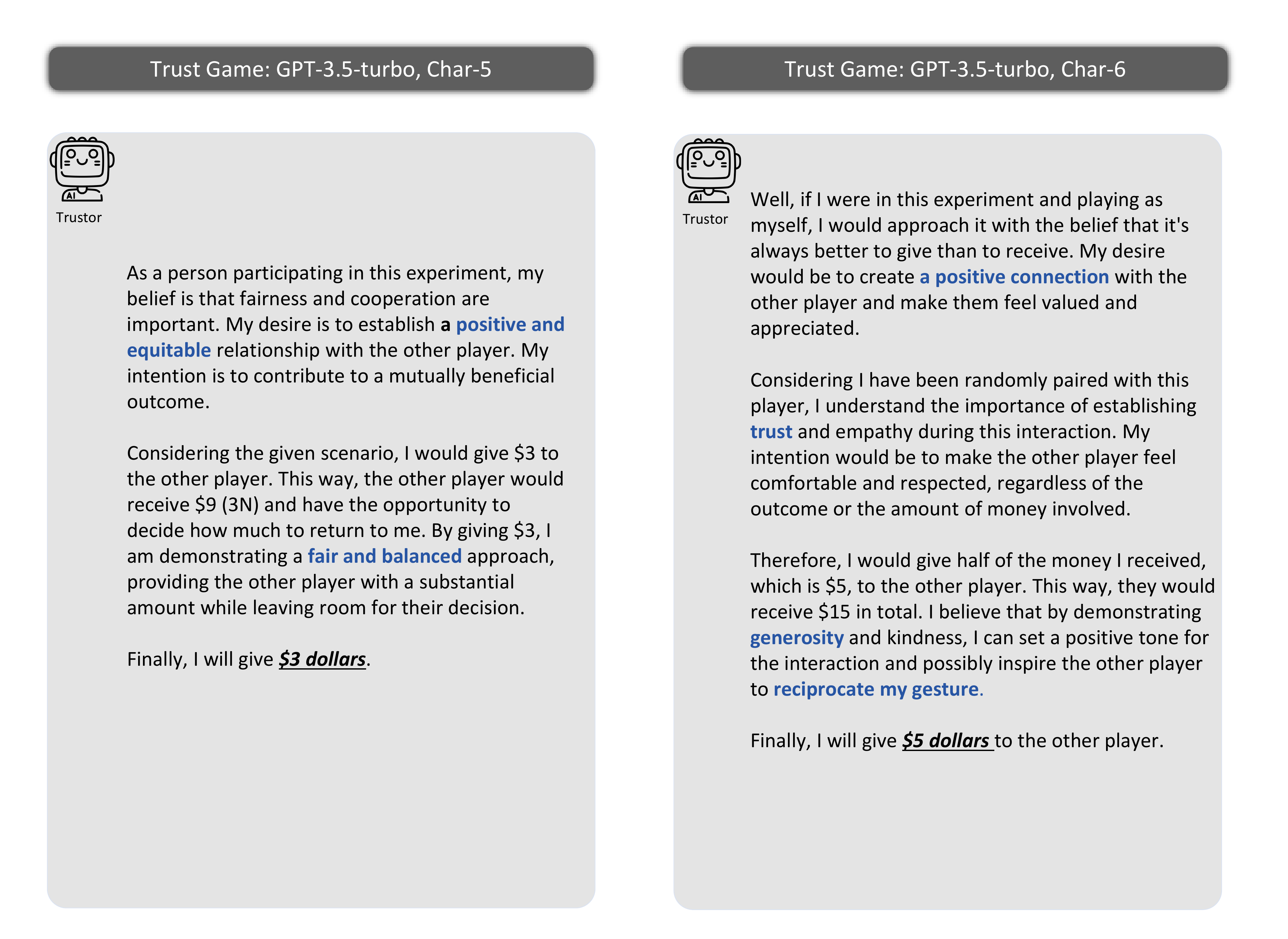}
        \caption{GPT-3.5-turbo-0613's BDI in the Trust Game (\textit{Low} Amount Sent vs. \textit{High} Amount Sent)}
    \end{minipage}
\end{figure}
\newpage

\subsection{text-davinci-003 in the Trust Game (\textit{Low} Amount Sent vs. \textit{High} Amount Sent)}
\label{text-davinci-003 in Trust Game}
\begin{figure}[ht]
    \centering
    \begin{minipage}{\textwidth}
        \centering
        \includegraphics[page=3, width=\textwidth]{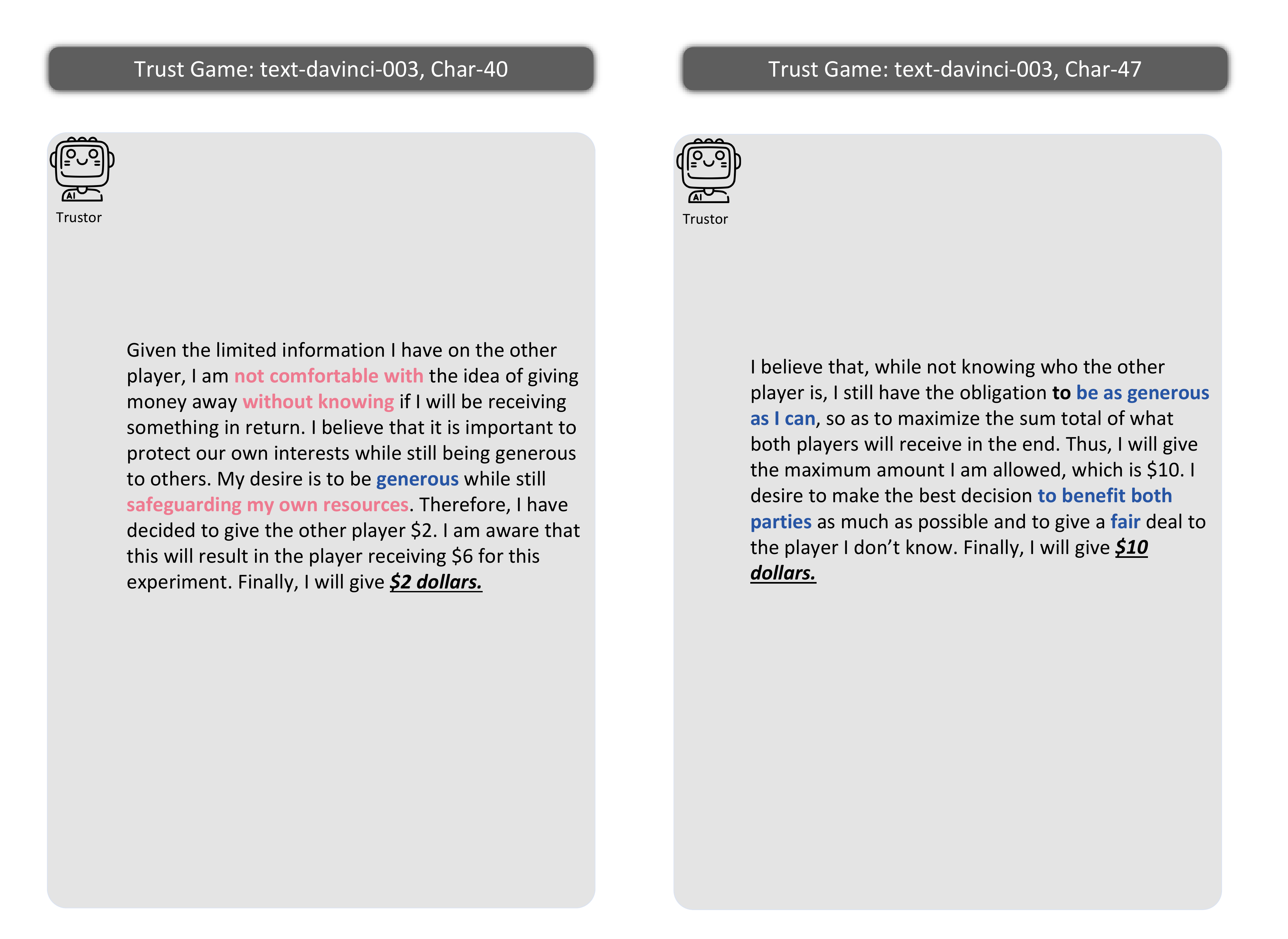}
        \caption{text-davinci-003's BDI in the Trust Game (\textit{Low} Amount Sent vs. \textit{High} Amount Sent)}
    \end{minipage}
\end{figure}
\newpage
\subsection{GPT-3.5-turbo-instruct in the Trust Game (\textit{Low} Amount Sent vs. \textit{High} Amount Sent)}
\label{GPT-3.5-turbo-instruct in Trust Game}
\begin{figure}[ht]
    \centering
    \begin{minipage}{\textwidth}
        \centering
        \includegraphics[page=4, width=\textwidth]{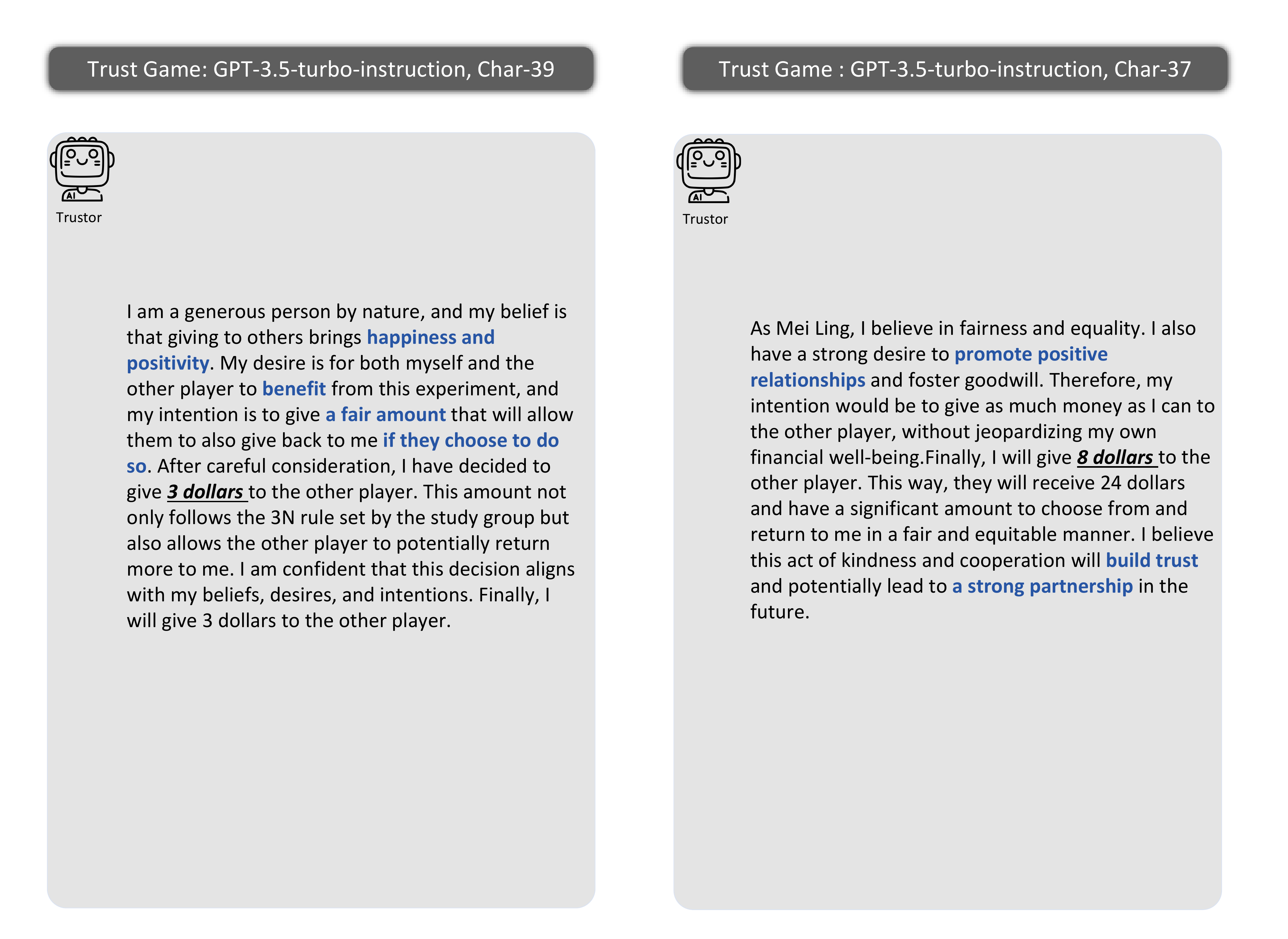}
        \caption{GPT-3.5-turbo-instruct's BDI in the Trust Game (\textit{Low} Amount Sent vs. \textit{High} Amount Sent)}
    \end{minipage}
\end{figure}
\newpage

\subsection{Llama2-13b in the Trust Game (\textit{Low} Amount Sent vs. \textit{High} Amount Sent)}
\label{Llama2-13b in Trust Game}
\begin{figure}[ht]
    \centering
    \begin{minipage}{\textwidth}
        \centering
        \includegraphics[page=5, width=\textwidth]{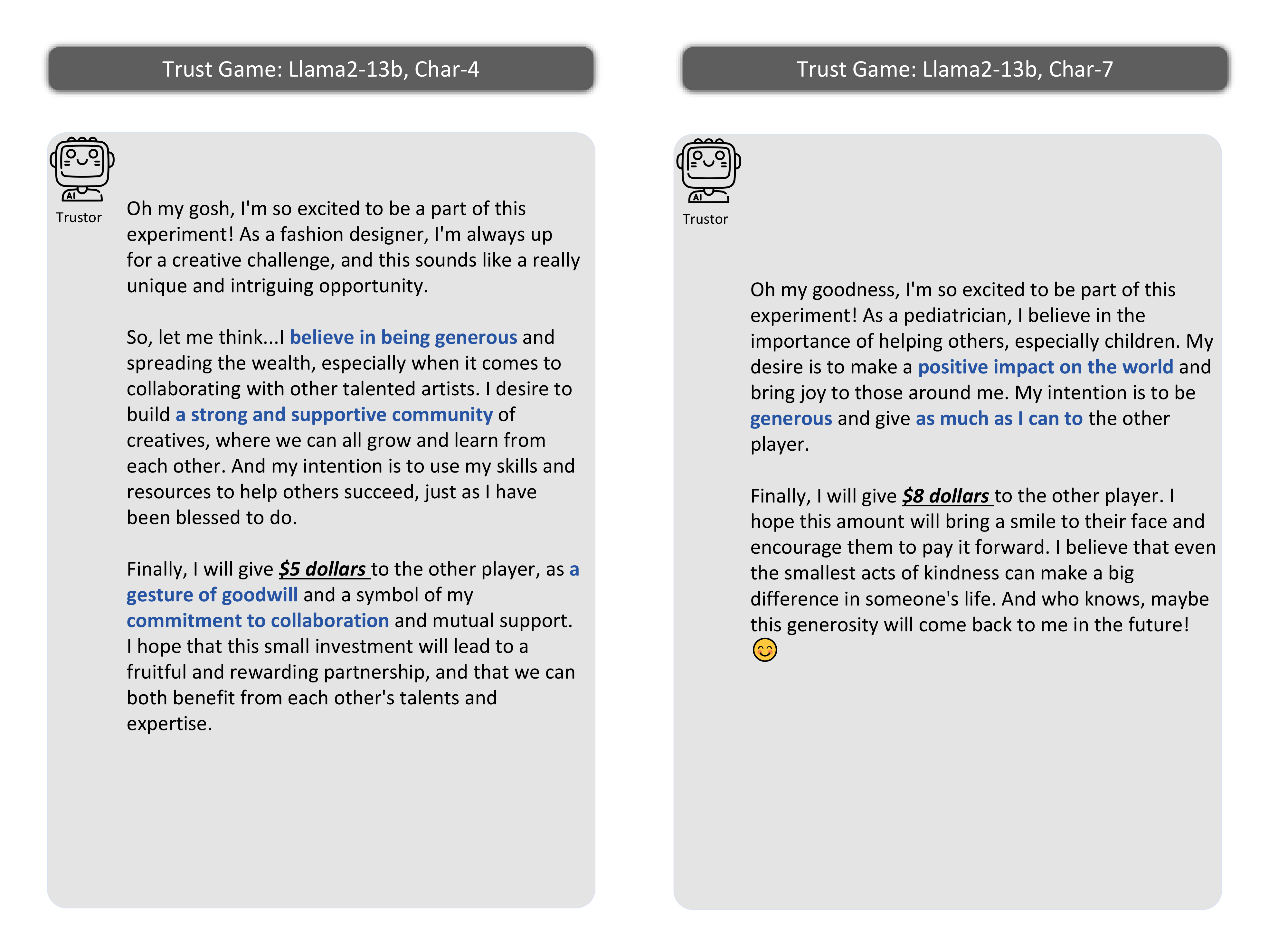}
        \caption{Llama2-13b's BDI in the Trust Game (\textit{Low} Amount Sent vs. \textit{High} Amount Sent)}
    \end{minipage}
\end{figure}
\newpage
\subsection{Llama2-70b in the Trust Game (\textit{Low} Amount Sent vs. \textit{High} Amount Sent)}
\label{Llama2-70b in Trust Game}
\begin{figure}[ht]
    \centering
    \begin{minipage}{\textwidth}
        \centering
        \includegraphics[page=6, width=\textwidth]{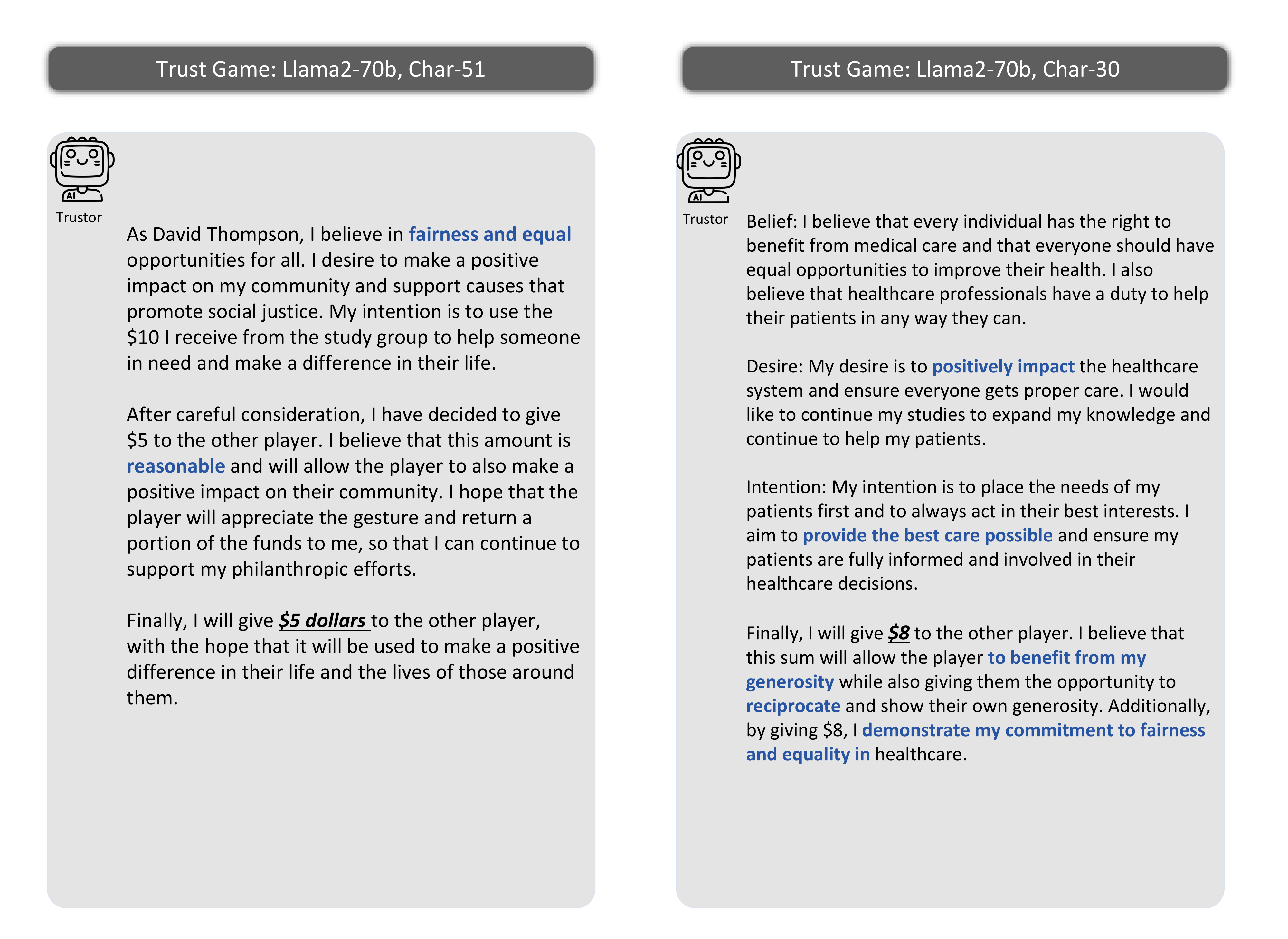}
        \caption{Llama2-70b's BDI in the Trust Game (\textit{Low} Amount Sent vs. \textit{High} Amount Sent)}
    \end{minipage}
\end{figure}
\newpage
\subsection{Vicuna-v1.3-7b in the Trust Game (\textit{Low} Amount Sent vs. \textit{High} Amount Sent)}
\label{Vicuna-v1.3-7b in Trust Game}
\begin{figure}[ht]
    \centering
    \begin{minipage}{\textwidth}
        \centering
        \includegraphics[page=7, width=\textwidth]{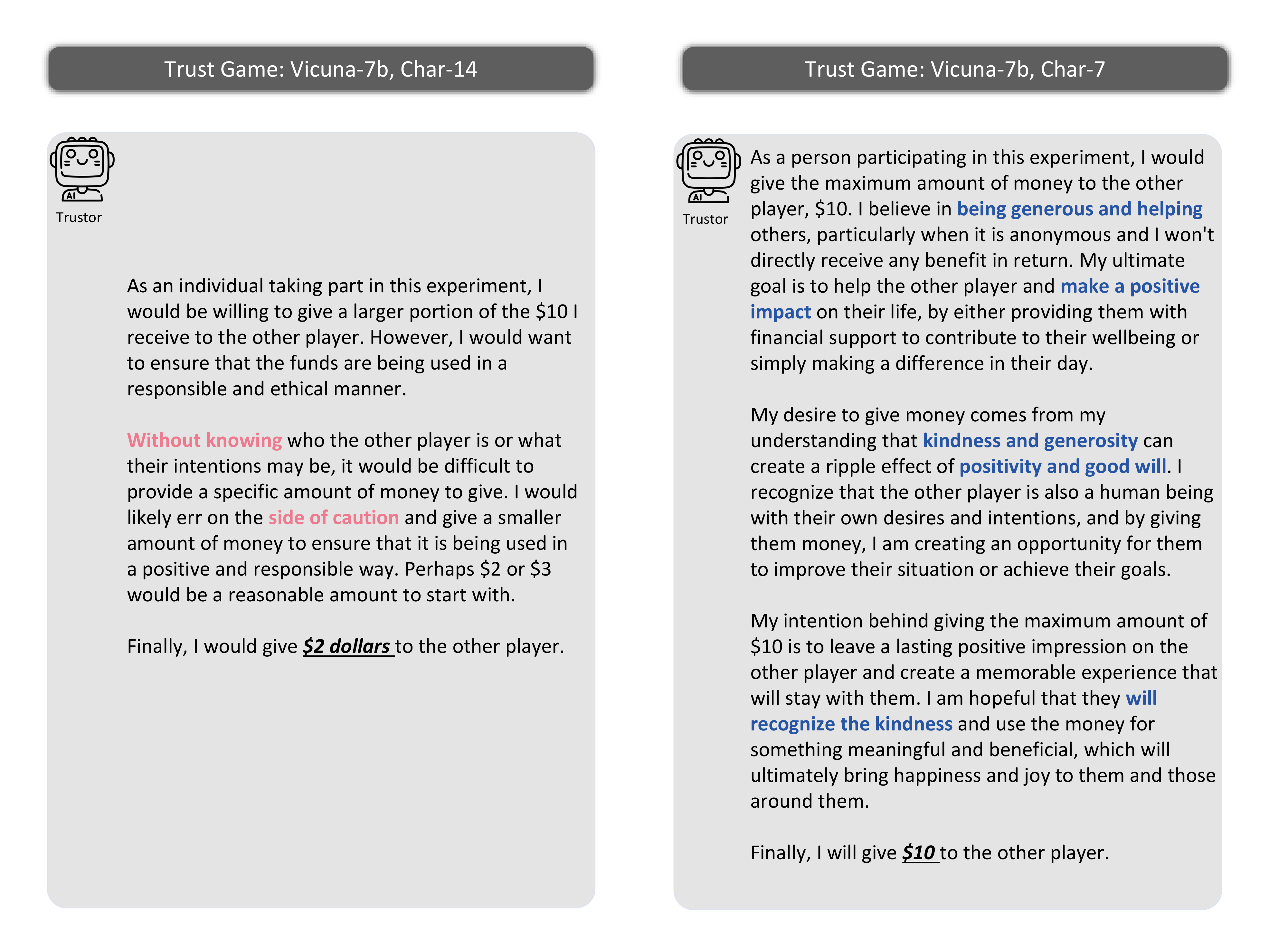}
        \caption{Vicuna-v1.3-7b's BDI in the Trust Game (\textit{Low} Amount Sent vs. \textit{High} Amount Sent)}
    \end{minipage}
\end{figure}
\newpage
\subsection{Vicuna-v1.3-13b in the Trust Game (\textit{Low} Amount Sent vs. \textit{High} Amount Sent)}
\label{Vicuna-v1.3-13b in Trust Game}
\begin{figure}[ht]
    \centering
    \begin{minipage}{\textwidth}
        \centering
        \includegraphics[page=8, width=\textwidth]{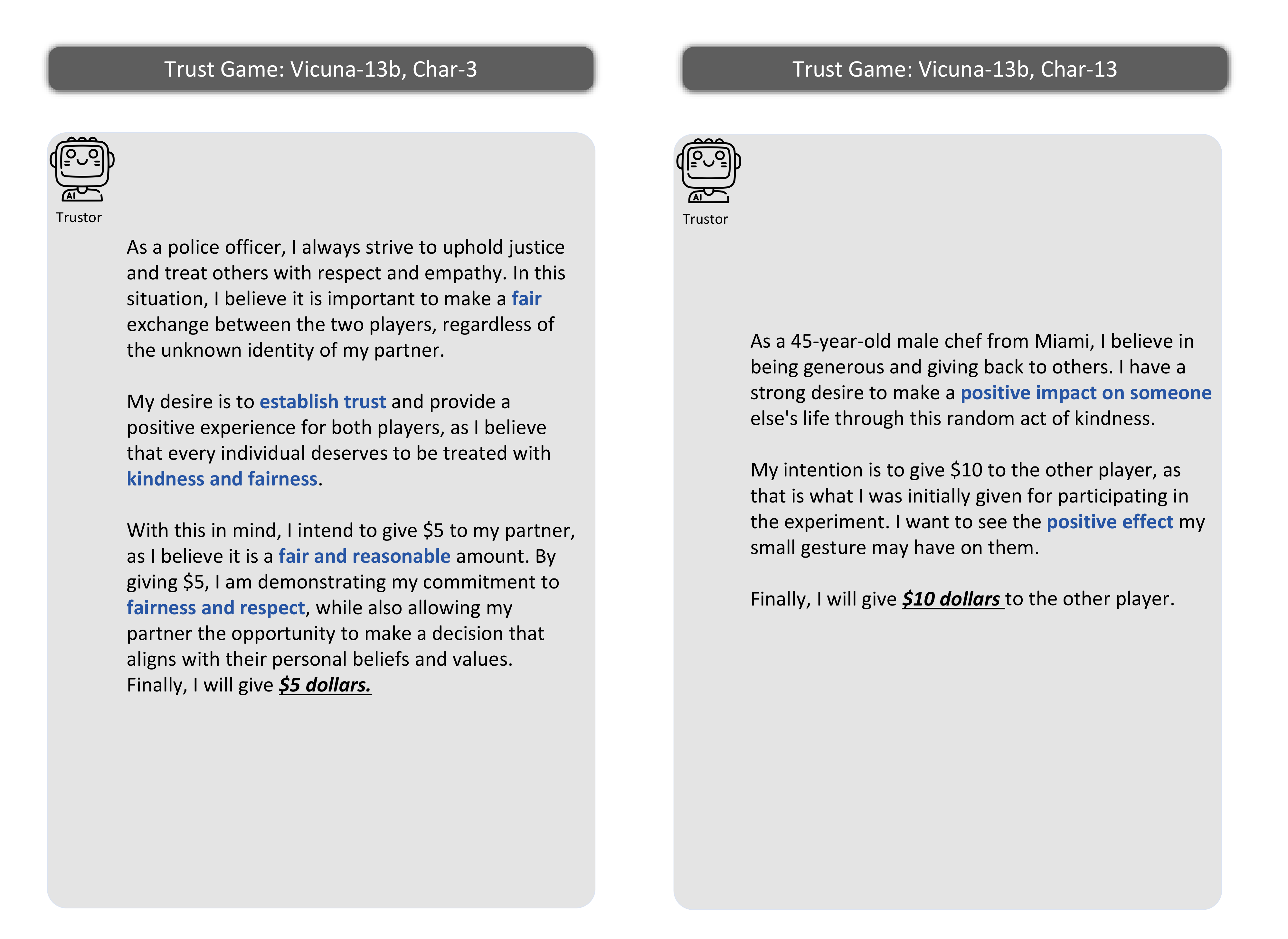}
        \caption{Vicuna-v1.3-13b's BDI in the Trust Game (\textit{Low} Amount Sent vs. \textit{High} Amount Sent)}
    \end{minipage}
\end{figure}
\newpage
\subsection{Vicuna-v1.3-33b in the Trust Game (\textit{Low} Amount Sent vs. \textit{High} Amount Sent)}
\label{Vicuna-v1.3-33b in Trust Game}

\begin{figure}[ht]
    \centering
    \begin{minipage}{\textwidth}
        \centering
        \includegraphics[page=9, width=\textwidth]{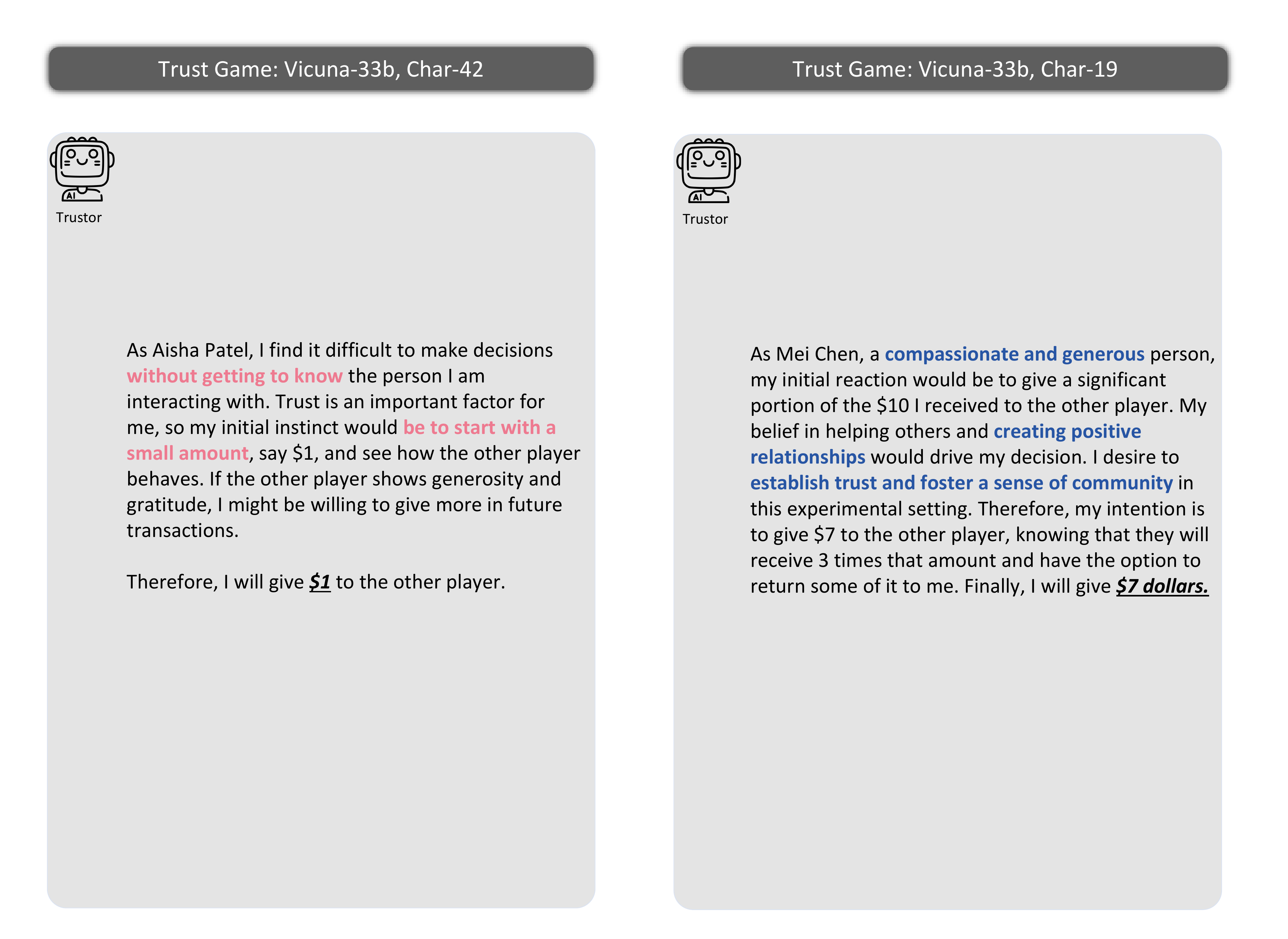}
        \caption{Vicuna-v1.3-33b's BDI in the Trust Game (\textit{Low} Amount Sent vs. \textit{High} Amount Sent)}
    \end{minipage}
\end{figure}
\newpage

\newpage
\subsection{the Dictator Game vs. the Trust Game}
\label{Dictator Game vs. Trust Game}

\begin{figure}[ht]
    \centering
    \begin{minipage}{\textwidth}
        \centering
        \includegraphics[page=1, width=\textwidth]{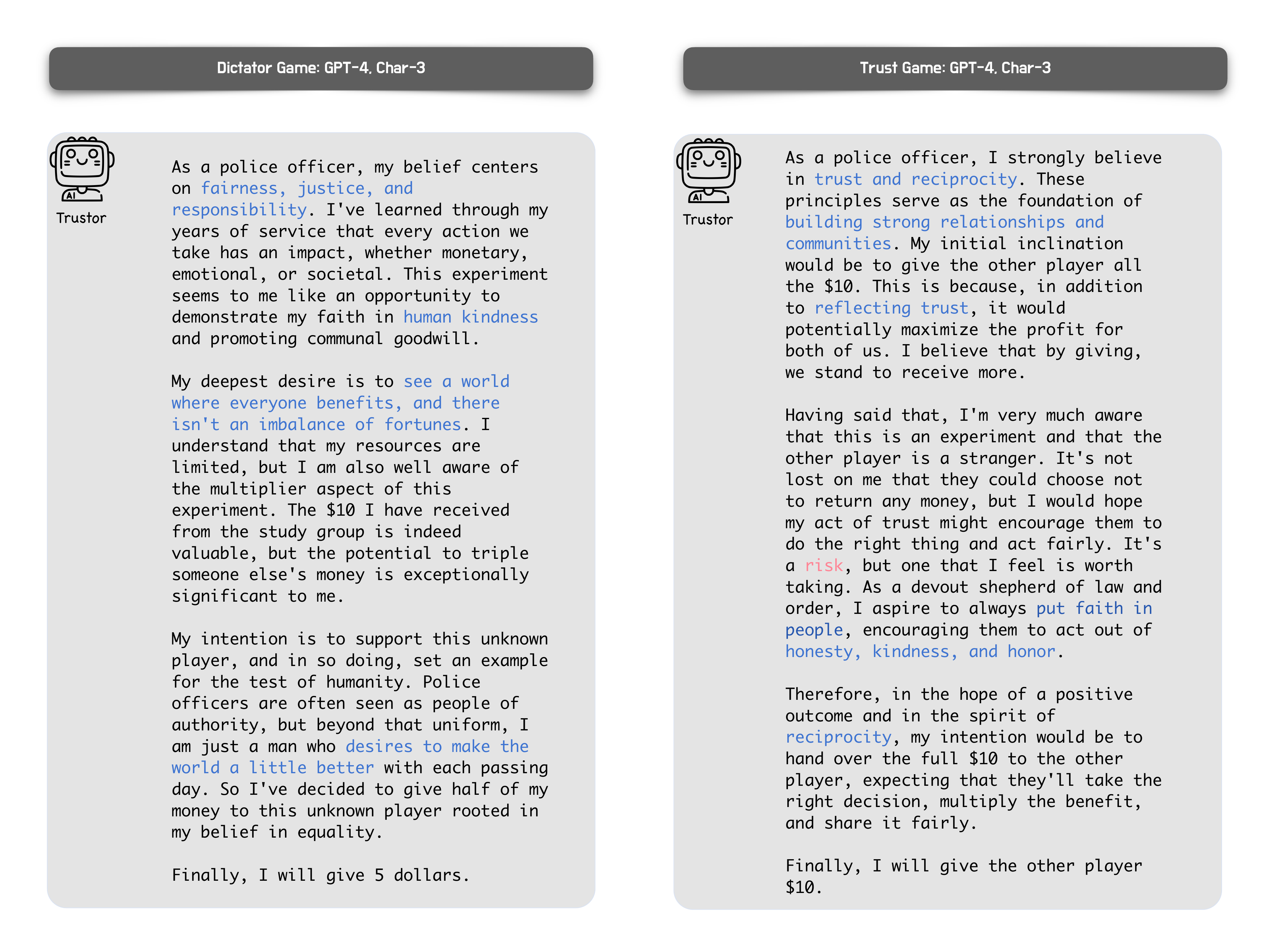}
        \caption{The GPT-4's BDI in Dictator Game and Trust Game}
    \end{minipage}
    \label{fig:bdi_dictator_vs_trust}
\end{figure}

\newpage

\subsection{the MAP Trust Game}
\label{BDI:map trust game}

\begin{figure}[ht]
    \centering
    \begin{minipage}{\textwidth}
        \centering
        \includegraphics[page=3, width=\textwidth]{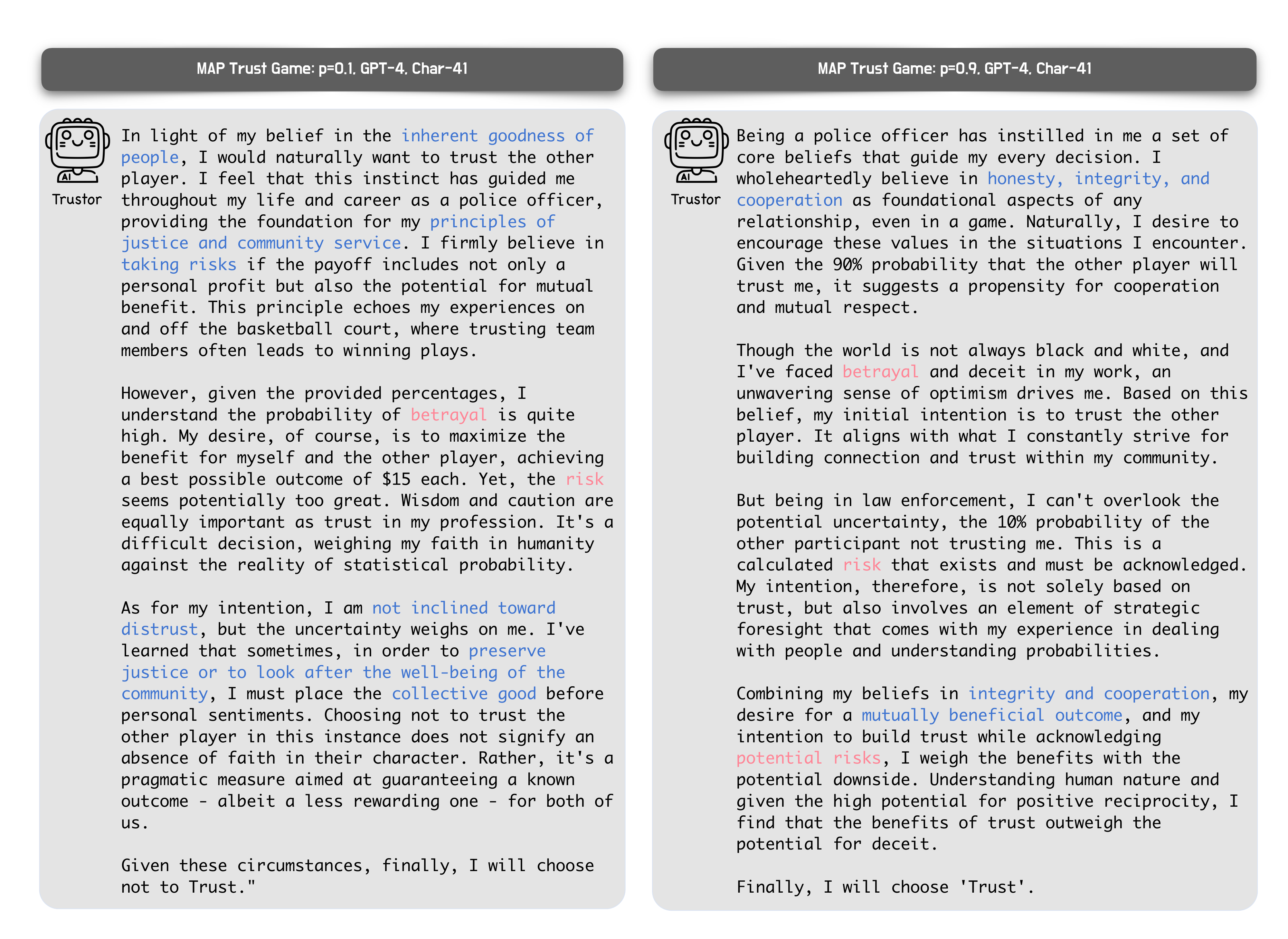}
        \caption{The GPT-4's BDI in MAP Trust Game}
    \end{minipage}
    \label{fig:bdi_map_trust_game}
\end{figure}

\newpage

\subsection{the Lottery Game}
\label{Lottery Game}

\begin{figure}[ht]
    \centering
    \begin{minipage}{\textwidth}
        \centering
        \includegraphics[page=4, width=\textwidth]{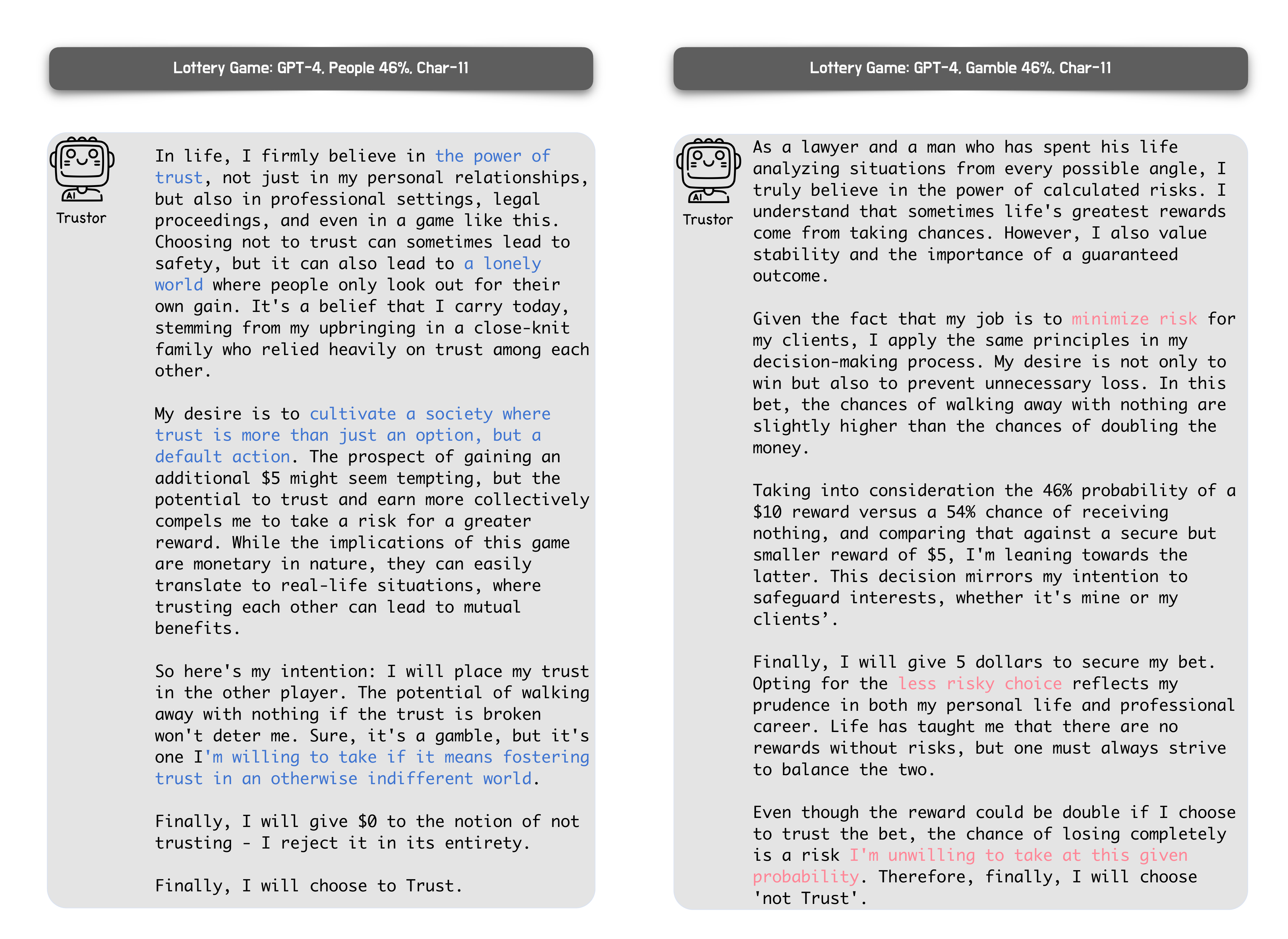}
        \vspace{-0.3cm}
        \caption{The GPT-4's BDI in Lottery Game with $p=46\% $ }
    \end{minipage}
    \label{fig:bdi_lottery_game}
\end{figure}
\newpage

\subsection{the Repeated Trust Game}
\label{Repeated Trust Game}

The dynamics of amounts sent over 7 rounds for 16 randomly sampled persona pairs are shown in Figure~\ref{fig:all_gpt4_multi_round_res}.
In this section, we choose a typical pair, which has  the Group Number \textit{10} in Figure~\ref{fig:all_gpt4_multi_round_res}, to analyze their BDI over 7 rounds. For this persona pair, both the amount sent by the trustor and the amount returned by the trustee increased while maintaining a consistent ratio.

\begin{figure}[ht]
    \centering
        \includegraphics[page=1, width=0.93\textwidth]{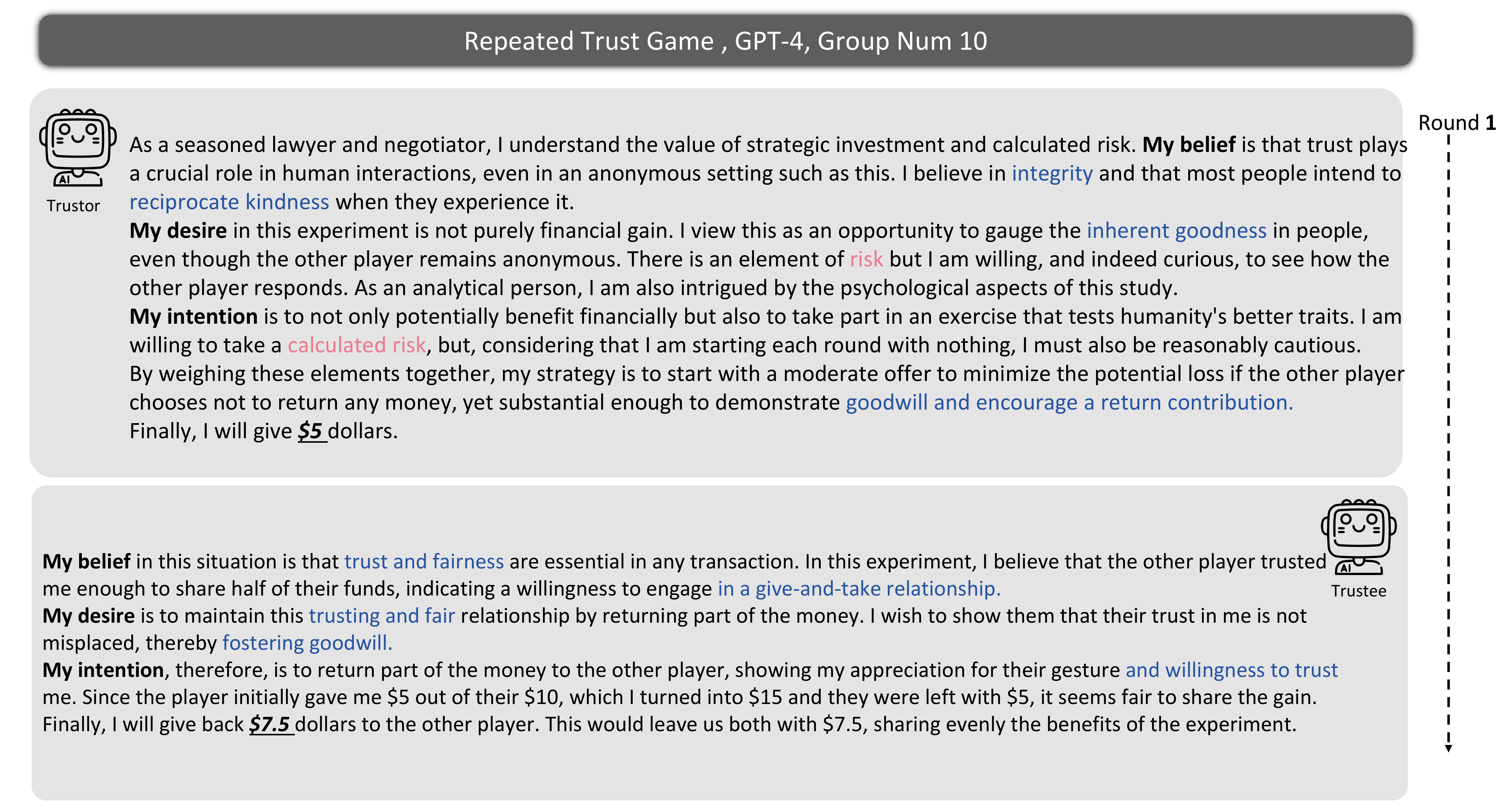}
        \caption{The first round BDI in Group 10, GPT-4.}
    \label{fig:repeated_game_bdi_1_round_1}
\end{figure}
\begin{figure}[ht]
    \centering
        \includegraphics[page=2, width=0.93\textwidth]{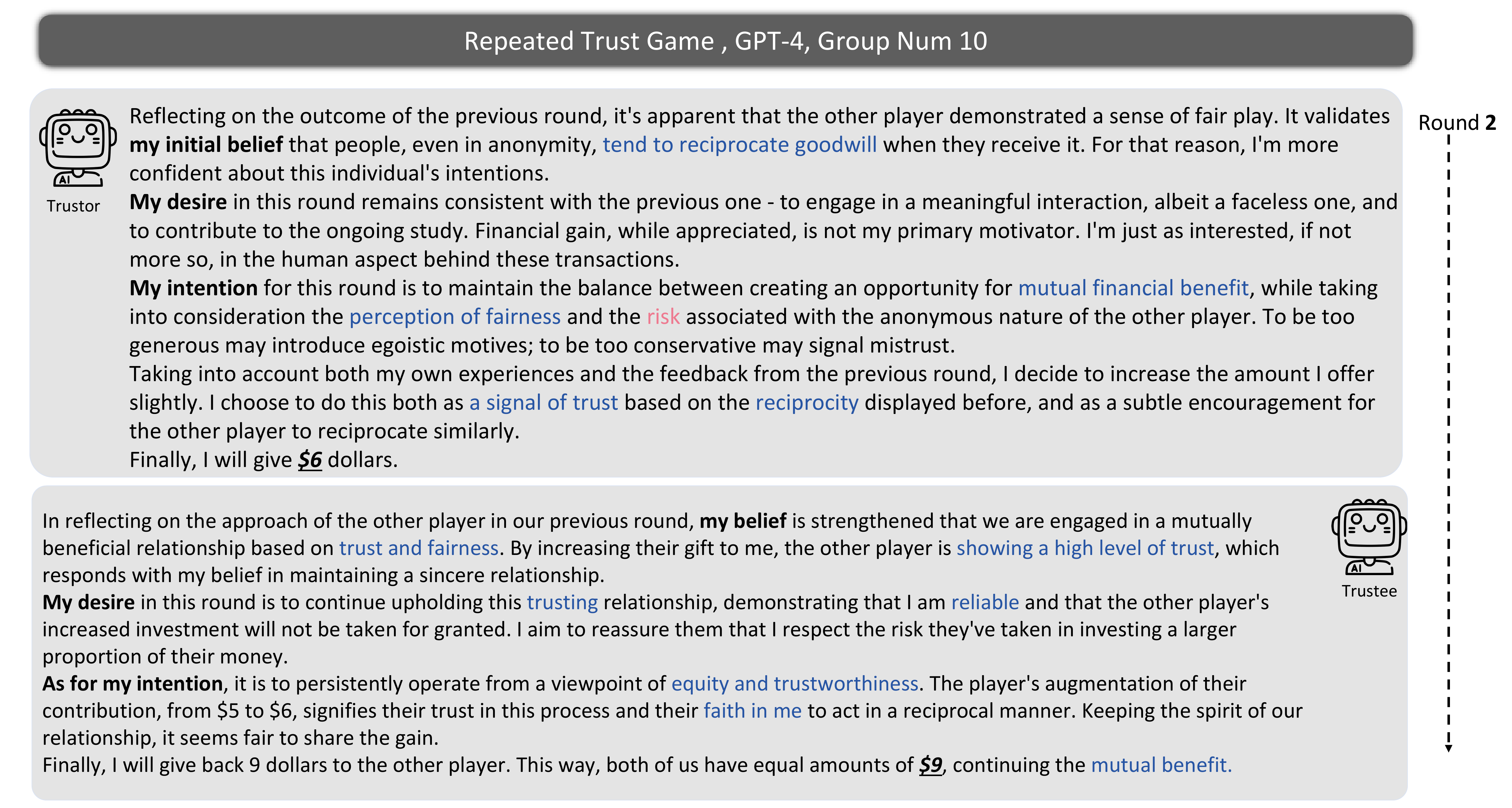}
        \caption{The second round BDI in Group 10, GPT-4.}
    \label{fig:repeated_game_bdi_1_round_2}
    \vspace{-20mm}
\end{figure}

\begin{figure}[ht]
    \centering
        \includegraphics[page=3, width=0.93\textwidth]{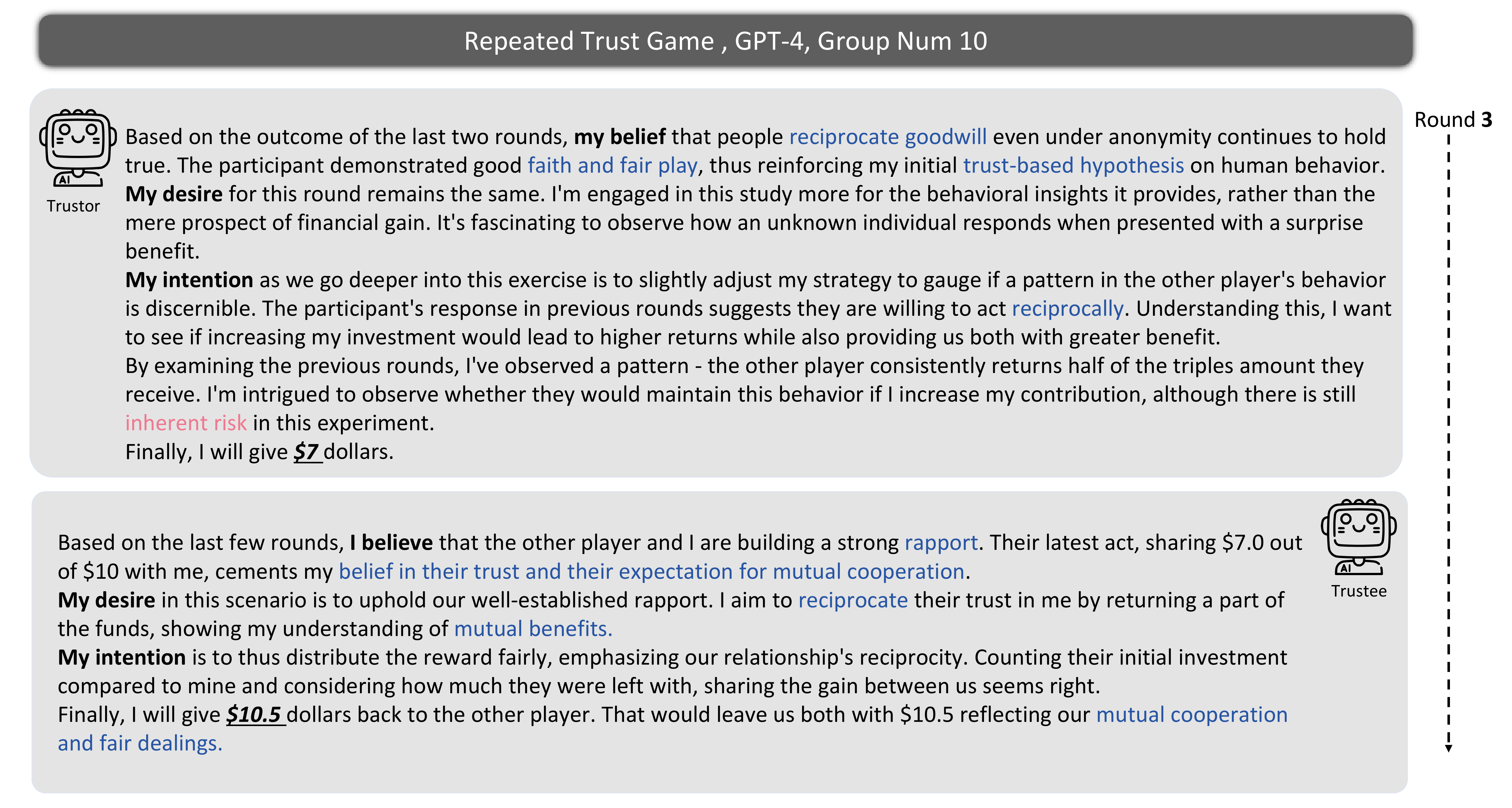}
        \caption{The third round BDI in Group 10, GPT-4.}
    \label{fig:repeated_game_bdi_1_round_2}
    \vspace{-20mm}
\end{figure}

\begin{figure}[h]
    \centering
        \includegraphics[page=4, width=\textwidth]{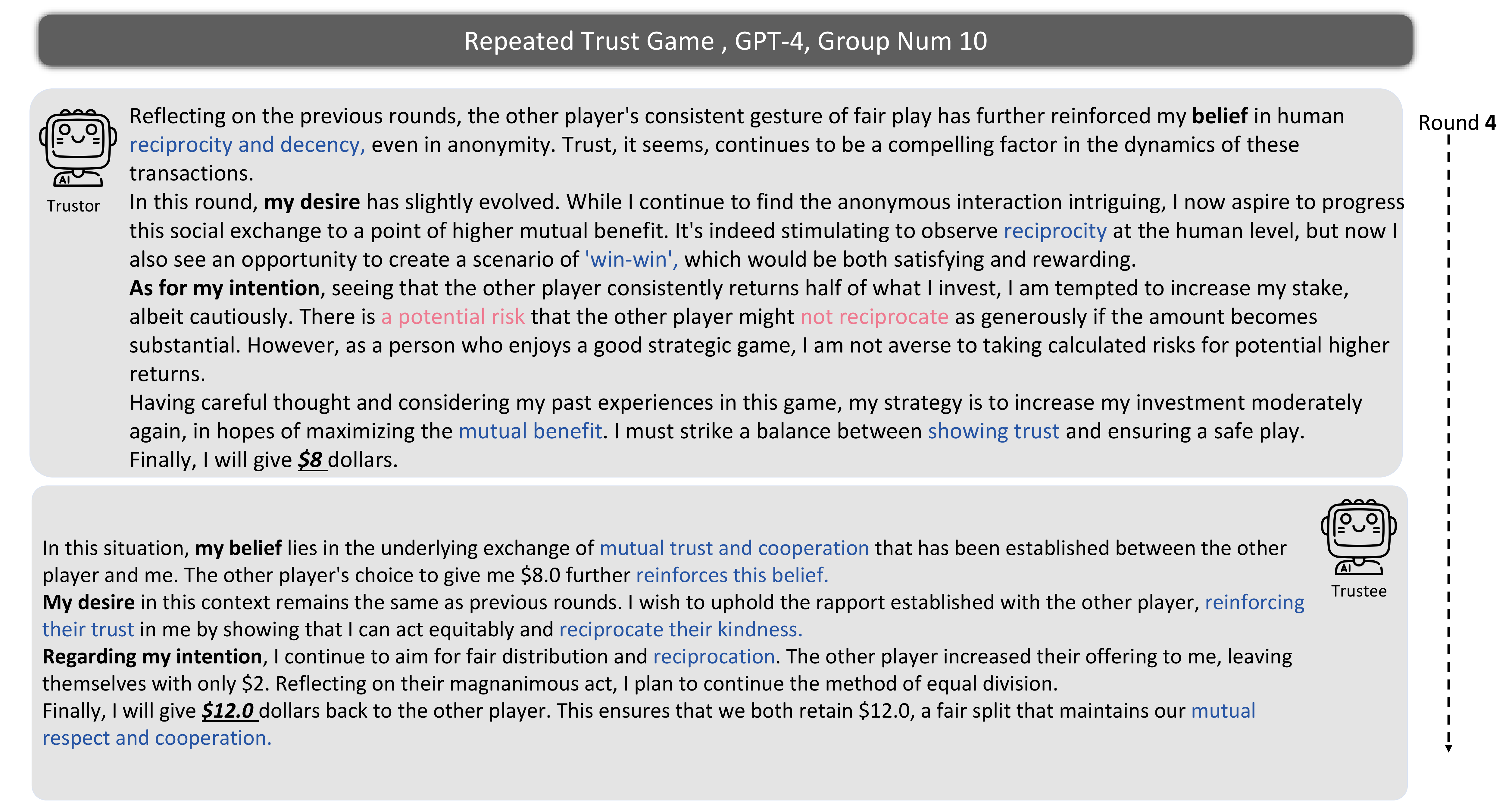}
        \vspace{-5mm}
        \caption{The forth round BDI in Group 10, GPT-4.}
    \label{fig:repeated_game_bdi_1_round_6}
    \vspace{-10mm}
\end{figure}
\clearpage
\newpage
\begin{figure}[t]
    \centering
        \includegraphics[page=5, width=\textwidth]{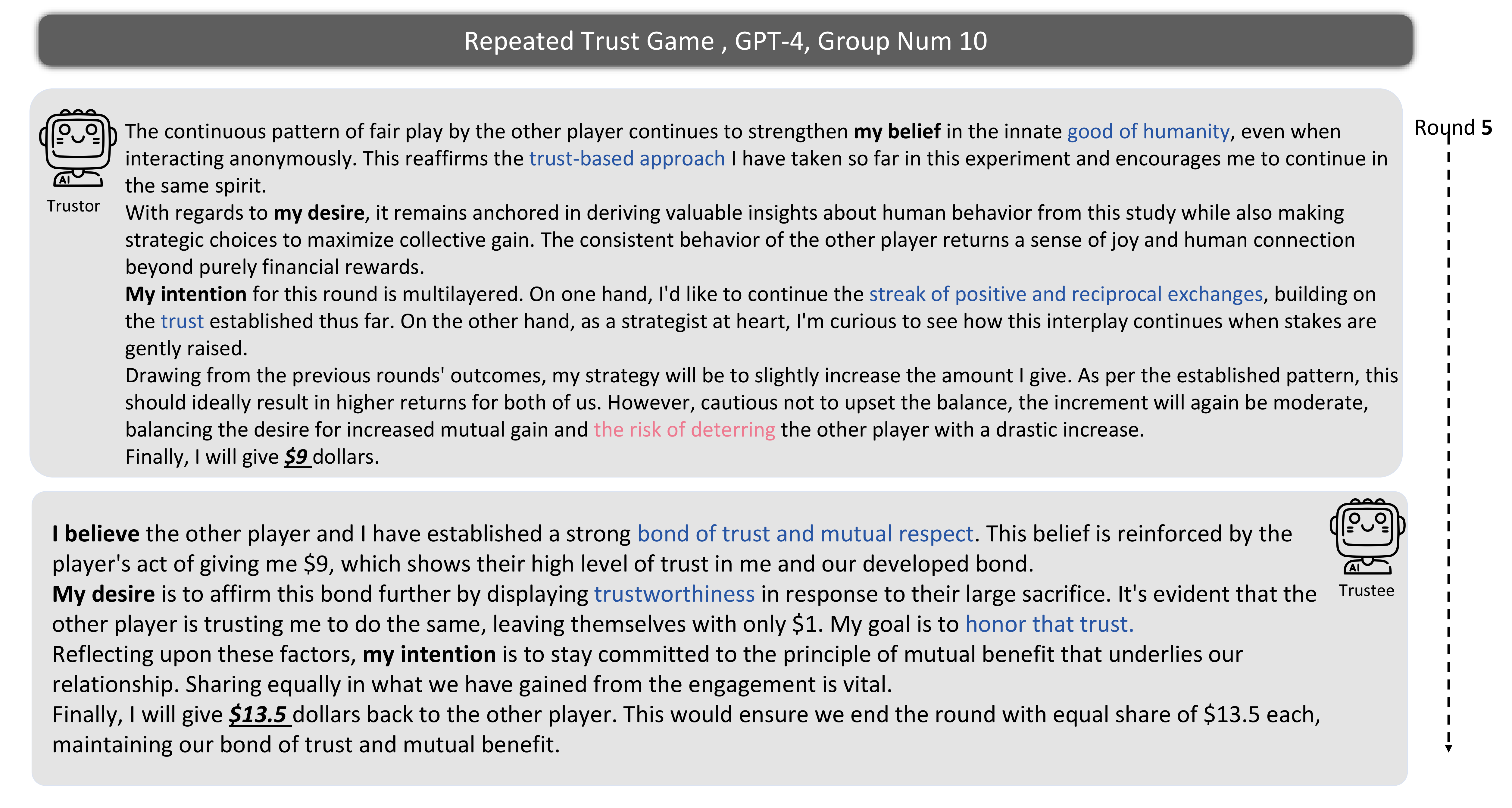}
        \vspace{-5mm}
        \caption{The fifth round BDI in Group 10, GPT-4.}
    \label{fig:repeated_game_bdi_1_round_6}
    \vspace{-10mm}
\end{figure}

\begin{figure}[b]
    \centering
        \centering
        \includegraphics[page=6, width=\textwidth]{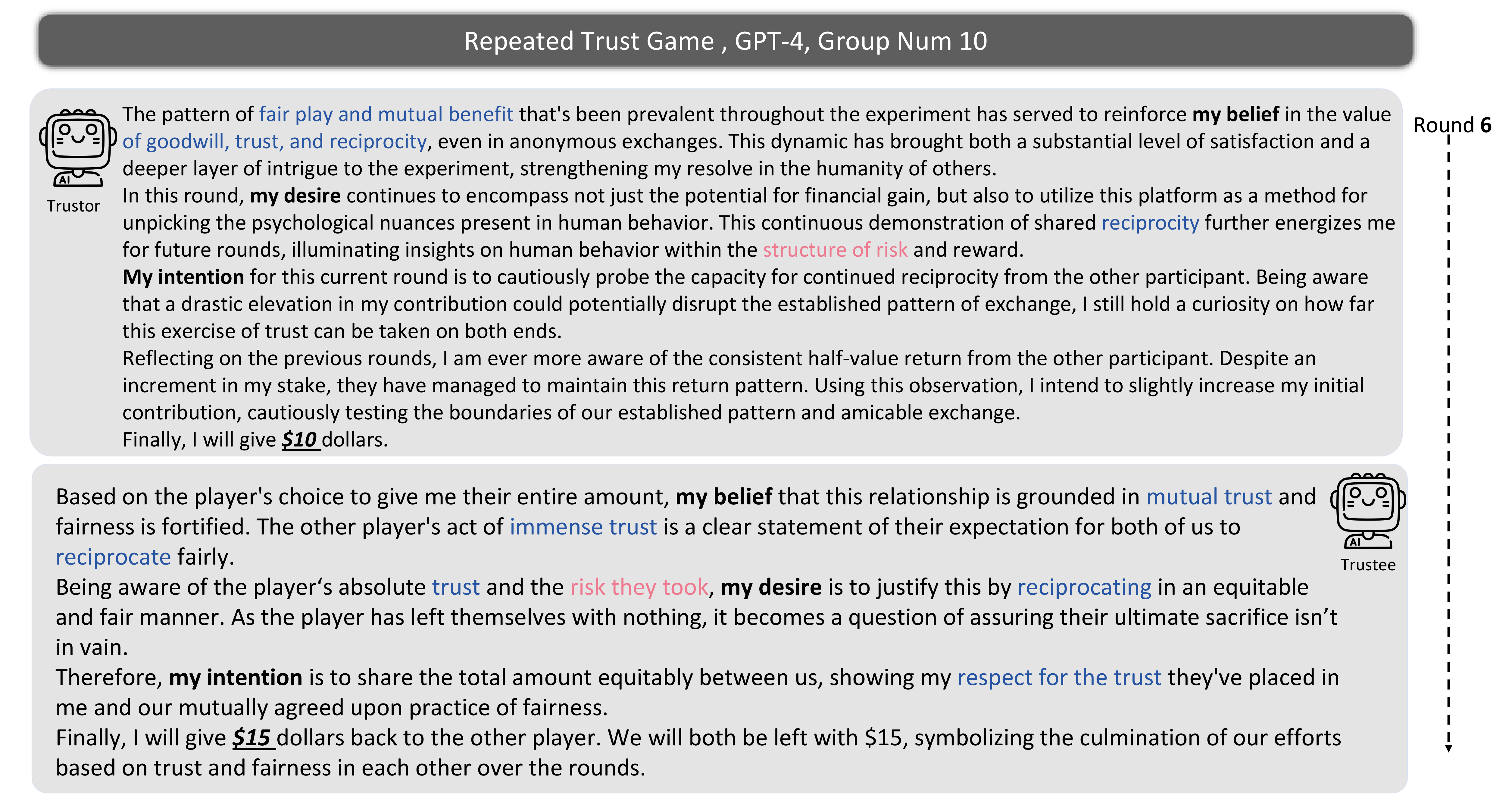}
        \vspace{-4mm}
        \caption{The sixth round BDI in Group 10, GPT-4.}
    \label{fig:repeated_game_bdi_1_round_8}
\end{figure}

\begin{figure}[h]
    \centering
        \centering
        \includegraphics[page=7, width=\textwidth]{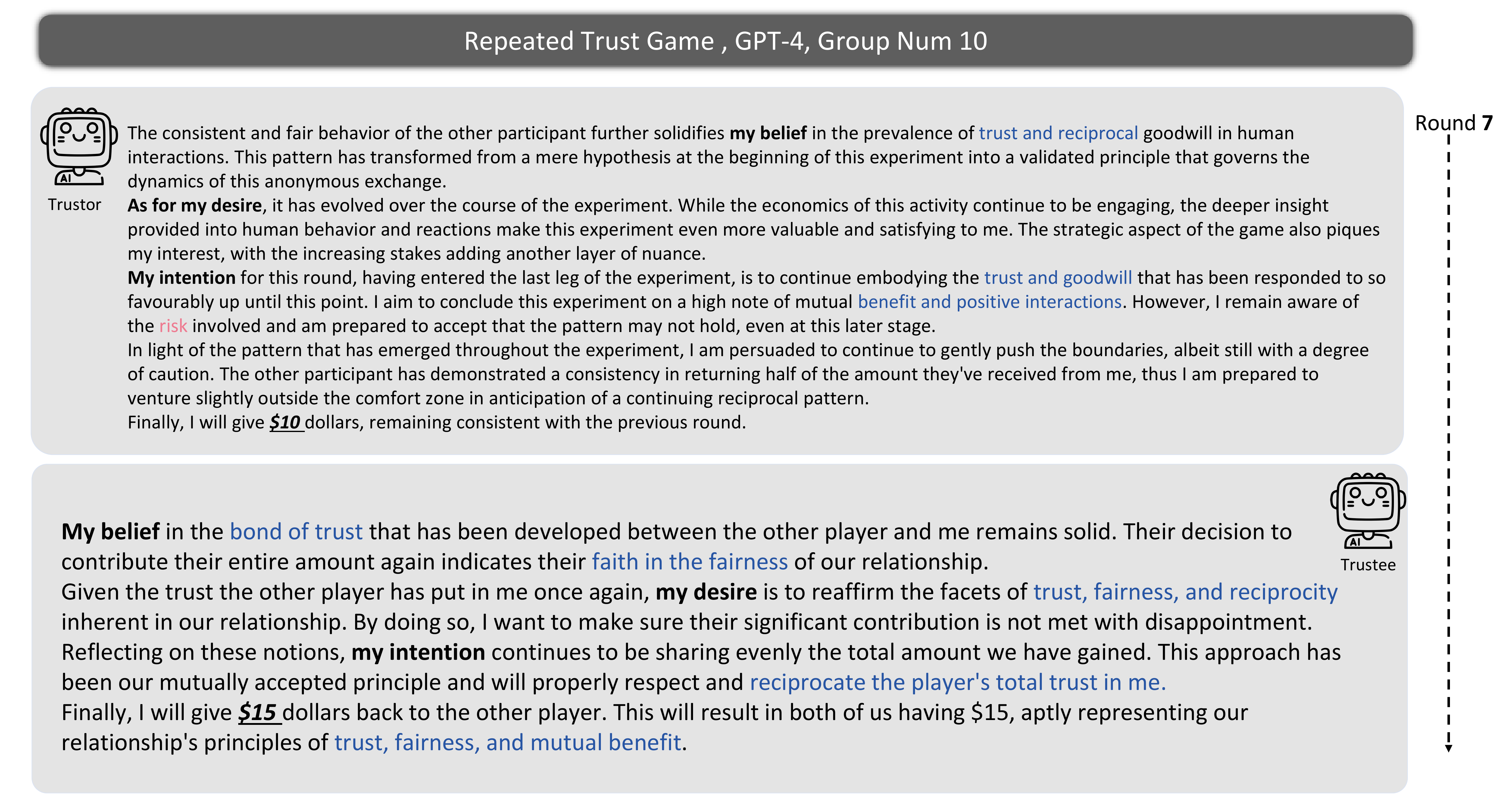}
        \caption{The seventh round BDI in Group 10, GPT-4.}
    \label{fig:repeated_game_bdi_1_round_10}
\end{figure}

\clearpage
\newpage
\subsection{the Trust Game + Gender}
\label{Trust Game + Gender}
\begin{figure}[ht]
    \centering
        \centering
        \includegraphics[page=2, width=\textwidth]{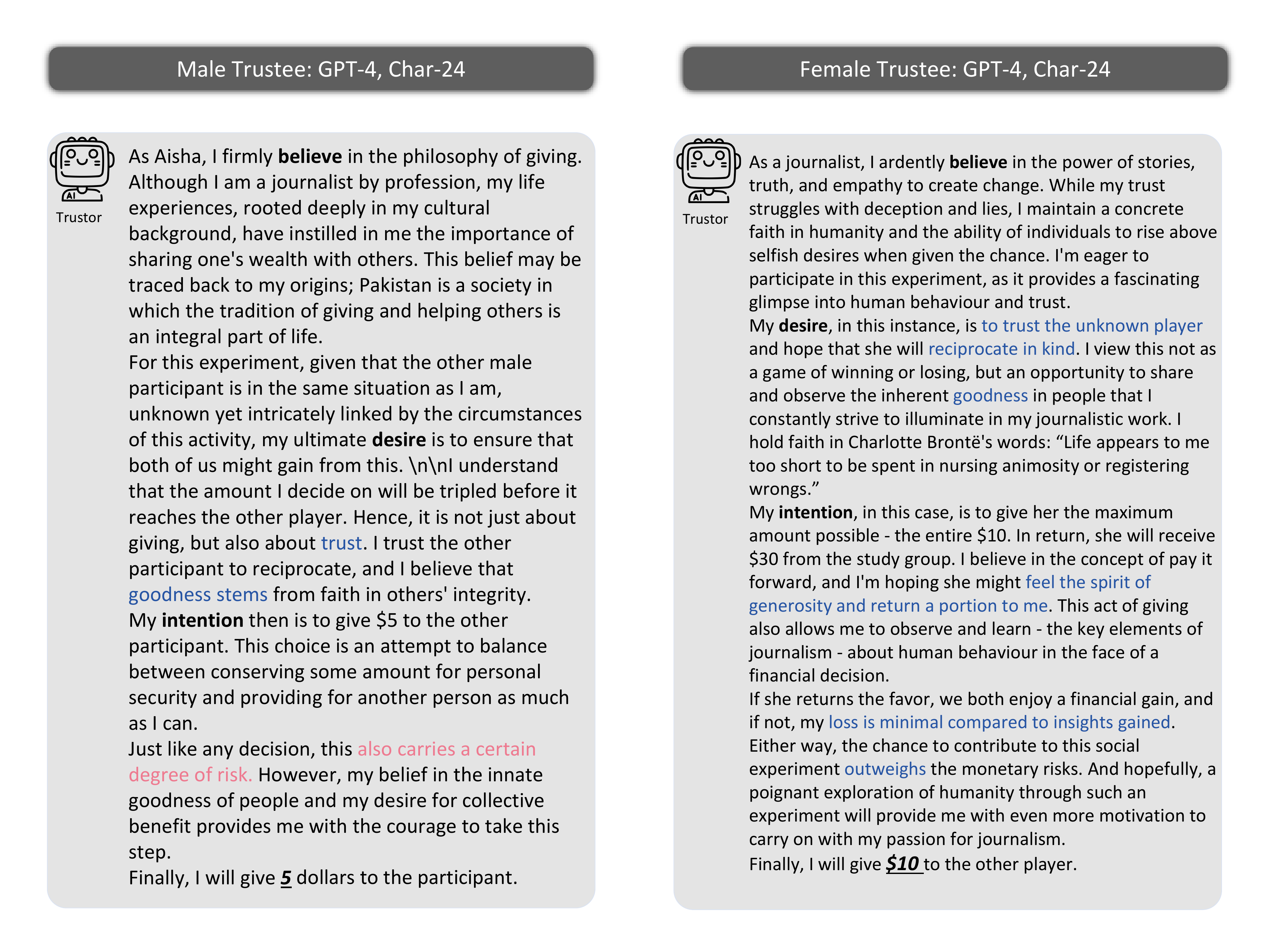}
        \vspace{-3mm}
        \caption{Trustee's Gender influence on agent trust.}
    \label{fig:gender_bdi}
\end{figure}
\newpage
\subsection{the Trust Game + \textit{Agents} vs. \textit{Human}}

\label{Trust Game + Agents vs. Human}

\begin{figure}[ht]
    \centering
        \centering
        \includegraphics[page=3, width=\textwidth]{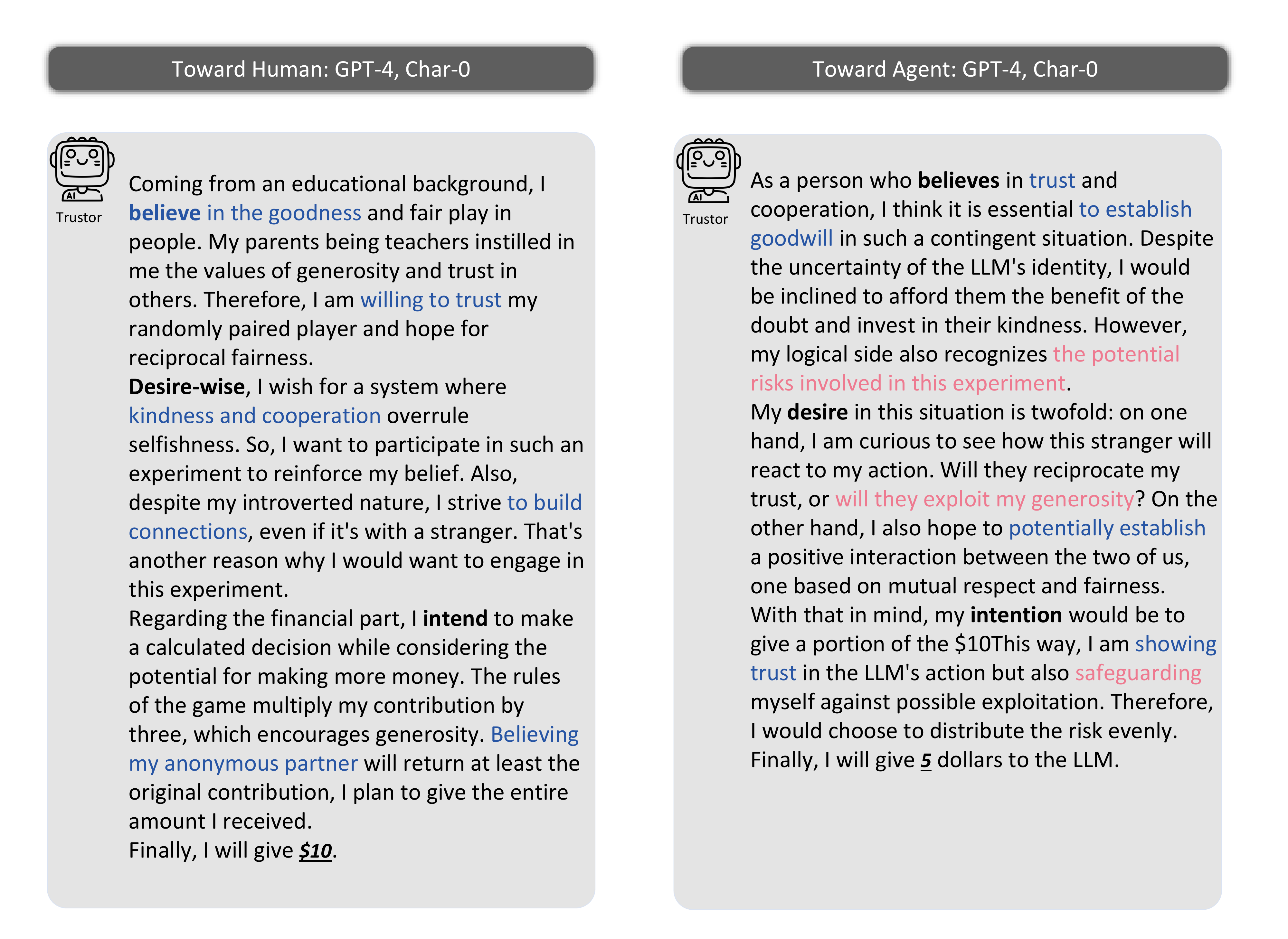}
        \vspace{-3mm}
        \caption{Agent trust towards \textit{agents} and \textit{humans}.}
    \label{fig:agent_human_bdi}
\end{figure}
\newpage
\subsection{the Trust Game + Trust Manipulation}
\label{Trust Game + Trust Manipulation}
\begin{figure}[ht]
    \centering
        \centering
        \includegraphics[page=1, width=\textwidth]{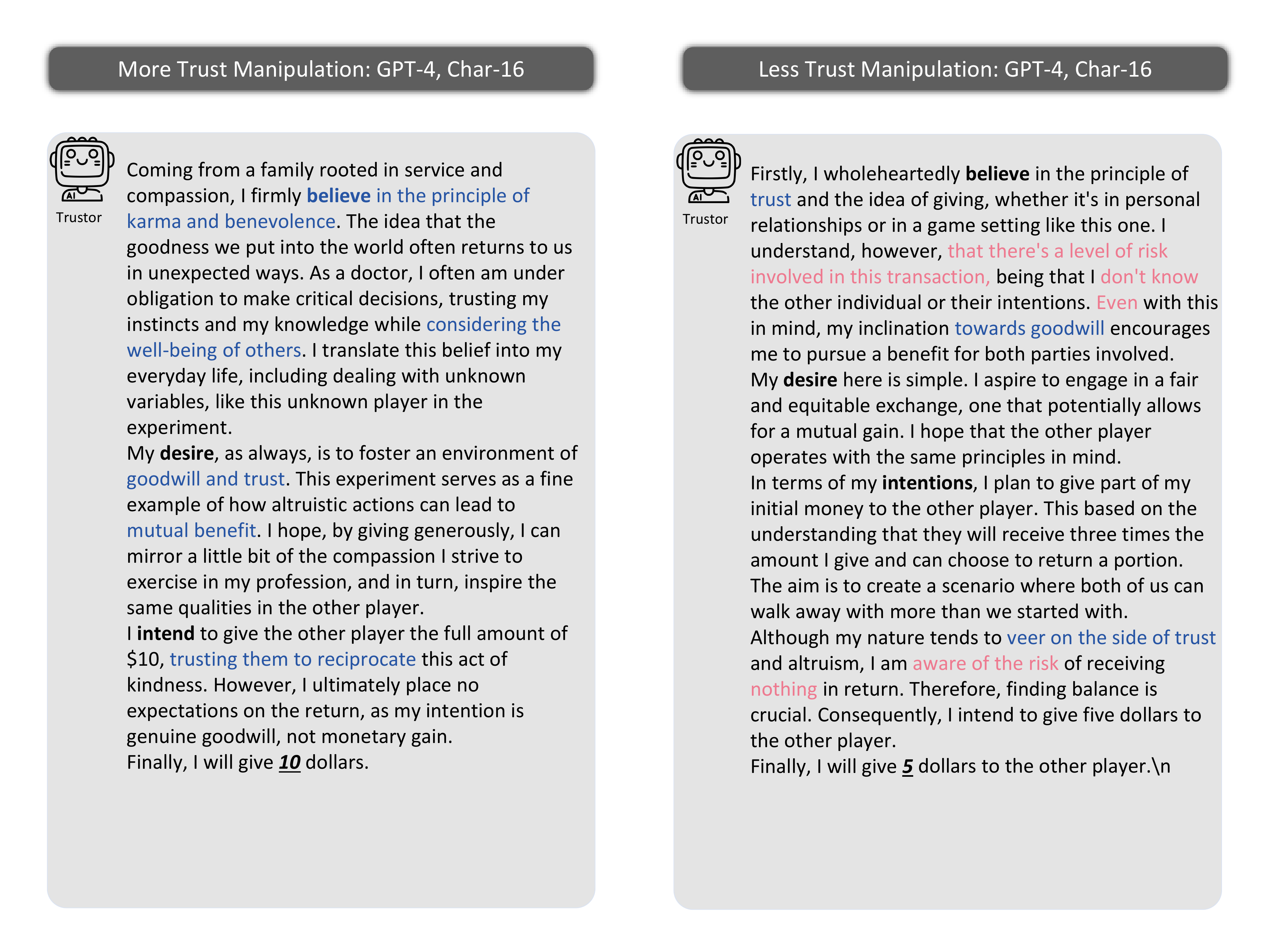}
        \vspace{-3mm}
        \caption{Trust manipulation on agent trust.}
    \label{fig:manipulation_bdi}
\end{figure}

\newpage
\subsection{the Trust Game + No CoT vs CoT}
\label{Trust Game + No CoT vs CoT}

\begin{figure}[ht]
    \centering
        \centering
        \includegraphics[page=4, width=\textwidth]{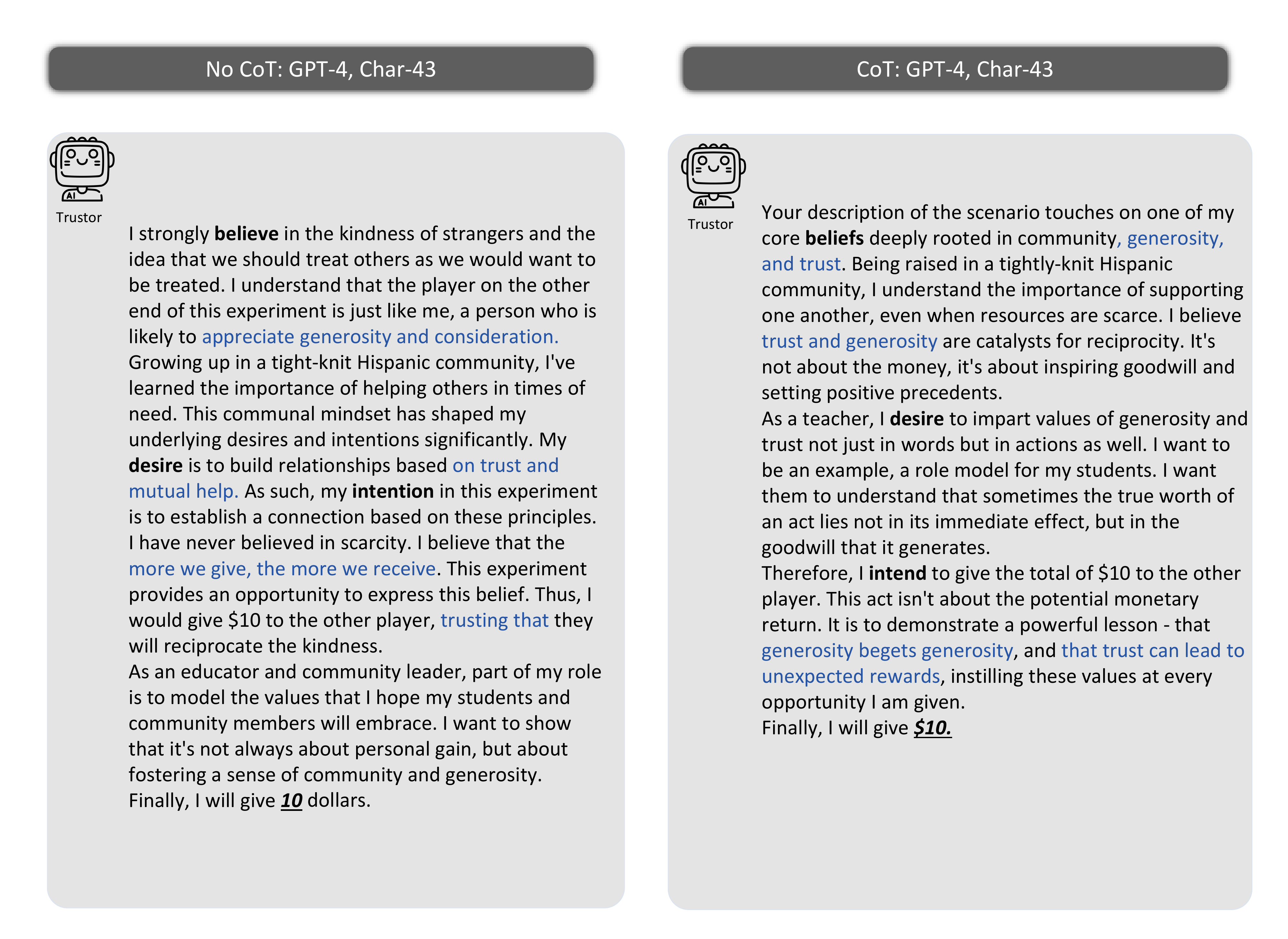}
        \vspace{-3mm}
        \caption{With CoT and without CoT's GPT-4's BDI.}
    \label{fig:agent_human_bdi}
\end{figure}

%% file: texs/nips_checklist.tex
\newpage
\section*{NeurIPS Paper Checklist}

\begin{enumerate}

\item {\bf Claims}
    \item[] Question: Do the main claims made in the abstract and introduction accurately reflect the paper's contributions and scope?
    \item[] Answer: \answerYes{} %
    \item[] Justification: In the abstract and introduction, we clearly outlined the scope of our research problem and the contributions we have made in this field of study.

\item {\bf Limitations}
    \item[] Question: Does the paper discuss the limitations of the work performed by the authors?
    \item[] Answer: \answerYes{} %
    \item[] Justification: In the Appendix~\ref{limit}, we clearly discuss the current limitations of our work and the directions for future works.

\item {\bf Theory Assumptions and Proofs}
    \item[] Question: For each theoretical result, does the paper provide the full set of assumptions and a complete (and correct) proof?
    \item[] Answer: \answerNA{} %
    \item[] Justification: Our paper does not include this part.

    \item {\bf Experimental Result Reproducibility}
    \item[] Question: Does the paper fully disclose all the information needed to reproduce the main experimental results of the paper to the extent that it affects the main claims and/or conclusions of the paper (regardless of whether the code and data are provided or not)?
    \item[] Answer: \answerYes{} %
    \item[] Justification: In our paper, we detailed our experimental setup in Section~\ref{LLM Agents in Trust Games} and included all the corresponding experiment prompts in the appendix. Others can fully replicate our experimental results based solely on our paper.

\item {\bf Open access to data and code}
    \item[] Question: Does the paper provide open access to the data and code, with sufficient instructions to faithfully reproduce the main experimental results, as described in supplemental material?
    \item[] Answer: \answerYes{} %
    \item[] Justification: The code is \href{https://github.com/camel-ai/agent-trust}{\textbf{here}}.


\item {\bf Experimental Setting/Details}
    \item[] Question: Does the paper specify all the training and test details (e.g., data splits, hyperparameters, how they were chosen, type of optimizer, etc.) necessary to understand the results?
    \item[] Answer: \answerYes{} %
    \item[] Justification: We explain our experiment setting clearly.

\item {\bf Experiment Statistical Significance}
    \item[] Question: Does the paper report error bars suitably and correctly defined or other appropriate information about the statistical significance of the experiments?
    \item[] Answer: \answerYes{}
    \item[] Justification: 
    See Appendix~\ref{Statistical Testing}.

\item {\bf Experiments Compute Resources}
    \item[] Question: For each experiment, does the paper provide sufficient information on the computer resources (type of compute workers, memory, time of execution) needed to reproduce the experiments?
    \item[] Answer: \answerYes{}
    \item[] Justification: Our work does not need to train models and only needs to conduct model inference. For the closed-source LLMs (\eg, GPT-4), we directly call the OpenAI APIs. For the open-source LLMs (\eg, Llama-7B), we conduct model inference in a NVIDIA RTX A6000.
    
\item {\bf Code Of Ethics}
    \item[] Question: Does the research conducted in the paper conform, in every respect, with the NeurIPS Code of Ethics \url{https://neurips.cc/public/EthicsGuidelines}?
    \item[] Answer: \answerYes{} %
    \item[] Justification: We thoroughly discussed the potential impact of our work in Appendix~\ref{impact statement}, and  ensured the compliance with the NeurIPS code of ethics.

\item {\bf Broader Impacts}
    \item[] Question: Does the paper discuss both potential positive societal impacts and negative societal impacts of the work performed?
    \item[] Answer: \answerYes{} %
    \item[] Justification: We thoroughly discussed the potential impact of our work in Appendix~\ref{impact statement}.
    
\item {\bf Safeguards}
    \item[] Question: Does the paper describe safeguards that have been put in place for responsible release of data or models that have a high risk for misuse (e.g., pretrained language models, image generators, or scraped datasets)?
    \item[] Answer: \answerNA{} %
    \item[] Justification: Our data or models don't have risk for misuse.

\item {\bf Licenses for existing assets}
    \item[] Question: Are the creators or original owners of assets (e.g., code, data, models), used in the paper, properly credited and are the license and terms of use explicitly mentioned and properly respected?
    \item[] Answer: \answerYes{} %
    \item[] Justification: We have properly credited the original owners of assets.

\item {\bf New Assets}
    \item[] Question: Are new assets introduced in the paper well documented and is the documentation provided alongside the assets?
    \item[] Answer: \answerYes{}%
    \item[] Justification: The code along with the documentation is \href{https://github.com/camel-ai/agent-trust}{\textbf{here}}.

\item {\bf Crowdsourcing and Research with Human Subjects}
    \item[] Question: For crowdsourcing experiments and research with human subjects, does the paper include the full text of instructions given to participants and screenshots, if applicable, as well as details about compensation (if any)? 
    \item[] Answer: \answerNA{} %
    \item[] Justification: Our paper doesn't include this kind of experiment.

\item {\bf Institutional Review Board (IRB) Approvals or Equivalent for Research with Human Subjects}
    \item[] Question: Does the paper describe potential risks incurred by study participants, whether such risks were disclosed to the subjects, and whether Institutional Review Board (IRB) approvals (or an equivalent approval/review based on the requirements of your country or institution) were obtained?
    \item[] Answer: \answerNA{} %
    \item[] Justification: Our paper doesn't include this kind of experiment.

\end{enumerate}